\documentclass{article}
\usepackage[final]{colm2026_conference}

\usepackage[nopatch=footnote]{microtype}
\usepackage{hyperref}
\usepackage{url}
\usepackage{booktabs}
\usepackage{amsmath,enumitem}
\usepackage{siunitx}
\usepackage{xspace}
\usepackage{tcolorbox}

\usepackage{lineno}

\definecolor{darkblue}{rgb}{0, 0, 0.5}
\hypersetup{colorlinks=true, citecolor=darkblue, linkcolor=darkblue, urlcolor=darkblue}

\usepackage{cinzel}
\definecolor{cinered}{RGB}{180,30,30}
\newcommand{\reframed}{\kern-0.3em{\cinzelblack%
Re\tikz[baseline=(f.base)]{\node[draw=cinered, rounded corners=1.5pt, inner sep=1.5pt, outer sep=0pt, line width=0.6pt, text=cinered](f){frame};}d}\xspace}

\usepackage{pgfplots}
\pgfplotsset{compat=1.18}

\usepackage{subcaption}

\definecolor{reframed-green}{HTML}{236623}
\definecolor{reframed-silver}{HTML}{B8B8B7}
\definecolor{reframed-gray}{HTML}{CFCFCE}
\definecolor{reframed-yellow}{HTML}{E8B017}
\definecolor{reframed-gold}{HTML}{DA9100}
\definecolor{reframed-red}{HTML}{B93B32}

\usepackage{pifont}
\usepackage[table]{xcolor}

\newcommand*\colourcheck[2]{%
	\expandafter\newcommand\csname #1check\endcsname{\textcolor{#2}{\ding{52}}}%
}
\colourcheck{gold}{reframed-gold}
\colourcheck{silver}{reframed-silver}

\newcommand{\redcross}{\textcolor{reframed-red}{\ding{56}}}

\newcommand{\videoIcon}{\raisebox{-0.2em}{\includegraphics[height=1em]{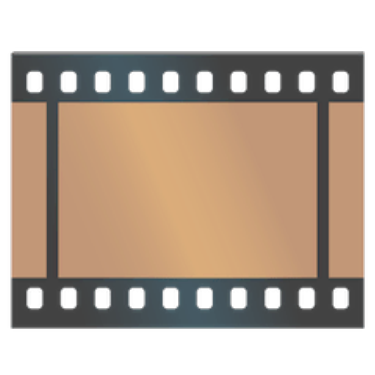}}}
\newcommand{\audioIcon}{\raisebox{-0.2em}{\includegraphics[height=1em]{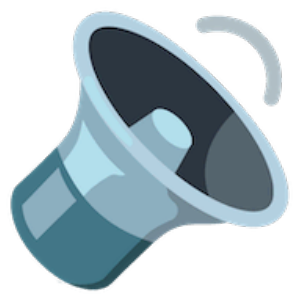}}}
\newcommand{\adIcon}{\raisebox{-0.2em}{\includegraphics[height=1em]{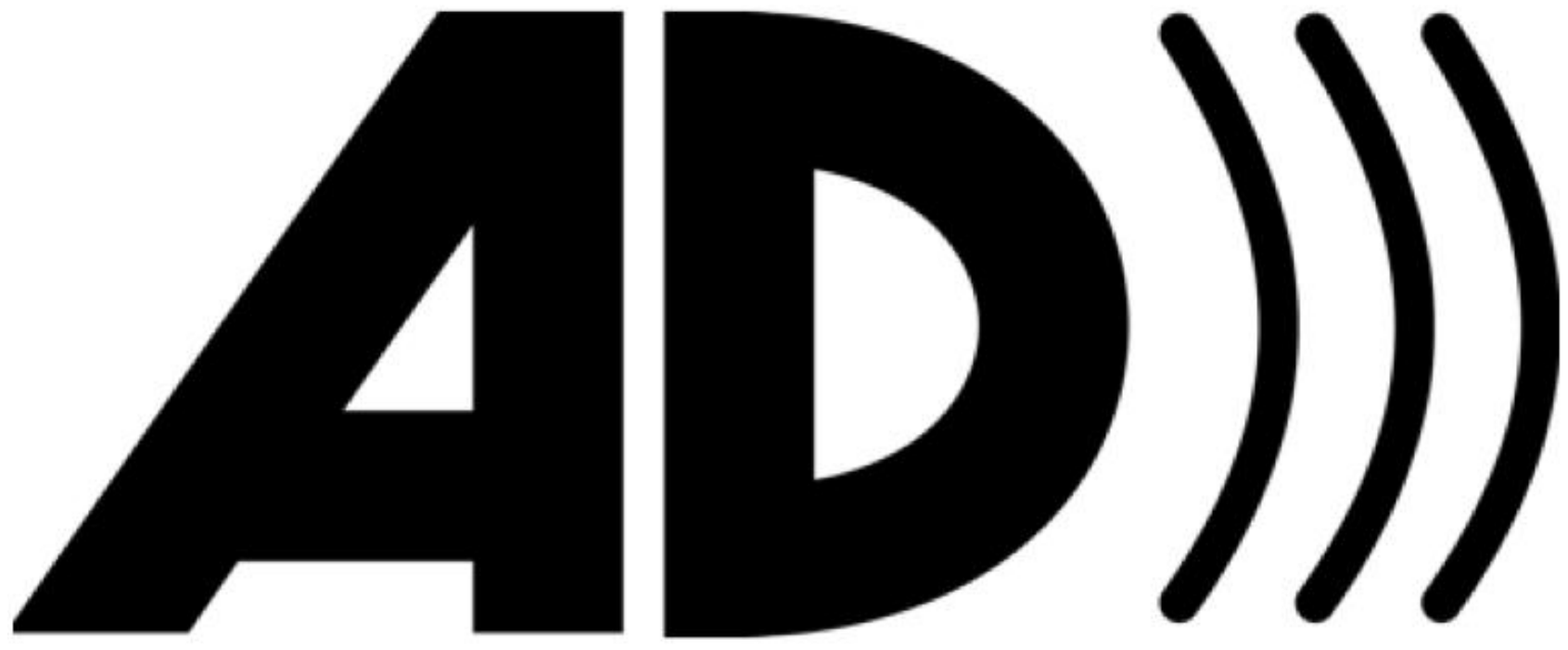}}}
\newcommand{\subtitlesIcon}{\raisebox{-0.2em}{\includegraphics[height=1em]{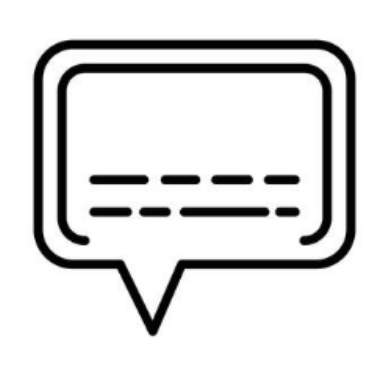}}}
\newcommand{\screenplayIcon}{\raisebox{-0.2em}{\includegraphics[height=1em]{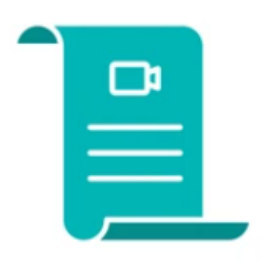}}}

\usepackage{placeins}

\usepackage{wrapfig}

\usepackage{float}

\usepgfplotslibrary{fillbetween}
\usetikzlibrary{backgrounds}
\pgfdeclarelayer{background}
\pgfdeclarelayer{main}
\pgfdeclarelayer{foreground}
\pgfsetlayers{background,main,foreground}

\newcommand{\citeposs}[1]{\citeauthor{#1}'s (\citeyear{#1})}

\usepackage{listings}

\lstdefinelanguage{XML}
{
	morestring=[b]",
	morestring=[s]{>}{<},
	morecomment=[s]{<?}{?>},
	stringstyle=\color{black},
	identifierstyle=\color{darkblue},
	keywordstyle=\color{cyan},
	morekeywords={xmlns,version,type}
}

\usepackage{tcolorbox}

\usepackage{tikz}
\usepackage{graphicx}

\usetikzlibrary{arrows.meta, positioning, calc, patterns}
\pgfdeclarepatternformonly{hatchgap}{\pgfqpoint{-1pt}{-1pt}}{\pgfqpoint{7pt}{7pt}}{\pgfqpoint{6pt}{6pt}}{
	\pgfsetlinewidth{2.12pt}
	\pgfpathmoveto{\pgfqpoint{0pt}{0pt}}
	\pgfpathlineto{\pgfqpoint{6pt}{6pt}}
	\pgfusepath{stroke}
}

\usepackage{algorithm}
\usepackage{algpseudocode}

\lstdefinestyle{xmlstyle}{
	basicstyle=\small\ttfamily,
	backgroundcolor=\color{xmlbgcolor},
	breaklines=true,
	frame=single,
}
\tcbuselibrary{breakable}

\pgfplotsset{
	/pgfplots/ybar legend/.style={
			/pgfplots/legend image code/.code={%
					\draw[##1,/tikz/.cd,yshift=-0.25em]
					(0cm,0cm) rectangle (3pt,0.8em);},
		},
}

\usepgfplotslibrary{statistics}

\usepackage[rightcaption]{sidecap}

\usepackage{fvextra}
\tcbuselibrary{breakable}

\title{\reframed: Towards Realistic Audio Description\\Generation for Movies}

\author{Igor Sterner, Mirella Lapata, Alex Lascarides
\& Frank Keller \\
  School of Informatics \\
  University of Edinburgh \\
  United Kingdom \\
  {\tt igor.sterner@ed.ac.uk, }%
  \{%
  {\tt mlap,alex,keller}%
  \}%
  {\tt @inf.ed.ac.uk}\\
  \url{https://igorsterner.github.io/reframed/}
  }

\begin{document}

\ifcolmsubmission
\linenumbers
\fi

\maketitle

\begin{abstract}
Audio Description (AD) is a verbal narration of key visual content in videos, enabling access for visually impaired audiences. Unlike standard video captioning, AD is a structured editorial task: descriptions must be inserted into gaps in dialogue and must convey only what is needed to understand the narrative being told. However, existing approaches formulate AD generation in an artificial setting where both the content and timing of descriptions are pre-specified, reducing the task to clip-level captioning. They further rely on noisy transcription and alignment pipelines, and lack the rich parallel data required for modeling narrative context. We introduce a new formulation of AD generation in which models must jointly decide \emph{what} to describe and \emph{when} to do it. To support this, we present \reframed, a high-quality dataset of 2{,}023 videos that span 3{,}302 scenes from 206 movies, with professional AD transcripts (both American and British versions), professional subtitles and aligned screenplays. We also provide a manually curated challenge set that pairs full movies with multiple AD references, together with evaluation protocols that leverage dialogue gaps and multi-reference comparisons. Experiments with state-of-the-art AD systems and multimodal LLMs show that they outperform trivial baselines but fall far short of expert human performance. Our dataset and benchmark establish a new foundation for research on video understanding.
\end{abstract}

\section{Introduction}

Audio Description (AD) is an accessibility service for visually impaired individuals.
It is a verbal narration of key visual information necessary for understanding a video.
The narration is timed to fit into natural gaps in dialogue, so that it minimally interrupts the original soundtrack.
It is recorded and delivered via an earpiece or a secondary audio channel.
Both American and British legislation will soon mandate AD at scale: Title II of the Americans with Disabilities Act will require public entities to provide AD for pre-recorded video, while the UK's Media Act sets a streaming quota of 10\% by 2030.
These mandates create an urgent need for automating AD in a way that addresses real-world requirements.
However, current systems have only tackled limited aspects of the task.

For movies, the full task of creating AD requires a describer to first understand the movie's narrative arc before deciding which visual elements are needed to be able to follow the story.
The describer must then formulate concise and evocative language that fits within dialogue gaps.
Consider the AD in Figure~\ref{fig:dragon-tattoo}, which is for the final scene in the 2011 murder-mystery \emph{The Girl with the Dragon Tattoo}.\footnote{The video is available from second 37 onwards at \url{https://rottentomatoes.com/m/the_girl_with_the_dragon_tattoo/videos/ofdvX951xl2S}}
This excerpt shows that the creation of AD requires fine-grained visual understanding, character tracking, location and object tracking, conveying location changes, time, and mood.
The AD also does not interrupt the dialogue.

\begin{figure}[t]
    \centering
    \input{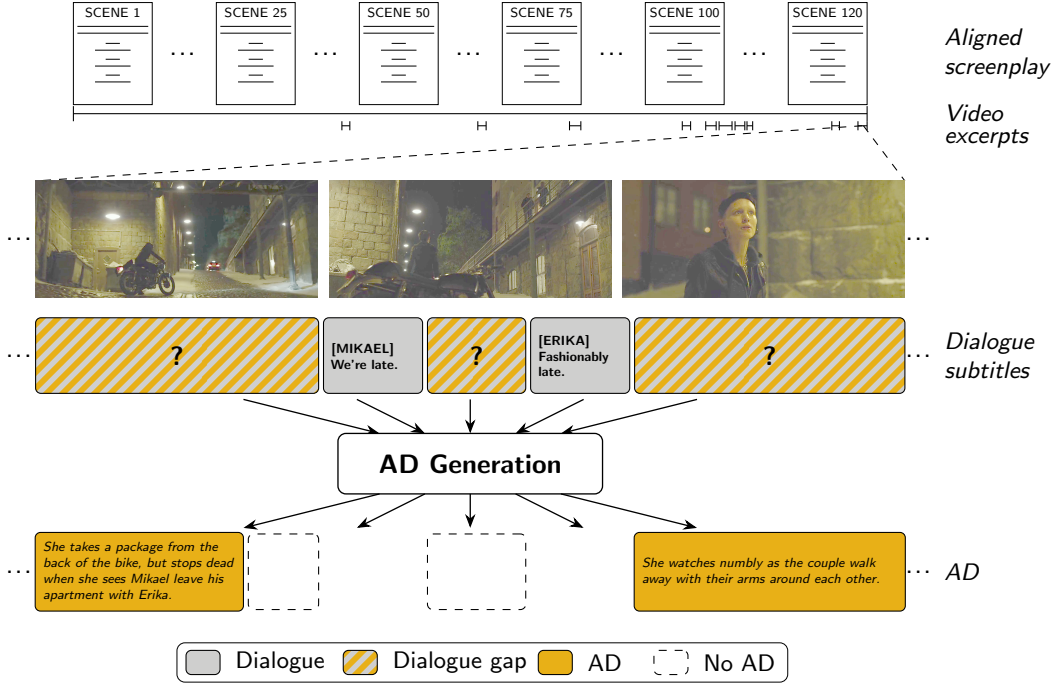}
    \vspace{-10pt}
    \caption{The \protect\reframed task formulation. Given input of a video, dialogue subtitles, and the spans of gaps between them, a model must decide both \textit{when} and \textit{what} to describe. Narrative context can be made available from the movie's screenplay and other scene videos.}
    \label{fig:headline-figure}
\end{figure}

AD presents distinctive challenges that current systems are far from solving.
In addition to fine-grained multimodal understanding over hours of content, AD generation requires precise control over when to describe and the ability to determine which elements of the movie are \emph{narratively} salient. Such elements are not always \emph{visually} salient: in the above scene, the two lovers are visible only from a distance (the middle frame of Figure~\ref{fig:headline-figure}), yet their actions are essential to the narrative and must be described.

\begin{figure}[t]
\centering
\begin{tcolorbox}[
  colback=reframed-gray!10,
  colframe=reframed-silver,
  boxrule=0.5pt,
  arc=3pt,
  left=0pt, right=0pt, top=0pt, bottom=0pt,
  width=\columnwidth,
  fontupper=\small
]

\textit{Night. Lisbeth rides her motorcycle down a cobbled street with snow piled at the curbside. She parks on a corner outside Mikael's apartment and removes her crash helmet. She takes a package from the back of the bike, but stops dead when she sees Mikael leave his apartment with Erika.
}

\smallskip
\textbf{[MIKAEL] We're late.}\\
\textbf{[ERIKA] Fashionably late.}
\smallskip

\textit{She watches numbly as the couple walk away with their arms around each other. Lisbeth looks down as Mikael and Erika get into a waiting taxi. Lisbeth turns away and flings the package which has the card attached to it into a nearby dumpster. She puts her crash helmet back on, gets on her bike and starts it up. A solitary figure, she rides off into the night.}
\end{tcolorbox}
\vspace{-8pt}
\caption{Excerpt of Audio Description (\emph{italics}) and natural dialogue (\textbf{bold}) from \textit{The Girl with the Dragon Tattoo} (2011). The AD requires fine-grained visual understanding, character and object tracking, and precise temporal placement,  all without interrupting the soundtrack.}
\label{fig:dragon-tattoo}
\vspace{-5pt}
\end{figure}

Recent multimodal LLMs are candidates for meeting these challenges. They exhibit narrative understanding and can track characters, infer intent, and maintain coherence over long contexts, while grounding this reasoning in visual input. Moreover, the structured nature of AD, with fixed candidate gaps (between dialogue) and clear stylistic conventions, aligns naturally with the instruction-following capabilities of current models. Their ability to meet these challenges, however, cannot be meaningfully assessed as existing datasets and task formulations do not reflect the demands of AD in practice.

AD is created for nearly every major movie release, sometimes in multiple versions for a single movie, offering a rich source of expert-authored data that current datasets do not fully exploit.
They all rely on unvalidated automatic speech recognition (ASR) to transcribe audio AD tracks, resulting in transcriptions we found to be extremely noisy.
More fundamentally, prior work reduces AD generation to an artificial setting in which both AD \emph{content-selection} and \emph{placement} are fixed: models are given the clip that corresponds to the elements described in the reference AD.
This collapses the problem to clip-level video captioning, ignoring the core editorial challenge of AD: deciding which elements to describe and when.
Finally, existing datasets provide almost none of the rich parallel data (screenplays, professional dialogue subtitles, multiple AD versions) that would benefit novel treatments of the task.

To address these limitations, we reformulate AD generation as a joint decision problem over \emph{when} and \emph{what} to describe, and introduce \reframed, a dataset and benchmark for this setting. Our task formulation matches real-world requirements: given a video and gaps in dialogue, a model must determine where descriptions are needed and what to say. Uniquely, for both training and evaluation data we provide \emph{two} reference English AD versions, screenplays aligned at the level of scenes, professional English dialogue subtitles and SDH subtitles, which include salient non-verbal audio for hard-of-hearing audiences. Moreover, we adapt existing evaluation metrics to our novel AD task and benchmark current AD generation systems and LLMs under this formulation.
Our contributions are:

\begin{itemize}
\item \textbf{A realistic task formulation} for AD generation (a model must decide when and what to describe), together with evaluation metrics that exploit multiple references.
\item \reframed, a dataset of 2{,}023 video excerpts (3{,}302 scenes) from 206 movies with dual AD versions (US and UK), dialogue subtitles, SDH, screenplays, and video.
\item  The \reframed \textbf{challenge set} of 10 fully manually annotated movies with 2--3 professionally transcribed AD versions each, scene boundaries, AD-to-scene labels, screenplay alignment, and professional subtitles.
\item  \textbf{A suite of benchmarks} for ASR quality, temporal alignment, screenplay alignment, scene segmentation, AD splitting, and AD generation on the above data.
\end{itemize}

\FloatBarrier

\section{Related Work}

\textbf{AD as Video Captioning.}\label{sec:related_work_generation} The task has commonly been framed as a form of video captioning: given a short video clip, generate a natural language sentence that describes its visually salient content.
Most current work follows this formulation and finetunes multimodal models for the task, augmenting input with additional signals such as broader visual context, character information, or scene-level cues \citep{wang-2021-toward,han-2023-autoad_i,han-2023-autoad_ii,han-2024-autoad_iii,lin-2024-movie_seq, lee-2025-ssms_ad,deganutti-2025-dante, wang-2025-uni_ad, ye-2025-focusedad}.
DistinctAD \citep{fang-2024-distinct_ad} trains using character information, visual features and text from nearby clips, with architectural components and learning objectives that reward distinctive descriptions.
\citet{xie-2024-autoad_zero} propose to first generate a detailed description of a clip and then summarize it into AD; in Shot-by-Shot \citep{xie-2025-shot_by_shot}, this approach is improved by incorporating knowledge of shot boundaries and scale.
Other work relies on LLM prompting \citep{lin-2023-mm_vid,chu-2024-llm_ad,zhang-2024-mm_narator,ye-2024-mmad, khandelwal-2025-coherent};
\citet{gao-2025-ad_review} provide a broader survey of the area.
A related line of work incorporates screenplays as narrative context.
Screenplays have been estimated to contain approximately 60\% of the information required for AD generation \citep{lakritz-2006-semi}, and later work shows that screenplay context can improve generation quality \citep{rimle-2020-enriching,campos-2023-machine,park-2025-narr_ad}.
Despite this progress, all these approaches address a simplified formulation of AD creation: they assume AD placement is given and only generate the description sentence.
Our benchmark is the first to require systems to determine placement and content jointly.

\textbf{Movie Datasets with AD.} Many movie datasets exist without AD \citep{tapaswi-2016-movieqa,gu-2018-ava,vicol-2018-moviegraphs,huang-2020-movienet,bain-2020-condensed,wu-2021-towards,sun-2023-symon,sun-2024-m_symon,wu-2024-longvideobench,zaranis-2025-mf2}.
The first large-scale dataset containing AD is LSMDC \citep{rohrbach-2017-lsmdc}, which pairs semi-automatically transcribed AD sentences with manually extracted movie clips that correspond with the described elements, replacing character names with placeholders.
MAD \citep{soldan-2022-mad} scales this up by fully automatically transcribing AD tracks and aligning them to full movies via audio cross-correlation.
Evaluation material reuses LSMDC timecoding, which is suited to their task of video grounding.
\citet{han-2023-autoad_i} repurpose a subset of this evaluation data for a full-movie AD generation benchmark; the adoption of the LSMDC timecoding means that the time-span of the element to be described in the generation is provided as input.
The placeholder substitution for character names is also retained.
\citet{han-2024-autoad_iii} introduce CMD-AD by aligning AD segments to movie scenes from YouTube and allowing temporal offsets and speed adjustments to account for differences in broadcast frame rates (e.g.,~23.976fps NTSC vs.\ 25fps PAL).

These datasets share three limitations. First, AD transcripts are produced automatically, yet transcription quality has never been evaluated; in Appendix~\ref{sec:asr} we show this is a major issue. Second, the alignment of AD tracks to movie scenes has also not been validated; in Appendix~\ref{sec:scene-alignment} we show that the method used for the construction of CMD-AD is very noisy. Third, the MAD-based evaluation set inherits LSMDC timecoding, transcription errors and character name replacement, issues that have propagated into subsequent work.

\textbf{Evaluation of AD Generation.}\label{sec:related_work_evaluation} Appendix Table~\ref{tab:metrics} summarizes the many metrics used to evaluate AD generation. They include n-gram \citep[e.g. METEOR,][]{banerjee-2005-meteor}, image-captioning \citep[e.g. CIDEr,][]{vedantam-2015-cider}, and QA-based evaluation \citep{kala-2025-what}.
Dense video captioning metrics such as SODA \citep{fujita-2020-soda} and its temporal counterpart \citep{batra-2022-temporal} align generated and reference elements using their time spans before scoring them with F-measure.
However, these metrics inherit limitations from the simplified task setting. Because prior work assumes fixed AD placement, evaluation operates at the level of individual captions matched to predefined clips and therefore cannot assess whether a system correctly decides \emph{when} to insert a description as well as \emph{what} visual element needs to be described. Prior work also ignores the task's subjectivity and typically uses only a single reference. We adapt existing metrics to our open-ended task formulation, and use multiple references throughout.

\FloatBarrier

\begin{table}[t]
\begin{small}
\centering
\begin{tabular}{llc@{\hspace{1mm}}c@{\hspace{1mm}}c@{\hspace{1mm}}c@{\hspace{1mm}}c@{\hspace{2mm}}r@{\hspace{2mm}}rc@{\hspace{1mm}}c@{\hspace{1mm}}c@{\hspace{1mm}}c@{\hspace{1mm}}c@{\hspace{2mm}}r@{\hspace{2mm}}r@{\hspace{2mm}}r}
\toprule
\multicolumn{2}{c}{} & \multicolumn{7}{c}{TRAINING} & \multicolumn{7}{c}{TESTING}  & TASK \\
Name & Type & \videoIcon & \audioIcon & \subtitlesIcon & \screenplayIcon & \adIcon & hrs & \#refs & \videoIcon & \audioIcon & \subtitlesIcon & \screenplayIcon & \adIcon & hrs & \#refs & \\
\cmidrule(lr){1-1} \cmidrule(lr){2-2} \cmidrule(lr){3-9} \cmidrule(lr){10-16} \cmidrule(lr){17-17}
LSMDC (AD) & 5s clips &  \goldcheck & \goldcheck & \goldcheck & \redcross & \goldcheck & 134 & 1 & \goldcheck & \goldcheck & \goldcheck & \redcross & \silvercheck & 12 & 1 & VC \\
MAD & full movies & \redcross & \redcross & \redcross & \redcross & \silvercheck & 990 & 1 & \redcross & \redcross & \redcross & \redcross & \silvercheck & 217 & 1 & VG \\
MAD-v2-eval & full movies & -- & -- & -- & -- & -- & -- & -- & \redcross & \redcross & \silvercheck & \redcross & \silvercheck & 19 & 1 & VC \\
\rowcolor{reframed-yellow!30}
MAD-v3-eval & full movies & -- & -- & -- & -- & -- & -- & -- & \redcross & \redcross & \goldcheck & \redcross & \goldcheck & 19 & 1 & ASR \\
\rowcolor{reframed-yellow!30}
\reframed & full movies & -- & -- & -- & -- & -- & -- &  -- &  \goldcheck & \goldcheck & \goldcheck & \goldcheck & \goldcheck & 21 & 2-3 & AD \\ \midrule
CMD-AD & 2m scenes & \goldcheck & \goldcheck & \silvercheck & \redcross & \silvercheck & 307 & 1 & \goldcheck & \goldcheck & \silvercheck & \redcross & \silvercheck & 22 & 1 & VC \\
\rowcolor{reframed-yellow!30}
\reframed & 2m scenes & \goldcheck & \goldcheck & \silvercheck  & \silvercheck  & \silvercheck & 77  & 2 &  \goldcheck & \goldcheck & \goldcheck & \goldcheck & \goldcheck & 4 & 2 & AD \\
\bottomrule
\end{tabular}
\caption{Comparison of publicly available AD datasets. We report whether video (\videoIcon), audio (\audioIcon), subtitles (\subtitlesIcon), screenplays (\screenplayIcon), and AD transcripts (\adIcon) are available at gold-standard quality (\goldcheck), silver-standard quality (\silvercheck; transcription or alignment), or not at all (\redcross), alongside total hours of content and number of AD references.
VC = video captioning; VG = video grounding; ASR = automatic speech recognition (and diarization) of AD; AD = our task of AD generation.
Training includes validation data, if it is provided.
CMD-AD hours are based on the available video data.
Our contributions are highlighted.
}
\label{tab:datasets}
\end{small}
\end{table}

\section[Reframed]{The \reframed Dataset}

We now provide an overview of the \reframed dataset and its construction, including quantitative validation of each tool used (results are summarized in Section~\ref{sec:construction-pipeline}, with more details in Appendix~\ref{sec:previous_methods} and \ref{sec:new_data_collection_techniques}). Our design is guided by the goal of supporting AD generation as a joint decision problem over \emph{when} and \emph{what} to describe, informed by narrative context.

\subsection{Data and Benchmark Overview}\label{sec:overview-data}

The dataset is built from movie excerpts licensed by Fandango, with raw videos without watermarking available publicly on \url{rottentomatoes.com}.
Narrative context is provided from full-movie screenplays that we align to the movie timelines as well as other videos from the same movie.
We select movies with at least five movie excerpt videos and listings on \url{audiovault.net} for \emph{both} American and British AD versions. Table~\ref{tab:datasets} compares our dataset against publicly available alternatives, showing our contributions of providing more parallel data streams for models to learn from, multiple AD references for both training and evaluation material, and gold standard data for evaluation.
Following prior work \citep{bamman-2024-film}, we exclude animated movies as they represent a substantially different visual domain.

\textbf{AD Transcription and Alignment.}
We obtain transcription for all AD tracks: for evaluation movies, they are the result of professional human transcription; training movies rely on a Speechmatics-based pipeline (Speechmatics are a specialist ASR and diarization provider).
For each movie, we align the video excerpts to the full AD track (both American and British versions) using cross-correlation, accounting for possible PAL--NTSC speed differences and enforcing consistent alignment offsets across scenes from the same movie.

\textbf{Parallel Data Streams.} For each movie excerpt, we provide professional English dialogue-only and SDH subtitles sourced from \url{opensubtitles.org}.
They are aligned to video using ASR-based timestamp matching (Appendix~\ref{appx:sec:subtitles}). For the 85 movies with parsed screenplays available in MovieSum \citep{saxena-2024-moviesum},
we align screenplay scenes to movie scenes using Vecalign \citep{thompson-2019-vecalign} applied to joint dialogue and AD representations, enabling robust alignment despite script deviations.

\textbf{AD Splitting and Scene Segmentation.}
We split AD transcripts into segments that each correspond to one visual element.
For all evaluation data, the splitting is a part of the professional human transcription; for training data, it is from a learned splitting model.
We additionally apply a second learned model that predicts scene boundaries from AD segments.
Appendix~\ref{appx:intro-example} shows split elements for our earlier example.

\textbf{Dataset Splits.} The final dataset contains 206~movies and 2{,}023 video excerpts (average duration 144 seconds), spanning 3{,}302 scenes; 85 movies have aligned screenplays. We split the data into training, validation, and test sets such that the genre distribution is balanced across the splits (Appendix~\ref{sec:splitting}). The training set excludes movies appearing in previous evaluation sets. The validation set consists of~10 movies from existing evaluation datasets, with aligned screenplays, while the test set contains~10 recent movies (2023--2025) not present in prior datasets.
Validation and test movie excerpts benefit from professionally transcribed AD rather than automatic transcription (Appendix~\ref{appx:sec:subtitling}).

\textbf{Challenge Benchmark.} For full-movie evaluation, we provide a manually annotated benchmark of 10 movies, each with 2--3 AD versions, exact scene boundaries, per-element scene labels, scene sequences, screenplay alignment, and professional dialogue and SDH subtitles (Appendix~\ref{appx:sec:challenge}).
All annotations were performed manually, either by a professional post-production company or by a trained annotator.
Seven of these movies are drawn from MAD-v3-eval; we additionally include \textit{Dune: Part Two} (2024), \textit{Barbie} (2023), and \textit{Priscilla} (2023), selected for their award-winning AD tracks.
The resulting set of ten movies spans 20 years of cinema and covers a diverse range of genres --- 14 out of the 19 IMDb genre categories in the training set are represented.

\subsection{The Construction Pipeline} \label{sec:construction-pipeline}

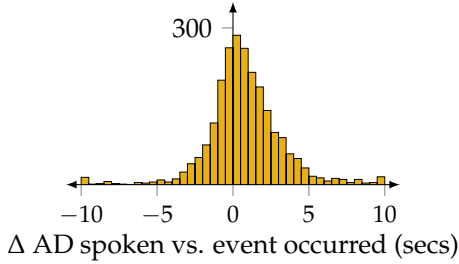
\begin{SCfigure}[][t]
  \centering
  \begin{tikzpicture}
  \begin{axis}[
      xlabel={$\Delta$ AD spoken vs. event occurred (secs)},
      axis lines=middle,
      inner axis line style={<->, >=latex},
      xlabel style={at={(axis description cs:0.5,0)}, anchor=north, yshift=-15pt},
      xmin=-11, xmax=11,
      ymin=0,
      ymax=350,
      ytick={300},
      xtick={-10,-5,0,5,10},
      tick align=outside,
      extra x ticks={0},
      extra x tick style={
          tick label style={anchor=north}
        },
      xticklabel style={font=\small},
      height=4cm,
      width=6cm
    ]
    \addplot[
      hist={
          bins=40,
          data min=-10,
          data max=10,
        },
      fill=reframed-yellow,
      draw=black,
      line width=0.3pt,
    ] table[y=diff, col sep=space]{data/time_diffs.dat};

  \end{axis}
\end{tikzpicture}
  \caption{Histogram comparing (a) MAD-v2-eval timecoding (v2; when described elements occur) and (b) new professional timecoding (v3; when the AD is spoken). Texts are normalized and aligned at the word level, and v2 segments matching a concatenation of one or more v3 segments are retained. We plot the midpoint differences.}
  \label{fig:timediff}
\end{SCfigure}

\textbf{Gold-standard Transcription.}
All downstream evaluation depends on gold-standard AD transcription, which is unavailable from existing data.
We hence contracted a film post-production company to perform fully manual AD transcription and timecoding to the nearest frame (approximately 0.04 seconds), and to provide segments each corresponding to a single visual element of the movie, which we will call description elements.
We call this new resource MAD-v3-eval.
Analysis of these annotations shows that AD narration occurs within 10 seconds of the described elements, but not necessarily at exactly the same time as them, reflecting the need to place AD within available dialogue gaps (Figure~\ref{fig:timediff}).

\textbf{AD Transcription and Diarization.} The first pipeline stage takes a full-length AD audio track as input and produces a transcription with speaker labels and timestamps. We evaluate multiple ASR systems against MAD-v3-eval and find that existing AD corpora are extremely noisy (WhisperX achieves a best overall word error rate, WER=12.8; the MAD-v2-eval transcriptions achieve 22.1). A system based on Speechmatics with character-name adaptation achieves WER=2.9 and is used to transcribe all training data (Appendix~\ref{sec:asr}).

\textbf{Temporal Alignment.} The second stage aligns scene videos to full-length AD tracks using cross-correlation \citep{soldan-2022-mad}, accounting for frame-rate differences between PAL and NTSC formats (Appendix Figure~\ref{appx:fig:speedup}). This approach is robust for typical scene durations (100\% alignment accuracy to within one frame for scenes of length 32 seconds or greater, which accounts for $>$99\% of the video extracts in our data), and we enforce consistency of alignment offsets across scenes within each movie (Appendix~\ref{sec:scene-alignment}).

\textbf{AD Splitting and Scene Segmentation.}
ASR produces sentence-level output, but our evaluation data is segmented into description elements.
We therefore segment transcribed AD sentences using a RoBERTa-based model trained on gold-standard annotations ($F_1$=64.8 vs. a comma-splitting baseline $F_1$=36.9; Appendix~\ref{sec:fragmentation}).
A classifier over these segments then identifies scene boundaries, exploiting textual cues (e.g., ``Now'') that AD guidelines encourage describers to include ($F_1$=59.2; Appendix~\ref{sec:scene-segmentation}).
This produces scalable, silver-standard full-movie scene segmentation consistent across the AD versions.
We combine these with scene boundaries derived from the start and end of each video excerpt (Appendix~\ref{sec:scene-segmentation}).

\textbf{Screenplay Alignment.} We align screenplay scenes to movie scenes by representing dialogue and AD jointly as a screenplay-like sequence. We then apply Vecalign \citep{thompson-2019-vecalign} to perform embedding-based dynamic programming alignment with support for scene insertions, deletions, and many-to-one merges ($F_1$=80.9 vs. a text-based dialogue-only baseline $F_1$=63.8; Appendix~\ref{sec:screenplay-alignment}).
We apply this method to all movies with parsed screenplays available in MovieSum \citep[85/206 movies;][]{saxena-2024-moviesum}.
This results in screenplay segments for 611 videos, after a filtering step that requires that every scene within a video successfully maps to the screenplay.

\section{Evaluation Metrics}

The constrained nature of AD lends itself well to evaluation, unlike many other long-form generation tasks.
In particular, AD is anchored to dialogue gaps, description elements are typically inserted within a short temporal window of their occurrence (Figure~\ref{fig:timediff}), and multiple references are available in our data.

We evaluate AD generation under our task formulation, where models must jointly decide \emph{what} visual information is narratively worth describing, \emph{when} it can be inserted without interrupting dialogue, and \emph{how} to express it concisely enough to fit the available time.
These factors make no single evaluation metric sufficient.
We therefore combine lexical similarity metrics, temporal alignment metrics, and QA-based semantic metrics.
CIDEr and METEOR provide interpretable reference-based measures of content similarity.
SODA-T evaluates whether generated descriptions are temporally aligned with professional AD references, using METEOR-based alignment, while QEval-T evaluates whether generated AD supports the same temporally grounded question answers as the reference.
For evaluation on our challenge set, we additionally provide human-level results based on a third expert-created AD transcript.

\textbf{Dialogue Gap-based Evaluation.}
Our first approach evaluates performance at the level of dialogue gaps, directly reflecting the main constraint that applies to AD placement.
We collect gaps in the professional dialogue subtitles of at least 1 second, and assign references and generations to the gaps that fully contain them (allowing a 1 second gap collar).
This assignment accounts for 90.2\% of reference AD description elements in our challenge set: 87.8\% (N=10{,}644) for American AD and 93.1\% (N={8},734) for British AD, consistent with stricter British guidelines regarding AD placement.
The remaining cases concern subtitled audio that evidently was not considered to preclude AD narration, such as musical lyrics or background dialogue.
These remain covered by our non-dialogue gap metrics.
Dialogue gaps that are not assigned any reference AD are discarded.
Generated text within each gap is compared against the corresponding reference texts.

For the comparison, we compute CIDEr and METEOR scores.
CIDEr natively supports multiple references.
However, its length penalty makes it poorly suited to long references.
For CIDEr, we therefore only retain dialogue gaps shorter than 20 seconds (see Appendix Table~\ref{tab:gap_length} for results on longer dialogue gaps).
For METEOR, we follow prior work and take the maximum score across references.
We retain all dialogue gaps for METEOR.

\textbf{Alignment-based Evaluation.}
Because multiple nearby dialogue gaps may plausibly host some descriptions, strict gap-level matching is too rigid.
SODA metrics (Section~\ref{sec:related_work_evaluation}) allow us to relax the requirement that generated descriptions must align exactly with the reference gap.
However, these metrics are unsuitable for direct application to AD, because they rely on the intersection of time spans to align generated and reference descriptions.
Our adaptation of SODA is operationally similar to the original procedure: we compute an optimal one-to-one monotonic alignment between the reference and generated description elements using dynamic programming.
The key difference in our approach is the alignment criterion.
Our alignment optimizes only semantic similarity (METEOR $F_1$) between matched pairs.
Sorted by start time, let us denote the sequence of $n$ reference description elements as $\mathcal{D} = (d_i)_{i=1}^n$ and the sequence of $\hat{n}$ generated description elements as $\hat{\mathcal{D}} = (\hat{d}_j)_{j=1}^{\hat{n}}$.
The alignment obtains $\mathcal{A}^\star=\arg\max_{\mathcal{A}}\sum_{(i,j)\in\mathcal{A}}\mathrm{METEOR}(d_i, \hat{d}_j)$, computed exactly with dynamic programming (see Appendix \ref{sub:soda_alignment}).
We define SODA-M as the resulting $F_1$ measure over METEOR scores and SODA-T as the proportion of matched pairs with timespan midpoints differing by less than $\tau=10$ seconds (chosen based on Figure~\ref{fig:timediff}; Appendix Figure~\ref{fig:attribution-sweep} shows a sweep):
\[
\mathrm{SODA\text{-}M}
=\frac{\sum_{(i, j)\in\mathcal A^\star}\mathrm{METEOR}(d_i, \hat{d}_j)}{\tfrac{1}{2}\left(|\mathcal{D}|+|\hat{\mathcal{D}}|\right)},
\qquad
\mathrm{SODA\text{-}T}
=\frac{1}{|\mathcal{D}|}\sum_{(i,j)\in\mathcal A^\star}
\mathbb{1}\!\left[|m(I_i)-m(\hat{I}_j)|<\tau\right]
\]
where $m(I_i)$ is the midpoint of the time interval $I_i$ of description element $d_i$.
Following prior SODA implementations, we take the maximum of each score across references.

\textbf{QA-based Evaluation.} We assess whether generated AD supports downstream understanding using a QA-based evaluation that captures both semantic correctness and temporal grounding.
We prompt OLMo 3 32B Think to generate multiple-choice questions from reference AD.
Then, we filter aggressively for quality, retaining only questions answered correctly by OLMo 3 32B Base using each reference AD version and incorrectly when the supporting AD segment is removed.
This ensures that questions depend on specific narrative content described in both AD references and that the wrong answer distractors are appropriate.
Questions are answered using the base LLM by comparing multiple-choice logits.
Prompts and more details on filtering are in Appendix~\ref{sec:qa-prompts}.

We define \textbf{QEval} and \textbf{QEval-T}, capturing semantic accuracy and temporally grounded accuracy.
For each question, let $y_j$ and $\hat{y}_j$ denote the gold-standard and predicted answers.
$I_j$ is given by the segment used to generate the question.
We predict $\hat{I}_j$ as the segment whose removal most reduces the probability of the predicted answer.
This attribution-based approach identifies which part of the generated AD supports the answer, and achieves round-trip consistency of 92.6\% (see stages 3$\rightarrow$4 in Appendix Table~\ref{tab:filtering-stages}).

\textbf{QEval} measures answer correctness (the proportion of questions with system-predicted answer matching the gold-standard answer) and \textbf{QEval-T} also enforces temporal alignment:
\[
    \text{QEval} = \frac{1}{N} \sum_{j=1}^{N} \mathbb{1}(\hat{y}_j = y_j) \ ; \
    \text{QEval-T} = \frac{1}{N} \sum_{j=1}^{N} \mathbb{1}(\hat{y}_j = y_j) \cdot \mathcal{C}_j \ ; \ \mathcal{C}_j = \mathbb{1}(|m(\hat{I}_j) - m(I_j)| < \tau)
\]
where $\tau=10$s (Figure~\ref{fig:timediff}).
QEval-T therefore passes judgement on both the inclusion of specific narrative content (correct answer) and when it is grounded (correct temporal alignment).
We retain only QA pairs that satisfy QEval-T under both reference AD versions; on average, there is one QA pair for every 3.6 reference description elements.
Significance testing uses two-tailed Monte Carlo permutation tests with $R=10{,}000$ and $\alpha=0.05$.

\section{Experiments}

\subsection{Experimental Setup}

We demonstrate the application of our evaluation metrics to the \reframed challenge set;
results on the test set are in Appendix~\ref{appx:sec:extra-results}.
Our objectives are to assess:
(1) \textbf{upper and lower bounds} on task performance;
(2) progress made by \textbf{pre-existing models};
and
(3) the extent to which \textbf{current LLMs} can perform our task.

\textbf{Upper and Lower Bounds.}
As an approximate upper bound, we take a third professional AD transcript (available for one challenge movie) and evaluate it as a system output against the remaining two references. This gives a human-expert ceiling: i.e., the score a professional describer achieves on our metrics when judged against other professionals describing the same film.
As a lower bound, we show results from a greedy baseline that fills each dialogue gap with randomly sampled training-set description elements until the gap is full and no further description elements can be added.

\textbf{Pre-existing Modeling.}
Existing models were not designed for our task formulation (Section~\ref{sec:related_work_generation}), so we give them the best possible chance by providing all information their designers made available, including gold-standard AD placement and content selection.
For this evaluation, MAD-v2-eval timestamps are mapped to the new professional MAD-v3-eval timestamps, as is required for our evaluation ($>$99\% of MAD-v2-eval segments receive mapped timestamps; the remaining are discarded).
Seven of our ten challenge movies overlap with MAD-v2-eval, for which the required inputs are available; our measurements are an upper bound on these models' true performance.

\textbf{Current LLMs.}
Many recent LLMs support up to one million tokens of context, and some are trained on multi-modal input, making full-movie input feasible.
We evaluate two suitable model families.
\textbf{Qwen 3.5} \citep{qwen3.5} are open-weight models designed for video input and supporting context extension to one million tokens \citep[via YaRN-based RoPE scaling,][]{peng-2024-yarn}; we run the 27B dense model.
\textbf{Gemini 3.1 Flash-Lite} \citep{gemini3.1flashlite} is a proprietary model at comparable per-token cost; we run it with \texttt{thinking=high} and default media resolution.
All video input is at one frame per second with no audio.
Qwen inference uses vLLM on two H200 GPUs; Gemini is accessed via the official paid API.

The models receive video, timed dialogue subtitles and a symbolic listing of dialogue gaps exceeding one second, and are instructed to generate zero or more timed description sentences per gap.
In principle, they can take the full movie as input and output an AD script for it.
However, we found that after generating approximately ten minutes of an AD script (c.~1k tokens), generation degenerates.
For our experiments with full-movie input, we therefore instruct the models to generate descriptions for ten-minute chunks (specified via timestamps in the instructions).
We compare this approach against providing each ten-minute video chunk in isolation, i.e., without the full movie as context.
The generation process is iterative; generations for prior chunks are provided as a part of the input.

For our test set, where character names may not be inferable from the video alone, we provide manually annotated face crops of each character's first appearance alongside their name, in accordance with how a human describer would first assemble a character list (Appendix~\ref{appx:faces}).
Prompts are in Appendix~\ref{sec:prompts}.

\begin{table*}
\centering
\begin{small}
\begin{tabular}{l *{8}{S[table-format=2.1, table-column-width=1cm]}}
\toprule
& & & &
& \multicolumn{2}{c}{\textbf{Qwen 3.5}} & \multicolumn{2}{c}{\textbf{Gemini 3.1}} \\
\cmidrule(lr){6-7}\cmidrule(lr){8-9}
& {\raisebox{.2cm}[0pt]{\textbf{Expert}}} & {\raisebox{.2cm}[0pt]{\textbf{Random}}} & {\raisebox{.2cm}[0pt]{\textbf{D-AD}}} & {\raisebox{.2cm}[0pt]{\textbf{SbS}}}
& {full} & {chunk} & {full} & {chunk} \\
\midrule
\rowcolor{reframed-gray}\multicolumn{9}{c}{\textit{Dialogue-gaps}} \\ \midrule
CIDEr & \textbf{51.4} & 1.3 & 15.6 & 16.3 & 10.3 & 13.2 & 12.2 & 19.0 \\
METEOR & \textbf{20.1} & 4.4 & 7.1 & 7.0 & 6.1 & 8.1 & 6.1 & 8.2 \\
\midrule
\rowcolor{reframed-gray}\multicolumn{9}{c}{\textit{Aligned}} \\ \midrule
SODA-M & \textbf{16.3} & 4.0 & 8.3 & 8.0 & 6.8 & 7.4 & 6.8 & 7.9 \\
SODA-T & \textbf{81.0} & 27.1 & 49.0 & 53.0 & 26.0 & 38.4 & 27.1 & 36.0 \\
\midrule
\rowcolor{reframed-gray}\multicolumn{9}{c}{\textit{QA-based}} \\ \midrule
QEval & \textbf{69.6} & 34.1 & 35.7 & 39.9 & 40.0 & 43.9 & 40.1 & 42.3 \\
QEval-T & \textbf{61.2} & 2.3 & 8.8 & 17.0 & 9.4 & 17.3 & 10.6 & 16.4 \\
\bottomrule
\end{tabular}
\end{small}
\caption{Results on the \protect\reframed challenge set. \textit{Dialogue-gaps}: comparisons within each dialogue gap. \textit{Aligned}: scored textually (\mbox{SODA-M}) or temporally (\mbox{SODA-T}). \textit{QA-based}: AD-generated answers scored against reference questions; QEval-T also penalizes descriptions at the wrong time. \textit{full/chunk:} full movie vs.\ 10-minute video chunk as input.}
\label{tab:challenge-results}
\vspace{-10.5pt}
\end{table*}

\subsection{Results}

\textbf{How well do humans agree?}
Table~\ref{tab:challenge-results} summarizes results on our challenge set.
Despite the subjectivity of AD, inter-human agreement is high.
A QEval score of 69.6\% shows that professional describers agree well on which elements of a movie are narratively salient, and a QEval-T of 61.2\% confirms that they also describe them at similar times.
N-gram agreement is likewise high (CIDEr\,=\,51.4, METEOR\,=\,20.1), validating our dialogue gap-based evaluation.
SODA-M (16.3) is slightly lower than dialogue-gap METEOR, as expected given its stricter one-to-one alignment.
SODA-T is high (81.0\%) --- even when human describers formulate AD differently, their descriptions overlap enough for temporal alignment.

\textbf{Does the random baseline expose weaknesses in our metrics?}
The LLM evaluator answers 34.1\% of questions correctly from random descriptions (QEval; chance-level=20\%) demonstrating a degree of parametric knowledge and common-sense reasoning.
A baseline with an empty context achieves 41.4\% (all other metrics are zero), reinforcing this point.
Nevertheless, random performance drops to 2.3\% once temporal grounding is required (QEval-T).
SODA-T reaches 27.1\%: a result of monotonic alignment with random scores.
All other metrics are near-floor, confirming that random descriptions receive no undeserved credit from our metrics.

\textbf{How do pre-existing systems and LLMs perform?}
All differences between Shot-by-Shot (SbS) and DistinctAD (D-AD) against the random baseline are significant (for an evaluation of a wider range of systems, see Appendix Table~\ref{tab:full-challenge-results}).
The largest gains are on CIDEr (DistinctAD: 15.6 and Shot-by-Shot: 16.3, vs.\ 1.3).
SODA-T is trivially high, as expected, since these models are provided with gold-standard AD placements.
The more recent Shot-by-Shot outperforms DistinctAD on both QA-based metrics (both $p<0.001$).

The LLMs also outperform the random baseline.
However, with full-movie input, they achieve significantly lower scores across all dialogue gap and aligned metrics compared to the pre-existing models.
This reflects a difference in modeling approach: pre-existing models are provided with local context and generate plausible captions for each reference segment, whereas LLMs with full-movie input need to use timecoding to retrieve and describe the relevant part of their input.
 LLMs achieve the best results when provided with chunked input (ten minutes of context, as opposed to either a few seconds or the full movie), significantly outperforming the full-movie approach across all metrics.
These results show that current LLMs cannot yet leverage full-movie narrative context, which is a requirement for realistic AD generation.
Furthermore, AD requires precise temporal grounding, and our results show that this is currently less achievable with full-movie video input (SODA-T: Qwen 26.0\% vs. chunk 38.4\%, QEval-T: 9.4\% vs. 17.3\%).
\begin{figure}[t]
\begin{tcolorbox}[
  colback=reframed-gray!10,
  colframe=reframed-silver,
  boxrule=0.5pt,
  arc=3pt,
  left=0pt, right=0pt, top=8pt, bottom=8pt,
  width=\columnwidth
]
{\small
\vspace{-.2cm}
\textbf{[RON] He's supposed to be mad as a hatter, though, these days.}

\smallskip

{\color{reframed-gold!50!black}American AD: \textit{The professors watch warily as Mad Eye Moody hobbles past. He scans the room with both eyes, then uses the bulging eye like a telescope to zoom in on Harry. Dumbledore shakes Moody's hand.}}

{\color{reframed-gold!50!black}British AD: \textit{Moody has a scarred face and a large, protruding false eye fixed to a leather eye patch. It swivels erratically and lingers on Harry. One of Moody's feet is encased in a metal boot. Dumbledore...}}

\smallskip

{\color{reframed-red}Qwen 3.5 (full-movie input): \textit{The new professor shakes Dumbledore's hand while his multiple-lensed eye zooms in and out. He settles onto the bench next to Severus Snape, looking quite intimidating.}}

{\color{reframed-red}Gemini 3.1 (full-movie input): \textit{Filch stands alone in the aisles between the long tables, his hunched posture casting a gloomy presence over the Great Hall.}}

\smallskip

\textbf{[DUMBLEDORE] My dear old friend, thanks for coming.}

\smallskip

Who does the swiveling eye linger on?

\
\hfill
\textbf{A. Harry} \quad
B. Dumbledore \quad
C. Mr. Crouch \quad
D. Ron \quad
E. Moody

\vspace{-.2cm}
}
\end{tcolorbox}
\vspace{-8pt}
\caption{Two references and two model generated ADs based on a dialogue gap in the movie \textit{Harry Potter and the Goblet of Fire} (2005). Dialogue in \textbf{bold}, AD in \emph{italics}. The multiple-choice question is taken from our QA-based evaluation.}
\label{ex:harry}
\vspace{-9.5pt}
\end{figure}

Consider Figure~\ref{ex:harry}, which shows a twelve second dialogue gap twenty minutes into a \textit{Harry Potter} movie.
During this scene, two characters are introduced at a feast in a hall: Moody (a villain in disguise), and Dumbledore (a headmaster).
Both LLMs are able to retrieve the relevant moment from their 2-hour video input and describe it with largely faithful details about a figure in a hall (note Qwen hallucinates that he sits down, and Gemini incorrectly names the character).
However, they both fail to describe the narrative essence of the scene --- Moody's appearance and his spying on Harry Potter foreshadow his status as a villain in disguise, as described by both references.
The difference in length between the generations (21--26 words) and the references (34--35 words) is also notable.
Professional AD in our challenge set is tightly centered around 200 words per minute, close to guideline recommendations.
By contrast, Gemini generates descriptions that would need to be spoken much slower than the references (on average 110--118 words per minute; see Appendix Table~\ref{tab:wpm}).
For the same example, the pre-existing AD systems generate more text (30--47 words), but their descriptions are less coherent and faithful (DistinctAD describes Moody as wearing a suit and sunglasses, with Ron and Harry standing in a forest).

\textbf{What about the test set?}
Appendix Table~\ref{tab:test-set} gives \protect\reframed test set results.
Numbers are uniformly higher (Gemini: CIDEr 22.5, QEval-T 23.1\%; Qwen: CIDEr: 18.0, QEval-T: 21.7\%), suggesting this task is more tractable, though this gain stems from providing character names and faces: removing them drops CIDEr to 14.5 and 12.8 for Gemini and Qwen, respectively (both $p<0.001$).

\section{Conclusion and Future Work}

We introduce a new formulation of AD generation that reflects real-world requirements: a model must decide both when and what to describe. To support this task, we introduce \protect\reframed, a dataset of 2,023 movie excerpts paired with multiple AD transcripts, dialogue subtitles, SDH, and, for 85 movies, aligned screenplays, together with a fully annotated benchmark of 10 full movies. We propose evaluation metrics that account for both the temporal constraints of AD and its inherent subjectivity. Validation against expert human AD shows high agreement. Prior AD systems and zero-shot LLMs outperform a random baseline but remain far below human performance.

Our results show that current models struggle with three coupled challenges. \textbf{Content selection} requires identifying narratively salient characters, objects, actions, and visual details across long contexts. \textbf{Temporal placement} requires finding suitable gaps within dialogue or salient audio. \textbf{AD realization} requires producing descriptions that are informative yet match the available narration time.
\reframed supports future work in these directions by providing dialogue subtitles, SDH subtitles, scene boundaries, aligned screenplays, and multiple professional AD references.

\section*{Ethics Statement}

Providing equitable access to visual media requires AD to faithfully describe all key visual elements.
This includes potentially sensitive topics such as sexual content, abuse and violence.
This requirement places new pressure on the safety guardrails designed for current LLMs.
In our evaluation of Gemini, one full movie (\textit{Charlie St. Cloud (2010)}), two 10 minute chunks within the same movie, and two movie excerpts from \textit{Speak No Evil (2024)} triggered non-configurable safety guardrails.
This is likely to do with automatic detection of depictions of harm involving minors.
For full-movie evaluation of Gemini, we report results based on the subset of movies without the one flagged.
For our chunked and movie excerpt-based evaluation, we determined the output for these runs to be empty.

All AD transcripts for our evaluation data (validation, test and challenge splits) were the result of fully manual human transcription.
We paid a post-production company contracted for the task their full rate for professional feature-movie subtitling.

In terms of data release, copyright is of potential concern here.
The three main splits of our dataset contain data derived only from publicly available sources.
We release the movie video excerpts as links and release all the other text-based scene data in full.
The AD transcripts and dialogue subtitles are derived from excerpts of full-movie scripts, and our challenge set is based on full movies --- we share these after researchers agree to terms of use.
We release code necessary to evaluate future AD generation systems with the metrics proposed in this work.

Our dataset only represents English AD, since it forms the largest available data.
In our companion paper \citep{sterner-2026-comparing}, we test hypotheses concerning the differences between American and British AD (for results separated by references, see Appendix Table~\ref{tab:per-reference}).
We expect many aspects of the formulation and evaluation setup to transfer cross-lingually.
Professionally authored AD in many languages follows similar constraints regarding dialogue gaps and narrative accessibility.
On the other hand, stylistic conventions vary across regions.
Independently created AD tracks already exist in many languages for a subset of our movies, which could enable future cross-lingual extensions using our methodology.

\section*{Acknowledgements}

Sterner is funded by an EPSRC PhD studentship (project reference 2923920). Lapata gratefully acknowledges the support of the UK Engineering and Physical Sciences Research Council (grant EP/W002876/1).
We thank Arian Merati for his contributions to the evaluation of AD transcription and diarization.
We are grateful to the team at Speechmatics for providing API access to their models and Google for providing Gemini API credits.
We thank Inderjeet Mani, Argyrios Papoudakis, Olga Loginova, and three anonymous reviewers for their constructive feedback on this manuscript.

\bibliography{colm2026_conference}

@inproceedings{bain-2020-condensed,
  title         = {{Condensed Movies}: Story Based Retrieval with Contextual Embeddings},
  author        = {Bain, Max and Nagrani, Arsha and Brown, Andrew and Zisserman, Andrew},
  year          = {2020},
  month         = {November},
  booktitle     = {Proceedings of the Asian Conference on Computer Vision (ACCV)},
  pages         = {460--479},
  url           = {https://doi.org/10.1007/978-3-030-69541-5_28}
}

@inproceedings{soldan-2022-mad,
  title         = {{MAD}: A Scalable Dataset for Language Grounding in Videos From Movie Audio Descriptions},
  author        = {Soldan, Mattia and Pardo, Alejandro and Alc{\'a}zar, Juan Le{\'o}n and Caba Heilbron, Fabian and Zhao, Chen and Giancola, Silvio and Ghanem, Bernard},
  year          = {2022},
  month         = {June},
  booktitle     = {Proceedings of the Conference on Computer Vision and Pattern Recognition (CVPR)},
  pages         = {5016--5025},
  url           = {https://doi.org/10.1109/CVPR52688.2022.00497}
}

@inproceedings{han-2023-autoad_i,
  title         = {{AutoAD}: Movie Description in Context},
  author        = {Han, Tengda and Bain, Max and Nagrani, Arsha and Varol, G\"ul and Xie, Weidi and Zisserman, Andrew},
  year          = {2023},
  month         = {June},
  booktitle     = {Proceedings of the Conference on Computer Vision and Pattern Recognition (CVPR)},
  pages         = {18930--18940},
  url           = {https://doi.org/10.1109/CVPR52729.2023.01815}
}

@inproceedings{han-2023-autoad_ii,
  title         = {{AutoAD II}: The Sequel - Who, When, and What in Movie Audio Description},
  author        = {Han, Tengda and Bain, Max and Nagrani, Arsha and Varol, Gul and Xie, Weidi and Zisserman, Andrew},
  year          = {2023},
  month         = {October},
  booktitle     = {Proceedings of the International Conference on Computer Vision (ICCV)},
  pages         = {13645--13655},
  url           = {https://doi.org/10.1109/ICCV51070.2023.01255}
}

@inproceedings{han-2024-autoad_iii,
  title         = {{AutoAD III}: The Prequel - Back to the Pixels},
  author        = {Han, Tengda and Bain, Max and Nagrani, Arsha and Varol, G\"ul and Xie, Weidi and Zisserman, Andrew},
  year          = {2024},
  month         = {June},
  booktitle     = {Proceedings of the Conference on Computer Vision and Pattern Recognition (CVPR)},
  pages         = {18164--18174},
  url           = {https://doi.org/10.1109/CVPR52733.2024.01720}
}

@inproceedings{xie-2024-autoad_zero,
  title         = {{AutoAD-Zero}: A Training-Free Framework for Zero-Shot Audio Description},
  author        = {Xie, Junyu and Han, Tengda and Bain, Max and Nagrani, Arsha and Varol, G\"ul and Xie, Weidi and Zisserman, Andrew},
  year          = {2024},
  month         = {December},
  booktitle     = {Proceedings of the Asian Conference on Computer Vision (ACCV)},
  pages         = {2265--2281},
  url           = {https://doi.org/10.1007/978-981-96-0908-6_5}
}

@inproceedings{xie-2025-shot_by_shot,
  title         = {{Shot-by-Shot}: Film-Grammar-Aware Training-Free Audio Description Generation},
  author        = {Xie, Junyu and Han, Tengda and Bain, Max and Nagrani, Arsha and Khandelwal, Eshika and Varol, G\"ul and Xie, Weidi and Zisserman, Andrew},
  year          = {2025},
  month         = {October},
  booktitle     = {Proceedings of the International Conference on Computer Vision (ICCV)},
  pages         = {16503--16513},
  url           = {https://doi.org/10.1109/ICCV51701.2025.01532}
}

@inproceedings{zhang-2024-mm_narator,
  title         = {{MM-Narrator}: Narrating Long-form Videos with Multimodal In-Context Learning},
  author        = {Zhang, Chaoyi and Lin, Kevin and Yang, Zhengyuan and Wang, Jianfeng and Li, Linjie and Lin, Chung-Ching and Liu, Zicheng and Wang, Lijuan},
  year          = {2024},
  month         = {June},
  booktitle     = {Proceedings of the Conference on Computer Vision and Pattern Recognition (CVPR)},
  pages         = {13647--13657},
  url           = {https://doi.org/10.1109/CVPR52733.2024.01295}
}

@misc{chu-2024-llm_ad,
  title         = {{LLM-AD}: Large Language Model based Audio Description System},
  author        = {Peng Chu and Jiang Wang and Andre Abrantes},
  year          = {2024},
  url           = {https://doi.org/10.48550/arXiv.2405.00983},
  eprint        = {2405.00983v1},
  archiveprefix = {arXiv},
  primaryclass  = {cs.CV}
}

@inproceedings{ye-2024-mmad,
  title         = "{MMAD}: Multi-modal Movie Audio Description",
  author        = "Ye, Xiaojun and Chen, Junhao and Li, Xiang and Xin, Haidong and Li, Chao and Zhou, Sheng and Bu, Jiajun",
  year          = "2024",
  month         = may,
  booktitle     = {Proceedings of the Joint International Conference on Computational Linguistics, Language Resources and Evaluation (LREC-COLING)},
  publisher     = "ELRA and ICCL",
  address       = "Torino, Italia",
  pages         = "11415--11428",
  url           = "https://aclanthology.org/2024.lrec-main.998",
  editor        = "Calzolari, Nicoletta and Kan, Min-Yen and Hoste, Veronique and Lenci, Alessandro and Sakti, Sakriani and Xue, Nianwen"
}

@inproceedings{gao-2025-ad_review,
  title         = "Audio Description Generation in the Era of {LLM}s and {VLM}s: A Review of Transferable Generative {AI} Technologies",
  author        = "Gao, Yingqiang and Fischer, Lukas and Lintner, Alexa and Ebling, Sarah",
  year          = "2025",
  month         = apr,
  booktitle     = "Findings of the Association for Computational Linguistics: {NAACL}",
  publisher     = "Association for Computational Linguistics",
  address       = "Albuquerque, New Mexico",
  pages         = "471--490",
  isbn          = "979-8-89176-195-7",
  url           = {https://doi.org/10.18653/v1/2025.findings-naacl.29},
  editor        = "Chiruzzo, Luis and Ritter, Alan and Wang, Lu"
}

@inproceedings{fang-2024-distinct_ad,
  title         = {{DistinctAD}: Distinctive Audio Description Generation in Contexts},
  author        = {Fang, Bo and Wu, Wenhao and Wu, Qiangqiang and Song, Yuxin and Chan, Antoni B.},
  year          = {2025},
  month         = {June},
  booktitle     = {Proceedings of the Computer Vision and Pattern Recognition Conference (CVPR)},
  pages         = {13571--13581},
  url           = {https://doi.org/10.1109/CVPR52734.2025.01267}
}

@inproceedings{lin-2024-movie_seq,
  title         = {Learning video context as interleaved multimodal sequences},
  author        = {Lin, Kevin Qinghong and Zhang, Pengchuan and Gao, Difei and Xia, Xide and Chen, Joya and Gao, Ziteng and Xie, Jinheng and Xiao, Xuhong and Shou, Mike Zheng},
  year          = {2024},
  booktitle     = "Proceedings of the European Conference on Computer Vision (ECCV)",
  pages         = {375--396},
  url           = {https://doi.org/10.1007/978-3-031-72967-6_21}
}

@inproceedings{yue-2023-movie101,
  title         = "Movie101: A New Movie Understanding Benchmark",
  author        = "Yue, Zihao and Zhang, Qi and Hu, Anwen and Zhang, Liang and Wang, Ziheng and Jin, Qin",
  year          = "2023",
  month         = jul,
  booktitle     = "Proceedings of the Association for Computational Linguistics (ACL)",
  publisher     = "Association for Computational Linguistics",
  address       = "Toronto, Canada",
  pages         = "4669--4684",
  url           = {https://doi.org/10.18653/v1/2023.acl-long.257},
  editor        = "Rogers, Anna and Boyd-Graber, Jordan and Okazaki, Naoaki"
}

@inproceedings{yue-2024-movie101_v2,
  title         = "Movie101v2: Improved Movie Narration Benchmark",
  author        = "Yue, Zihao and Zhang, Yepeng and Wang, Ziheng and Jin, Qin",
  year          = "2025",
  month         = jul,
  booktitle     = "Proceedings of the Association for Computational Linguistics (ACL)",
  publisher     = "Association for Computational Linguistics",
  address       = "Vienna, Austria",
  pages         = "17081--17095",
  isbn          = "979-8-89176-251-0",
  url           = {https://doi.org/10.18653/v1/2025.acl-long.836},
  editor        = "Che, Wanxiang and Nabende, Joyce and Shutova, Ekaterina and Pilehvar, Mohammad Taher"
}

@inproceedings{rohrbach-2015-mpii_md,
  title         = {A Dataset for Movie Description},
  author        = {Rohrbach, Anna and Rohrbach, Marcus and Tandon, Niket and Schiele, Bernt},
  year          = {2015},
  month         = {June},
  booktitle     = {Proceedings of the Conference on Computer Vision and Pattern Recognition (CVPR)},
  url           = {https://doi.org/10.1109/CVPR.2015.7298940}
}

@article{rohrbach-2017-lsmdc,
  title         = {Movie description},
  author        = {Rohrbach, Anna and Torabi, Atousa and Rohrbach, Marcus and Tandon, Niket and Pal, Christopher and Larochelle, Hugo and Courville, Aaron and Schiele, Bernt},
  year          = {2017},
  journal       = {International Journal of Computer Vision (IJCV)},
  publisher     = {Springer},
  volume        = {123},
  pages         = {94--120},
  url           = {https://doi.org/10.1007/s11263-016-0987-1}
}

@article{pitcher_cooper-2024-you_describe,
  title         = {{You Described, We Archived}: A Rich Audio Description Dataset},
  author        = {Pitcher-Cooper, Charity and Seth, Manali and Kao, Benjamin and Coughlan, James M. and Yoon, Ilmi},
  year          = {2024},
  month         = {Jan},
  journal       = {Journal on Technology and Persons with Disabilities},
  volume        = {11},
  pages         = {192--208},
  url           = {https://pubmed.ncbi.nlm.nih.gov/38516032/}
}

@inproceedings{lee-2025-ssms_ad,
  title         = {Now you see me: Context-aware automatic audio description},
  author        = {Lee, Seon-Ho and Wang, Jue and Fan, David and Zhang, Zhikang and Liu, Linda and Hao, Xiang and Bhat, Vimal and Li, Xinyu (Arthur)},
  year          = {2025},
  month         = {feb},
  booktitle     = {Proceedings of the Winter Conference on Applications of Computer Vision (WACV)},
  url           = {https://doi.org/10.1109/WACV61041.2025.00540}
}

@inproceedings{bain-2023-whisperx,
  title         = {{WhisperX}: Time-Accurate Speech Transcription of Long-Form Audio},
  author        = {Max Bain and Jaesung Huh and Tengda Han and Andrew Zisserman},
  year          = {2023},
  booktitle     = {Interspeech},
  pages         = {4489--4493},
  issn          = {2958-1796},
  url           = {https://doi.org/10.21437/Interspeech.2023-78}
}

@inproceedings{radford-2023-whisper,
  title         = {Robust speech recognition via large-scale weak supervision},
  author        = {Radford, Alec and Kim, Jong Wook and Xu, Tao and Brockman, Greg and McLeavey, Christine and Sutskever, Ilya},
  year          = {2023},
  booktitle     = {Proceedings of the International Conference on Machine Learning (ICML)},
  location      = {Honolulu, Hawaii, USA},
  url           = {https://doi.org/10.48550/arXiv.2212.04356},
  articleno     = {1182},
  numpages      = {27}
}

@article{lakritz-2006-semi,
  title         = {The semi-automatic generation of audio description from screenplays},
  author        = {Lakritz, James and Salway, Andrew},
  year          = {2006},
  journal       = {Dept. of Computing Technical Report CS-06-05, University of Surrey},
  publisher     = {Citeseer},
  url           = {https://andrewsalway.work/wp-content/uploads/2020/02/cs-06-05-1.pdf}
}

@article{campos-2023-machine,
  title         = {Machine generation of audio description for blind and visually impaired people},
  author        = {Campos, Virg{\'\i}nia P and Gon{\c{c}}alves, Luiz MG and Ribeiro, Wesnydy L and Ara{\'u}jo, Tiago MU and Do Rego, Tha{\'\i}s G and Figueiredo, Pedro HV and Vieira, Suanny FS and Costa, Thiago FS and Moraes, Caio C and Cruz, Alexandre CS and others},
  year          = {2023},
  journal       = {ACM Transactions on Accessible Computing},
  publisher     = {ACM New York, NY},
  volume        = {16},
  number        = {2},
  pages         = {1--28},
  url           = {https://doi.org/10.1145/3590955}
}

@inproceedings{rimle-2020-enriching,
  title         = {Enriching video captions with contextual text},
  author        = {Rimle, Philipp and Dogan-Sch{\"o}nberger, Pelin and Gross, Markus},
  year          = {2020},
  booktitle     = {Proceedings of the International Conference on Pattern Recognition (ICPR)},
  pages         = {5474--5481},
  url           = {https://doi.org/10.1109/ICPR48806.2021.9412008}
}

@inproceedings{oncescu-2021-quer_yd,
  title         = {{QuerYD}: A video dataset with high-quality text and audio narrations},
  author        = {Oncescu, Andreea-Maria and Henriques, Joao F and Liu, Yang and Zisserman, Andrew and Albanie, Samuel},
  year          = {2021},
  booktitle     = {Proceedings of the International Conference on Acoustics, Speech and Signal Processing (ICASSP)},
  pages         = {2265--2269},
  url           = {https://doi.org/10.1109/icassp39728.2021.9414640}
}

@misc{lin-2023-mm_vid,
  title         = {{MM-VID}: Advancing Video Understanding with {GPT-4V}(ision)},
  author        = {Kevin Lin and Faisal Ahmed and Linjie Li and Chung-Ching Lin and Ehsan Azarnasab and Zhengyuan Yang and Jianfeng Wang and Lin Liang and Zicheng Liu and Yumao Lu and Ce Liu and Lijuan Wang},
  year          = {2023},
  url           = {https://doi.org/10.48550/arXiv.2310.19773},
  eprint        = {2310.19773v1},
  archiveprefix = {arXiv},
  primaryclass  = {cs.CV}
}

@article{damen-2022-epic_kitchens_100,
  title         = {Rescaling Egocentric Vision: Collection, Pipeline and Challenges for {EPIC-KITCHENS-100}},
  author        = {Damen, Dima and Doughty, Hazel and Farinella, Giovanni Maria and Furnari, Antonino and Kazakos, Evangelos and Ma, Jian and Moltisanti, Davide and Munro, Jonathan and Perrett, Toby and Price, Will and Wray, Michael},
  year          = {2022},
  journal       = {International Journal of Computer Vision (IJCV)},
  volume        = {130},
  pages         = {33--55},
  url           = {https://doi.org/10.1007/s11263-021-01531-2}
}

@inproceedings{miech19howto100m,
  title         = {How{T}o100{M}: Learning a Text-Video Embedding by Watching Hundred Million Narrated Video Clips},
  author        = {Miech, Antoine and Zhukov, Dimitri and Alayrac, Jean-Baptiste and Tapaswi, Makarand and Laptev, Ivan and Sivic, Josef},
  year          = {2019},
  booktitle     = {Proceedings of the International Conference on Computer Vision (ICCV)},
  url           = {https://doi.org/10.1109/ICCV.2019.00272}
}

@inproceedings{dogan-2018-yms,
  title         = {A Neural Multi-Sequence Alignment TeCHnique ({NeuMATCH})},
  author        = {Dogan, Pelin and Li, Boyang and Sigal, Leonid and Gross, Markus},
  year          = {2018},
  month         = {June},
  booktitle     = {Proceedings of the Conference on Computer Vision and Pattern Recognition (CVPR)},
  pages         = {8749--8758},
  url           = {https://doi.org/10.1109/CVPR.2018.00912}
}

@inproceedings{sun-2023-symon,
  title         = {Synopses of Movie Narratives: a Video-Language Dataset for Story Understanding},
  author        = {Sun, Yidan and Chao, Qin and Ji, Yangfeng and Li, Boyang},
  year          = {2025},
  booktitle     = {Proceedings of the International Conference on Multimedia and Expo (ICME)},
  pages         = {1--6},
  url           = {https://doi.org/10.1109/ICME59968.2025.11209961}
}

@article{lu-2024-cvsv,
  title         = {Show Me a Video: A Large-Scale Narrated Video Dataset for Coherent Story Illustration},
  author        = {Lu, Yu and Ni, Feiyue and Wang, Haofan and Guo, Xiaofeng and Zhu, Linchao and Yang, Zongxin and Song, Ruihua and Cheng, Lele and Yang, Yi},
  year          = {2024},
  journal       = {IEEE Transactions on Multimedia},
  volume        = {26},
  pages         = {2456--2466},
  url           = {https://doi.org/10.1109/TMM.2023.3296944}
}

@inproceedings{sun-2024-m_symon,
  title         = "Multilingual Synopses of Movie Narratives: A Dataset for Vision-Language Story Understanding",
  author        = "Sun, Yidan and Yu, Jianfei and Li, Boyang",
  year          = "2024",
  month         = nov,
  booktitle     = "Findings of the Association for Computational Linguistics: {EMNLP}",
  publisher     = "Association for Computational Linguistics",
  address       = "Miami, Florida, USA",
  pages         = "13488--13504",
  url           = {https://doi.org/10.18653/v1/2024.findings-emnlp.788},
  editor        = "Al-Onaizan, Yaser and Bansal, Mohit and Chen, Yun-Nung"
}

@inproceedings{everingham-2006-hello,
  title         = {``{H}ello! {M}y name is... {B}uffy''--Automatic Naming of Characters in {TV} Video},
  author        = {Everingham, Mark and Sivic, Josef and Zisserman, Andrew},
  year          = {2006},
  booktitle     = {Proceedings of the British Machine Vision Conference ({BMVC})},
  volume        = {2},
  url           = {http://doi.org/10.5244/C.20.92},
  organization  = {Citeseer}
}

@inproceedings{lambert-2013-reordering,
  title         = {Scene reordering in movie script alignment},
  author        = {Lambert, Anne and Gu\'{e}gan, Marie and Zhou, Kai},
  year          = {2013},
  booktitle     = {Proceedings of the International Workshop on Content-Based Multimedia Indexing (CBMI)},
  pages         = {213--218},
  url           = {https://doi.org/10.1109/CBMI.2013.6576585}
}

@inproceedings{lin-2004-rouge,
  title         = {{ROUGE}: A Package for Automatic Evaluation of Summaries},
  author        = {Chin-Yew Lin},
  year          = {2004},
  booktitle     = {Text Summarization Branches Out},
  publisher     = {Association for Computational Linguistics},
  pages         = {74--81},
  url           = {https://aclanthology.org/W04-1013/}
}

@inproceedings{vedantam-2015-cider,
  title         = {{CIDEr}: Consensus-based Image Description Evaluation},
  author        = {Ramakrishna Vedantam and C. Lawrence Zitnick and Devi Parikh},
  year          = {2015},
  booktitle     = {Proceedings of the Conference on Computer Vision and Pattern Recognition (CVPR)},
  pages         = {4566--4575},
  url           = {https://doi.org/10.1109/CVPR.2015.7299087}
}

@inproceedings{anderson-2016-spice,
  title         = {{SPICE}: Semantic Propositional Image Caption Evaluation},
  author        = {Peter Anderson and Basura Fernando and Mark Johnson and Stephen Gould},
  year          = {2016},
  booktitle     = {Proceedings of the European Conference on Computer Vision (ECCV)},
  publisher     = {Springer},
  pages         = {382--398},
  url           = {https://doi.org/10.1007/978-3-319-46454-1_24}
}

@inproceedings{zhang-2019-bertscore,
  title         = {{BERTScore}: Evaluating Text Generation with {BERT}},
  author        = {Tianyi Zhang and Varsha Kishore and Felix Wu and Kilian Q. Weinberger and Yoav Artzi},
  year          = {2020},
  booktitle     = {Proceedings of the International Conference on Learning Representations (ICLR)},
  url           = {https://doi.org/10.48550/arXiv.1904.09675}
}

@inproceedings{wang-2025-uni_ad,
  title         = {Contextual AD Narration with Interleaved Multimodal Sequence},
  author        = {Wang, Hanlin and Tong, Zhan and Zheng, Kecheng and Shen, Yujun and Wang, Limin},
  year          = {2025},
  month         = {June},
  booktitle     = {Proceedings of the Computer Vision and Pattern Recognition Conference (CVPR)},
  pages         = {8372--8383},
  url           = {https://doi.org/10.1109/CVPR52734.2025.00784}
}

@inproceedings{papineni-2002-bleu,
  title         = "{BLEU}: a Method for Automatic Evaluation of Machine Translation",
  author        = "Papineni, Kishore and Roukos, Salim and Ward, Todd and Zhu, Wei-Jing",
  year          = "2002",
  month         = jul,
  booktitle     = {Proceedings of the Association for Computational Linguistics (ACL)},
  publisher     = "Association for Computational Linguistics",
  address       = "Philadelphia, Pennsylvania, USA",
  pages         = "311--318",
  url           = {https://doi.org/10.3115/1073083.1073135},
  editor        = "Isabelle, Pierre and Charniak, Eugene and Lin, Dekang"
}

@inproceedings{saxena-2024-moviesum,
  title         = "{M}ovie{S}um: An Abstractive Summarization Dataset for Movie Screenplays",
  author        = "Saxena, Rohit and Keller, Frank",
  year          = "2024",
  month         = aug,
  booktitle     = {Findings of the Association for Computational Linguistics: ACL},
  publisher     = "Association for Computational Linguistics",
  address       = "Bangkok, Thailand",
  pages         = "4043--4050",
  url           = {https://doi.org/10.18653/v1/2024.findings-acl.239},
  editor        = "Ku, Lun-Wei and Martins, Andre and Srikumar, Vivek"
}

@inproceedings{kala-2025-what,
  title         = "What You See is What You Ask: Evaluating Audio Descriptions",
  author        = "Kala, Divy and Khandelwal, Eshika and Tapaswi, Makarand",
  year          = "2025",
  month         = nov,
  booktitle     = {Proceedings of the Conference on Empirical Methods in Natural Language Processing (EMNLP)},
  publisher     = "Association for Computational Linguistics",
  address       = "Suzhou, China",
  pages         = "23496--23518",
  isbn          = "979-8-89176-332-6",
  url           = {https://doi.org/10.18653/v1/2025.emnlp-main.1199}
}

@inproceedings{zheng-2023-llm_as_a_judge,
  title         = {Judging {LLM}-as-a-judge with {MT}-bench and {C}hatbot {A}rena},
  author        = {Zheng, Lianmin and Chiang, Wei-Lin and Sheng, Ying and Zhuang, Siyuan and Wu, Zhanghao and Zhuang, Yonghao and Lin, Zi and Li, Zhuohan and Li, Dacheng and Xing, Eric P. and Zhang, Hao and Gonzalez, Joseph E. and Stoica, Ion},
  year          = {2023},
  booktitle     = {Proceedings of the International Conference on Neural Information Processing Systems (NeurIPS)},
  location      = {New Orleans, LA, USA},
  publisher     = {Curran Associates Inc.},
  address       = {Red Hook, NY, USA},
  url           = {https://dl.acm.org/doi/10.5555/3666122.3668142},
  articleno     = {2020},
  numpages      = {29}
}

@inproceedings{banerjee-2005-meteor,
  title         = "{METEOR}: An Automatic Metric for {MT} Evaluation with Improved Correlation with Human Judgments",
  author        = "Banerjee, Satanjeev and Lavie, Alon",
  year          = "2005",
  month         = jun,
  booktitle     = "Proceedings of the Workshop on Intrinsic and Extrinsic Evaluation Measures for Machine Translation and/or Summarization",
  publisher     = "Association for Computational Linguistics",
  address       = "Ann Arbor, Michigan",
  pages         = "65--72",
  url           = "https://aclanthology.org/W05-0909/",
  editor        = "Goldstein, Jade and Lavie, Alon and Lin, Chin-Yew and Voss, Clare"
}

@inproceedings{fujita-2020-soda,
  title         = {SODA: Story Oriented Dense Video Captioning Evaluation Framework},
  author        = {Fujita, Soichiro and Hirao, Tsutomu and Kamigaito, Hidetaka and Okumura, Manabu and Nagata, Masaaki},
  year          = {2020},
  booktitle     = "Proceedings of the European Conference on Computer Vision (ECCV)",
  pages         = {517--531},
  url           = {https://doi.org/10.1007/978-3-030-58539-6_31},
  numpages      = {15}
}

@inproceedings{batra-2022-temporal,
  title         = {A Closer Look at Temporal Ordering in the Segmentation of Instructional Videos},
  author        = {Batra, Anil and Gowda, Shreyank N and Keller, Frank and Sevilla-Lara, Laura},
  year          = {2022},
  booktitle     = {Proceedings of the British Machine Vision Conference (BMVC)},
  url           = {https://doi.org/10.48550/arXiv.2209.15501}
}

@inproceedings{venugopalan-2025-metrics,
  title         = {Metrics for Fine-Grained Evaluation of Inline Audio Descriptions},
  author        = {Subhashini Venugopalan and Amy Pavel and Tyler Roper and Jimmy Tobin and Emily Wilson and Jenny Tan and Alicia Martin and Anton Kast},
  year          = {2025},
  booktitle     = {Proceedings of the Workshop on Vision Foundation Models and Generative AI for Accessibility: Challenges and Opportunities},
  url           = {https://openreview.net/forum?id=HEzz8aNSW0}
}

@inproceedings{deganutti-2025-dante,
  title         = {{DANTE-AD}: Dual-Vision Attention Network for Long-Term Audio Description},
  author        = {Deganutti, Adrienne and Hadfield, Simon and Gilbert, Andrew},
  year          = {2025},
  booktitle     = {Proceedings of the Workshop on AI for Content Creation (AI4CC)},
  url           = {https://doi.org/10.48550/arXiv.2503.24096}
}

@inproceedings{park-2025-narr_ad,
  title         = {{NarrAD}: Automatic Generation of Audio Descriptions for Movies with Rich Narrative Context},
  author        = {Park, Jaehyeong and Ye, Juncheol and Lee, Seungkook and Ka, Hyun W. and Han, Dongsu},
  year          = {2025},
  month         = {February},
  booktitle     = {Proceedings of the Winter Conference on Applications of Computer Vision (WACV)},
  pages         = {409--419},
  url           = {https://doi.org/10.1109/WACV61041.2025.00050}
}

@inproceedings{tapaswi-2016-movieqa,
  title         = {Movie{QA}: Understanding stories in movies through question-answering},
  author        = {Tapaswi, Makarand and Zhu, Yukun and Stiefelhagen, Rainer and Torralba, Antonio and Urtasun, Raquel and Fidler, Sanja},
  year          = {2016},
  booktitle     = {Proceedings of the Conference on Computer Vision and Pattern Recognition (CVPR)},
  pages         = {4631--4640},
  url           = {https://doi.org/10.1109/CVPR.2016.501}
}

@inproceedings{huang-2020-movienet,
  title         = {Movie{N}et: A holistic dataset for movie understanding},
  author        = {Huang, Qingqiu and Xiong, Yu and Rao, Anyi and Wang, Jiaze and Lin, Dahua},
  year          = {2020},
  booktitle     = {Proceedings of the European Conference on Computer Vision (ECCV)},
  pages         = {709--727},
  url           = {https://doi.org/10.1007/978-3-030-58548-8_41}
}

@inproceedings{gu-2018-ava,
  title         = {{AVA}: A video dataset of spatio-temporally localized atomic visual actions},
  author        = {Gu, Chunhui and Sun, Chen and Ross, David A and Vondrick, Carl and Pantofaru, Caroline and Li, Yeqing and Vijayanarasimhan, Sudheendra and Toderici, George and Ricco, Susanna and Sukthankar, Rahul and others},
  year          = {2018},
  booktitle     = {Proceedings of the Conference on Computer Vision and Pattern Recognition (CVPR)},
  pages         = {6047--6056},
  url           = {https://doi.org/10.1109/CVPR.2018.00633}
}

@inproceedings{vicol-2018-moviegraphs,
  title         = {Movie{G}raphs: Towards understanding human-centric situations from videos},
  author        = {Vicol, Paul and Tapaswi, Makarand and Castrejon, Lluis and Fidler, Sanja},
  year          = {2018},
  booktitle     = {Proceedings of the Conference on Computer Vision and Pattern Recognition (CVPR)},
  pages         = {8581--8590},
  url           = {https://doi.org/10.1109/CVPR.2018.00895}
}

@misc{zaranis-2025-mf2,
  title         = {Movie Facts and Fibs ({MF}$^2$): A Benchmark for Long Movie Understanding},
  author        = {Emmanouil Zaranis and Ant\'{o}nio Farinhas and Saul Santos and Beatriz Canaverde and Miguel Moura Ramos and Aditya K Surikuchi and Andr\'{e} Viveiros and Baohao Liao and Elena Bueno-Benito and Nithin Sivakumaran and Pavlo Vasylenko and Shoubin Yu and Sonal Sannigrahi and Wafaa Mohammed and Ben Peters and Danae S\'{a}nchez Villegas and Elias Stengel-Eskin and Giuseppe Attanasio and Jaehong Yoon and Stella Frank and Alessandro Suglia and Chrysoula Zerva and Desmond Elliott and Mariella Dimiccoli and Mohit Bansal and Oswald Lanz and Raffaella Bernardi and Raquel Fern\'{a}ndez and Sandro Pezzelle and Vlad Niculae and Andr\'{e} F. T. Martins},
  year          = {2025},
  url           = {https://doi.org/10.48550/arXiv.2506.06275},
  eprint        = {2506.06275v1},
  archiveprefix = {arXiv},
  primaryclass  = {cs.CV}
}

@inproceedings{wu-2024-longvideobench,
  title         = {{LongVideoBench}: A Benchmark for Long-context Interleaved Video-Language Understanding},
  author        = {Haoning Wu and Dongxu Li and Bei Chen and Junnan Li},
  year          = {2024},
  booktitle     = {Proceedings of the International Conference on Neural Information Processing Systems (NeurIPS)},
  url           = {https://dl.acm.org/doi/10.5555/3737916.3738823}
}

@inproceedings{wu-2021-towards,
  title         = {Towards long-form video understanding},
  author        = {Wu, Chao-Yuan and Krahenbuhl, Philipp},
  year          = {2021},
  booktitle     = {Proceedings of the IEEE/CVF conference on computer vision and pattern recognition},
  pages         = {1884--1894},
  url           = {https://doi.org/10.1109/CVPR46437.2021.00192}
}

@inproceedings{winer-2017-automated,
  title         = {Automated screenplay annotation for extracting storytelling knowledge},
  author        = {Winer, David and Young, R},
  year          = {2017},
  booktitle     = {Proceedings of the Conference on Artificial Intelligence and Interactive Digital Entertainment (AIIDE)},
  volume        = {13},
  pages         = {273--280},
  url           = {https://doi.org/10.1609/aiide.v13i2.12994}
}

@inproceedings{agarwal-2014-parsing,
  title         = "Parsing Screenplays for Extracting Social Networks from Movies",
  author        = "Agarwal, Apoorv and Balasubramanian, Sriramkumar and Zheng, Jiehan and Dash, Sarthak",
  year          = "2014",
  month         = apr,
  booktitle     = "Proceedings of the Workshop on Computational Linguistics for Literature ({CLFL})",
  publisher     = "Association for Computational Linguistics",
  address       = "Gothenburg, Sweden",
  pages         = "50--58",
  url           = {https://doi.org/10.3115/v1/W14-0907},
  editor        = "Feldman, Anna and Kazantseva, Anna and Szpakowicz, Stan"
}

@inproceedings{turetsky-2004-screenplay,
  title         = {Screenplay alignment for closed-system speaker identification and analysis of feature films},
  author        = {Turetsky, Robert and Dimitrova, Nevenka},
  year          = {2004},
  booktitle     = {Proceedings of the International Conference on Multimedia and Expo (ICME)},
  volume        = {3},
  pages         = {1659--1662},
  url           = {https://doi.org/10.1109/ICME.2004.1394570},
  organization  = {IEEE}
}

@techreport{gil-2011-extraction,
  title         = {Extraction and Analysis of Character Interaction Networks from Plays and Movies},
  author        = {Gil, Sebastian and Kuenzel, Laney and Suen, Caroline},
  year          = {2011},
  url           = {https://snap.stanford.edu/class/cs224w-2011/proj/laneyk_Finalwriteup_v1.pdf},
  institution   = {Stanford University},
  type          = {Technical Report}
}

@inproceedings{thompson-2019-vecalign,
  title         = "{V}ecalign: Improved Sentence Alignment in Linear Time and Space",
  author        = "Thompson, Brian and Koehn, Philipp",
  year          = "2019",
  month         = nov,
  booktitle     = {Proceedings of the Conference on Empirical Methods in Natural Language Processing and the 9th International Joint Conference on Natural Language Processing (EMNLP-IJCNLP)},
  publisher     = "Association for Computational Linguistics",
  address       = "Hong Kong, China",
  pages         = "1342--1348",
  url           = {https://doi.org/10.18653/v1/D19-1136},
  editor        = "Inui, Kentaro and Jiang, Jing and Ng, Vincent and Wan, Xiaojun"
}

@misc{liu-2019-roberta,
  title         = {{RoBERTa}: A Robustly Optimized {BERT} Pretraining Approach},
  author        = {Yinhan Liu and Myle Ott and Naman Goyal and Jingfei Du and Mandar Joshi and Danqi Chen and Omer Levy and Mike Lewis and Luke Zettlemoyer and Veselin Stoyanov},
  year          = {2019},
  url           = {https://doi.org/10.48550/arXiv.1907.11692},
  eprint        = {1907.11692v1},
  archiveprefix = {arXiv},
  primaryclass  = {cs.CL}
}

@inproceedings{frohmann-2024-segment,
  title         = "{Segment Any Text}: A Universal Approach for Robust, Efficient and Adaptable Sentence Segmentation",
  author        = "Frohmann, Markus and Sterner, Igor and Vuli{\'c}, Ivan and Minixhofer, Benjamin and Schedl, Markus",
  year          = "2024",
  month         = nov,
  booktitle     = {Proceedings of the Conference on Empirical Methods in Natural Language Processing (EMNLP)},
  publisher     = "Association for Computational Linguistics",
  address       = "Miami, Florida, USA",
  pages         = "11908--11941",
  url           = {https://doi.org/10.18653/v1/2024.emnlp-main.665},
  editor        = "Al-Onaizan, Yaser and Bansal, Mohit and Chen, Yun-Nung"
}

@misc{qwen3.5,
  title         = {{Qwen3.5}: Towards Native Multimodal Agents},
  author        = {{Qwen Team}},
  year          = {2026},
  month         = {February},
  url           = {https://qwen.ai/blog?id=qwen3.5}
}

@inproceedings{peng-2024-yarn,
  title         = {Ya{RN}: Efficient Context Window Extension of Large Language Models},
  author        = {Bowen Peng and Jeffrey Quesnelle and Honglu Fan and Enrico Shippole},
  year          = {2024},
  booktitle     = {Proceedings of the International Conference on Learning Representations (ICLR)},
  url           = {https://doi.org/10.48550/arXiv.2309.00071}
}

@misc{gemini3.1flashlite,
  title         = {{Gemini 3.1 Flash-Lite}: Built for Intelligence at Scale},
  author        = {{Gemini Team}},
  year          = {2026},
  month         = {March},
  url           = {https://blog.google/innovation-and-ai/models-and-research/gemini-models/gemini-3-1-flash-lite/}
}

@inproceedings{liu-2024-infinigram,
  title         = {Infini-gram: Scaling Unbounded n-gram Language Models to a Trillion Tokens},
  author        = {Jiacheng Liu and Sewon Min and Luke Zettlemoyer and Yejin Choi and Hannaneh Hajishirzi},
  year          = {2024},
  booktitle     = {Proceedings of the Conference on Language Modeling (COLM)},
  url           = {https://doi.org/10.48550/arXiv.2401.17377}
}

@inproceedings{brown-2020-few,
  title         = {Language Models are Few-Shot Learners},
  author        = {Brown, Tom and Mann, Benjamin and Ryder, Nick and Subbiah, Melanie and Kaplan, Jared D and Dhariwal, Prafulla and Neelakantan, Arvind and Shyam, Pranav and Sastry, Girish and Askell, Amanda and Agarwal, Sandhini and Herbert-Voss, Ariel and Krueger, Gretchen and Henighan, Tom and Child, Rewon and Ramesh, Aditya and Ziegler, Daniel and Wu, Jeffrey and Winter, Clemens and Hesse, Chris and Chen, Mark and Sigler, Eric and Litwin, Mateusz and Gray, Scott and Chess, Benjamin and Clark, Jack and Berner, Christopher and McCandlish, Sam and Radford, Alec and Sutskever, Ilya and Amodei, Dario},
  year          = {2020},
  booktitle     = {Proceedings of the Conference on Neural Information Processing Systems ({NeurIPS})},
  publisher     = {Curran Associates, Inc.},
  volume        = {33},
  pages         = {1877--1901},
  url           = {https://dl.acm.org/doi/abs/10.5555/3495724.3495883},
  editor        = {H. Larochelle and M. Ranzato and R. Hadsell and M.F. Balcan and H. Lin}
}

@inproceedings{papalampidi-2021-movie,
  title         = {Movie Summarization via Sparse Graph Construction},
  author        = {Papalampidi, Pinelopi and Keller, Frank and Lapata, Mirella},
  year          = {2021},
  month         = {May},
  booktitle     = {Proceedings of the {AAAI} Conference on Artificial Intelligence},
  volume        = {35},
  pages         = {13631--13639},
  url           = {https://doi.org/10.1609/aaai.v35i15.17607}
}

@inproceedings{bouritsas-2018-multimodal,
  title         = {Multimodal visual concept learning with weakly supervised techniques},
  author        = {Bouritsas, Giorgos and Koutras, Petros and Zlatintsi, Athanasia and Maragos, Petros},
  year          = {2018},
  booktitle     = {Proceedings of the Conference on Computer Vision and Pattern Recognition (CVPR)},
  pages         = {4914--4923},
  url           = {https://doi.org/10.1109/CVPR.2018.00516}
}

@inproceedings{sivic-2009-who,
  title         = {``{W}ho are you?''-Learning person specific classifiers from video},
  author        = {Sivic, Josef and Everingham, Mark and Zisserman, Andrew},
  year          = {2009},
  booktitle     = {Proceedings of the Conference on Computer Vision and Pattern Recognition (CVPR)},
  pages         = {1145--1152},
  url           = {https://doi.org/10.1109/CVPR.2009.5206513}
}

@misc{zhang-2025-qwen3embedding,
  title         = {Qwen3 {E}mbedding: Advancing Text Embedding and Reranking Through Foundation Models},
  author        = {Yanzhao Zhang and Mingxin Li and Dingkun Long and Xin Zhang and Huan Lin and Baosong Yang and Pengjun Xie and An Yang and Dayiheng Liu and Junyang Lin and Fei Huang and Jingren Zhou},
  year          = {2025},
  url           = {https://doi.org/10.48550/arXiv.2506.05176},
  eprint        = {2506.05176v3},
  archiveprefix = {arXiv},
  primaryclass  = {cs.CL}
}

@inproceedings{wang-2021-toward,
  title         = {Toward Automatic Audio Description Generation for Accessible Videos},
  author        = {Wang, Yujia and Liang, Wei and Huang, Haikun and Zhang, Yongqi and Li, Dingzeyu and Yu, Lap-Fai},
  year          = {2021},
  booktitle     = {Proceedings of the CHI Conference on Human Factors in Computing Systems},
  location      = {Yokohama, Japan},
  publisher     = {Association for Computing Machinery},
  address       = {New York, NY, USA},
  series        = {CHI '21},
  isbn          = {9781450380966},
  url           = {https://doi.org/10.1145/3411764.3445347},
  articleno     = {277},
  numpages      = {12}
}

@misc{khandelwal-2025-coherent,
  title         = {More than a Moment: Towards Coherent Sequences of Audio Descriptions},
  author        = {Eshika Khandelwal and Junyu Xie and Tengda Han and Max Bain and Arsha Nagrani and Andrew Zisserman and G\"{u}l Varol and Makarand Tapaswi},
  year          = {2025},
  url           = {https://doi.org/10.48550/arXiv.2510.25440},
  eprint        = {2510.25440v1},
  archiveprefix = {arXiv},
  primaryclass  = {cs.CV}
}

@misc{ye-2025-focusedad,
  title         = {Focused{AD}: Character-centric Movie Audio Description},
  author        = {Xiaojun Ye and Chun Wang and Yiren Song and Sheng Zhou and Liangcheng Li and Jiajun Bu},
  year          = {2025},
  url           = {https://doi.org/10.48550/arXiv.2504.12157},
  eprint        = {2504.12157v3},
  archiveprefix = {arXiv},
  primaryclass  = {cs.CV}
}

@inproceedings{sterner-2026-comparing,
  title         = {Comparing {B}ritish and {A}merican Audio Description of Movies},
  author        = {Sterner, Igor and Lascarides, Alex and Keller, Frank},
  year          = {2026},
  month         = {June},
  booktitle     = {Proceedings of the International Workshop on Computational Models of Narrative (CMN)},
  url           = {https://homepages.inf.ed.ac.uk/alex/papers/cmn2026.pdf}
}

@article{bamman-2024-film,
  title         = {Measuring diversity in Hollywood through the large-scale computational analysis of film},
  author        = {David Bamman and Rachael Samberg and Richard Jean So and Naitian Zhou},
  year          = {2024},
  journal       = {Proceedings of the National Academy of Sciences (PNAS)},
  volume        = {121},
  number        = {46},
  pages         = {e2409770121},
  url           = {https://doi.org/10.1073/pnas.2409770121}
}
\bibliographystyle{colm2026_conference}

\appendix

\section{Other Datasets and Evaluation Metrics}

\paragraph{Other Datasets.}\label{sec:appx:datasets}
Other datasets also exist, such as TV-AD \citep{xie-2024-autoad_zero}, YouDescribe \citep{pitcher_cooper-2024-you_describe} (as used by \citet{oncescu-2021-quer_yd}) and Movie101 v1 and v2 \citep{yue-2023-movie101, yue-2024-movie101_v2}.
Datasets of movie summaries, also known as recaps, exist \citep{dogan-2018-yms,sun-2023-symon,sun-2024-m_symon,lu-2024-cvsv}.
Our choice to work with AD for entire movies is based on the fact that they are self-contained and have long-form, complex narratives.
Datasets of egocentric video with parallel descriptions are similar to AD in providing real-time narration of visual content. 
Epic Kitchens 100 \citep{damen-2022-epic_kitchens_100} and HowTo100M \citep{miech19howto100m} are two examples.
These corpora are specifically designed to tackle action recognition, which is not equivalent to the task of generating AD.

\paragraph{Evaluation.}
Table~\ref{tab:metrics} gives, to the best of our knowledge, an exhaustive summary of metrics that have been applied to the previous task of AD as video captioning.
Rows (1)--(4) are variants of n-gram metrics, (5)--(6) were designed for image captioning using semantic scene parses or neural embeddings, (7)--(8) are tailored to report on repetition and character naming in AD, (9--11) use LLMs, either generation or encoder models, (12) addresses only verbs in generated AD, and (13)--(14) are an existing QA evaluation setup for AD.

\begin{table}[h]
\centering
\begin{small}
\begin{tabular}{@{}l@{~}p{2.7cm}@{\hspace{-.6cm}}c@{~}c@{~~}p{4cm}@{}}
\toprule
& \textbf{Metric} & \textbf{First proposed by} & \textbf{For AD proposed by} & \multicolumn{1}{c}{\textbf{Description}} \\
\midrule
(1) & BLEU & \citet{papineni-2002-bleu} & \citet{rohrbach-2015-mpii_md} & n-gram precision + brevity penalty \\
(2) & METEOR & \citet{banerjee-2005-meteor} & \citet{rohrbach-2017-lsmdc} & $F_{\textrm{mean}}$ of unigram with stem and synonym matching + fragmentation penalty. \\
(3) & ROUGE-L & \citet{lin-2004-rouge} & \citet{han-2023-autoad_i} & $F_1$ of longest common subsequence\\
(4) & CIDEr & \citet{vedantam-2015-cider} & \citet{han-2023-autoad_i} & Cosine similarity of TF-IDF on n-grams  \\
(5) & SPICE & \citet{anderson-2016-spice} & \citet{han-2023-autoad_i} & $F_1$ of a semantic graph   \\
(6) & BERT-Score & \citet{zhang-2019-bertscore} & \citet{han-2023-autoad_i} & $F_1$ of BERT embeddings \\
(7) & Recall@k/N & --- & \citet{han-2023-autoad_ii} & Rank window-$N$ references by BERT-score and check if the gold pair is in top-$k$ \\
(8) & Character names & --- & \citet{han-2024-autoad_iii} & Proportion of reference characters mentioned \\
(9) & LLM-as-a-Judge & \citet{zheng-2023-llm_as_a_judge} & \citet{han-2024-autoad_iii} & Prompt an LLM for a score in [0, 5] \\
(10) & LLM-as-a-Judge (via AD guidelines) & & \citet{venugopalan-2025-metrics} & Prompt an LLM for scores according to criteria in AD guidelines \\
(11) & Sentence similarity &  & \citet{xie-2025-shot_by_shot} & Max sentence similarity in a window \\
(12) & Verb lemmas &  & \citet{xie-2025-shot_by_shot} & Proportion of reference verbs mentioned \\
(13) & Fact-based QA &  & \citet{kala-2025-what} & Multiple-choice QA using LLMs + correct rationale \\
(14) & Narrative-based QA &  & \citet{kala-2025-what} & As above, but based on aligned plot summary sentences \\
\bottomrule
\end{tabular}
\end{small}
\caption{Evaluation metrics for AD generation. To the best of our knowledge, an exhaustive list of metrics previously applied to AD (framed as video captioning), grouped by type: n-gram overlap (1--4), image-captioning (5--6), AD-specific repetition and character naming (7--8), LLM-based (9--11), verb coverage (12), and QA-based (13--14). \textit{First proposed by} gives each metric's origin; \textit{For AD proposed by} gives the first work to apply it to AD.}
\label{tab:metrics}
\end{table}

\newpage

\section{Professional AD Transcription}
 \label{appx:sec:subtitling}

For AD transcription in the evaluation sets of \reframed and for our use for evaluating data collection techniques, we contracted the work of a movie post-production company.
They performed fully manual AD transcription and timecoding to the nearest frame (approximately 0.04 seconds).
The delivery consists of segments of transcription, each with a start and end timecode.
One segment was defined as a single event, action or other visual element being described (they are frequently sentence-fragments).

We chose a company with experience with feature-film subtitling and AD, and one that assured us that they do not use any automatic AI tools in their work.
For each movie that we provided them, the company researches the movie and the spellings of particularly hard characters and movie-related words.
The timestamps are manually determined based on clicking between individual frames in the video and listening to when the audio begins.
The data went through a full round of professional quality control from the company.

\section{Evaluating Pre-existing Data Collection Techniques}\label{sec:previous_methods}

Here we evaluate the effectiveness of pre-existing data collection techniques.
The first task is for a system to take as input a movie-length AD track, and output a transcription of it (text of what was said) and diarization (speaker labels for each span within the transcription; needed to differentiate character dialogue from the AD narrator).
The second task is for a system to take as input a short movie clip and a full-length AD movie track, and output the offset and speedup of the clip with respect to the full track.

In order to define the first task, we need a gold-standard AD transcription.
For the second task, we need aligned full-length movie and AD tracks, and then we can create evaluation data by extracting segments from the movie track and applying speed changes.
For these evaluations, we used a newly transcribed version of MAD-v2-eval, which we call MAD-v3-eval.
We purchased the 10 movies in this evaluation set.
Figure~\ref{fig:timediff} on page \pageref{fig:timediff} shows the differences between MAD-v2-eval and MAD-v3-eval timestamps.
A positive difference indicates that the AD described the event before it occurred, and vice-versa.
A Gaussian best fit shows that on average AD describes events 0.8 seconds in advance, and the standard deviation is 2.4 seconds.
This spread demonstrates that a central part of the task of creating AD is identifying suitable gaps in which to place a description that is nearby to its corresponding visual content; the visual event being described is not necessarily at exactly at the same time as when it is described in the AD.

\subsection{Evaluating Automatic AD Transcription.}\label{sec:asr}
We compare the capabilities of three ASR systems on the task of transcribing AD: WhisperX \citep{bain-2023-whisperx} as used for the CMD-AD data; industry APIs from Microsoft (as used for the MAD training data); and Speechmatics (a reputable company that specialises in ASR).
We can also evaluate the quality of the MAD-v2-eval transcripts against our new reference.
Characters play a central role in movie narratives, and hence it is vital that they are transcribed correctly.
We apply pre-existing methods for guiding ASR models to spell particular words correctly, which for us is character names.
For more details, see Section~\ref{appx:sec:asr}.

\begin{table}

\centering

\begin{small}

\begin{tabular}{lr@{\hspace{2mm}}r@{\hspace{2mm}}r@{\hspace{2mm}}r@{\hspace{2mm}}r@{\hspace{2mm}}r@{\hspace{2mm}}r@{\hspace{2mm}}r@{\hspace{2mm}}r@{\hspace{2mm}}r@{\hspace{2mm}}r@{\hspace{2mm}}r@{\hspace{2mm}}r@{\hspace{2mm}}r@{\hspace{2mm}}}
\toprule
& \multicolumn{4}{c}{\bfseries ASR} & \multicolumn{3}{c}{\bfseries Diarization} & \multicolumn{1}{c}{\bfseries Names} & \multicolumn{2}{c}{\bfseries Times} & \multicolumn{4}{c}{\bfseries Overall} \\
\cmidrule(lr){2-5} \cmidrule(lr){6-8} \cmidrule(lr){9-9} \cmidrule(lr){10-11} \cmidrule(lr){12-15}
&  {{SR}} & {{DR}} & {{IR}} & {{WER}} & {{DER}} & {{MR}} & {{FAR}} & {{Micro}} & {{$0.2$}} & {{$0.4$}} & {{SR}} & {{DR}} & {{IR}} & {{WER}} \\
\midrule
\textbf{MAD-v2-eval} & & & & & & & & &  & & 4.6 & 16.7 & 0.8 & 22.1 \\ \midrule
\textbf{Microsoft} & 4.5 & 1.2 & 1.8 & 7.5 & 20.6 & 0.4 & 20.2 & 25.9  & 2.8 & 0.3 & 4.6 & 0.6 & 12.6 & 17.8 \\
\textbf{WhisperX} & 2.9 & 2.7 & 1.0 & 6.6 & 14.9 & 3.7 & 11.3 & 22.3 & 5.1 & 1.0  & 3.6 & 2.4 & 7.7 & 13.7 \\
\textbf{Speechmatics} & \textbf{2.3} & \textbf{0.7} & \textbf{0.6} & \textbf{3.6} & \textbf{2.5} & \textbf{0.3} & \textbf{2.2} & \textbf{16.2} & \textbf{0.7} & \textbf{0.0} & \textbf{2.3} & \textbf{0.5} & \textbf{0.8} & \textbf{3.6} \\ \midrule
\multicolumn{15}{c}{\bfseries Character-name adapted} \\ \midrule
\textbf{WhisperX}  & 1.7 & 2.6 & 1.1 & 5.4 & 14.1 & 3.4 & 10.7 & 7.2 &  5.3 & 1.0 & 2.2 & 2.5 & 8.1 & 12.8 \\
\textbf{Speechmatics} & \textbf{1.5} & \textbf{0.7} & \textbf{0.6} & \textbf{2.8} & \textbf{2.7} & \textbf{0.2 }& \textbf{2.5} & \textbf{4.8} & \textbf{0.8} & \textbf{0.1} & \textbf{1.5} & \textbf{0.5} & \textbf{0.8} & \textbf{2.9} \\
\bottomrule
\end{tabular}
\end{small}
\caption{Automatic AD transcription error rates (\%); lower is better, best per column in \textbf{bold}. \textbf{ASR} (with gold-standard diarization): word error rate (WER) and its substitution (SR), deletion (DR), and insertion (IR) components. \textbf{Diarization}: binary diarization error rate (DER), with miss rate (MR) and false alarm rate (FAR). \textbf{Names}: character-name error rate, micro-averaged over reference characters. \textbf{Times}: timestamp error rate within a collar of $0.2$ or $0.4$ seconds. \textbf{Overall}: end-to-end WER (and SR/DR/IR) after automatic diarization. The lower block repeats the systems with character-name adaptation. See Appendix~\ref{appx:sec:asr} and Tables~\ref{appx:tab:standard}--\ref{appx:tab:character_guided} for full details and all system variants.}
\label{tab:asr}
\end{table}

Results are in Table~\ref{tab:asr}.
MAD-v2-eval achieves overall WER of 22.1\%, largely as a result of a high deletion rate: evidence that this data is of insufficient quality.
WhisperX is better with a WER of 12.8\%.
Across all metrics, the Speechmatics systems perform best with overall WER of 2.9\%.
The good overall performance can be traced back, in part, to excellent diarization performance.
They also provide the only systems with $\le$1\% timestamp error rate within 0.2 seconds.
With our character adaptation, it spells $\le$5\% of character names incorrectly.

\subsection{Evaluating Temporal Alignment of Scenes to Whole Movies.}\label{sec:scene-alignment}

\begin{figure}[t]
\centering
\begin{tikzpicture}

  \begin{pgfonlayer}{background}
    \begin{axis}[
        ybar interval,
        xlabel={Speedup},
        ylabel={\# Count},
        xmin=0.90,
        xmax=1.05,
        ymax=14000,
        axis x line=bottom,
        axis y line=left,
        x tick label as interval,
        yticklabel style={
            /pgf/number format/fixed,
            /pgf/number format/precision=5
          },
        xticklabel style={
            anchor=east,
            yshift=-5pt,
            /pgf/number format/fixed,
            /pgf/number format/precision=2,
            /pgf/number format/fixed zerofill
          },
        scaled y ticks=false,
        ticklabel style={font=\small},
        xtick={0.90, 0.91,..., 1.05},
        scaled x ticks=false,
        width=\linewidth,
        height=5cm,
        xmajorgrids=false,
        ymajorgrids=true
      ]
      \addplot+[fill=reframed-red,draw=black] table {data/speedup.dat};

      \coordinate (insetleft) at (axis description cs:0.3,0.8);
      \coordinate (insetright) at (axis description cs:0.95,0.8);
    \end{axis}
  \end{pgfonlayer}

  \begin{pgfonlayer}{foreground}
    \begin{axis}[
        at={(insetleft)},
        anchor=north east,
        small,
        ybar interval,
        xmin=0.955,
        xmax=0.96,
        ymin=0,
        ymax=3500,
        xtick={.955, .956,..., .960},
        axis x line=bottom,
        axis y line=left,
        width=4cm,
        height=3cm,
        xmajorgrids=false,
        ymajorgrids=true,
        xticklabel style={
            font=\tiny,
            rotate=45,
            anchor=east,
            /pgf/number format/fixed,
            /pgf/number format/precision=3,
            /pgf/number format/.prefix=empty
          },
        scaled y ticks=false,
        ticklabel style={font=\tiny},
        axis background/.style={fill=white},
        name=insetleftAxis,
      ]
      \addplot+[fill=reframed-red,draw=black] table {data/speedup.dat};
    \end{axis}
  \end{pgfonlayer}

  \begin{pgfonlayer}{foreground}
    \begin{axis}[
        at={(insetright)},
        anchor=north east,
        small,
        ybar interval,
        xmin=0.995,
        xmax=1.005,
        ymin=0,
        ymax=14000,
        xtick={.995, .996,..., 1.005},
        axis x line=bottom,
        axis y line=left,
        width=4cm,
        height=3cm,
        xmajorgrids=false,
        ymajorgrids=true,
        xticklabel style={
            font=\tiny,
            rotate=45,
            anchor=east,
            /pgf/number format/fixed,
            /pgf/number format/precision=3,
            /pgf/number format/.prefix=empty,
            /pgf/number format/fixed zerofill
          },
        scaled y ticks=false,
        ticklabel style={font=\tiny},
        axis background/.style={fill=white},
        name=insetrightAxis,
      ]
      \addplot+[fill=reframed-red,draw=black] table {data/speedup.dat};

    \end{axis}
  \end{pgfonlayer}

  \begin{pgfonlayer}{main}
    \fill [black!10] ([shift={(-2pt,-2pt)}] insetleftAxis.outer south west)
    rectangle ([shift={(+2pt,+2pt)}] insetleftAxis.outer north east);
  \end{pgfonlayer}

  \begin{pgfonlayer}{main}
    \fill [black!10] ([shift={(-2pt,-2pt)}] insetrightAxis.outer south west)
    rectangle ([shift={(+2pt,+2pt)}] insetrightAxis.outer north east);
  \end{pgfonlayer}

\end{tikzpicture}
\vspace{-15pt}
\caption{A histogram of the speedups computed by \citet{han-2024-autoad_iii} for CMD-AD.}
\label{appx:fig:speedup}
\end{figure}
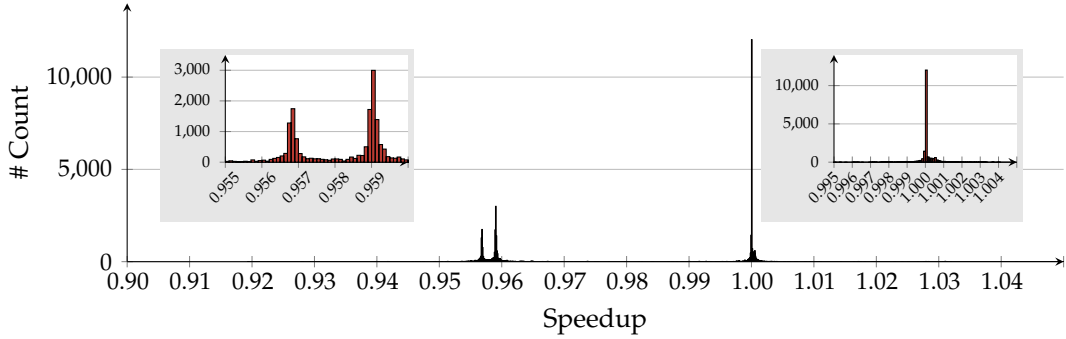

The task now is to temporally identify where in a full-length AD track a given movie scene came from.
This task is made more challenging in practice because of the speed differences caused by the different frame rates of the two main standards: NTSC (as used in North
America, which uses 24000/1001 frames per second) and PAL (as used in Europe, which uses 25 frames per second).
In order to treat potential speed differences, all we need to do is allow for one version to be NTSC and the other PAL, or both are NTSC/PAL.
Figure~\ref{appx:fig:speedup} shows reverse-engineered data released by \citet{han-2024-autoad_iii} --- experimentally we find that their predicted speedups are consistent with these two theoretical results.

Hence our task definition is: given a full-length AD track and a movie scene, identify the time of the start of the movie scene in the AD track and determine if the scene needs a PAL-NTSC speeddown, no speed change, or a NTSC-PAL speedup.
\begin{wraptable}{r}{6cm}
\centering
\begin{small}
\begin{tabular}{lc}
\toprule
\textbf{Parameter} & \textbf{Value} \\
\midrule
Target sample rate & 16000 \\
Number of Mel bands & 128 \\
FFT window size & 1024 \\
Hop length & 512 \\
RANSAC window size & 50 \\
\bottomrule
\end{tabular}
\end{small}
\caption{Hyperparameters for audio feature extraction and cross-correlation alignment.}
\label{tab:align_params}
\end{wraptable}
We create many samples of each case by using the pre-aligned full movies in MAD-v3-eval, and we do so for many different lengths of movie scene.
\citeposs{han-2024-autoad_iii} RANSAC-based method can directly be applied to the task; for evaluation, we consider any predicted speed difference within 0.01 of the reference to be correct.
\citeposs{soldan-2022-mad} cross-correlation can only predict the offset, so we adapt it by repeating the cross-correlation computation for each of the possible speed changes and use the one with highest correlation score to be the prediction.
We consider offsets to be correct if they are within 0.04 seconds of the reference.
Table~\ref{tab:align_params} details the specific hyperparameters used for feature extraction and alignment.

\begin{figure}
    \centering
    \begin{subfigure}[t]{0.48\textwidth}
    \begin{tikzpicture}
  \begin{axis}[
      xmax=1049,
      ymin=0, ymax=100,
      ymajorgrids=true,
      xmode=log,
      xlabel=Clip duration (s),
      ylabel=Offset accuracy (\%),
      grid style=dashed,
      height=5cm,
      width=\linewidth,
      ylabel style={font=\small},
      xlabel style={font=\small},
      xticklabel style={font=\small},
      yticklabel style={font=\small},
      height=5cm,
      axis x line*=bottom,
      axis y line*=left,
      axis line style={->, >=latex},
      legend pos=north west,
      legend style={
          font=\small,
          cells={anchor=west},
          opacity=0.7,
        },
    ]
    \addplot[reframed-green, thick, mark=x] coordinates{(0.002, 0.00) (0.004, 0.00) (0.008, 0.00) (0.016, 1.00) (0.032, 6.00) (0.064, 30.00) (0.128, 59.00) (0.256, 68.00) (0.512, 63.00) (1.024, 68.00) (2.048, 77.00) (4.096, 82.00) (8.192, 91.00) (16.384, 99.00) (32.768, 100.00) (65.536, 100.00) (131.072, 100.00) (262.144, 100.00) (524.288, 100.00) (1048.576, 100.00)};

    \addplot[reframed-red, thick, mark=x] coordinates{(2.048, 0.00) (4.096, 0.00) (8.192, 0.00) (16.384, 0.00) (32.768, 0.00) (65.536, 0.00) (131.072, 0.00) (262.144, 12.00) (524.288, 43.00) (1048.576, 92.00) };

    \legend{\textsc{Cross-correlation}, \textsc{RANSAC}}

  \end{axis}
\end{tikzpicture}
    \caption{Accuracy of determining movie scene offset.}
    \end{subfigure}
    \hfill
    \begin{subfigure}[t]{0.48\textwidth}
    \begin{tikzpicture}
  \begin{axis}[
      xmax=1049,
      ymin=0, ymax=100,
      ymajorgrids=true,
      xmode=log,
      xlabel=Clip duration (s),
      ylabel=Speedup accuracy (\%),
      grid style=dashed,
      height=5cm,
      width=\linewidth,
      ylabel style={font=\small},
      xlabel style={font=\small},
      xticklabel style={font=\small},
      yticklabel style={font=\small},
      height=5cm,
      axis x line*=bottom,
      axis y line*=left,
      axis line style={->, >=latex},
    ]
    \addplot[reframed-green, thick, mark=x] coordinates{((0.002, 45.00) (0.004, 45.00) (0.008, 33.00) (0.016, 15.00) (0.032, 17.00) (0.064, 43.00) (0.128, 38.00) (0.256, 56.00) (0.512, 36.00) (1.024, 21.00) (2.048, 63.00) (4.096, 85.00) (8.192, 88.00) (16.384, 96.00) (32.768, 100.00) (65.536, 100.00) (131.072, 100.00) (262.144, 100.00) (524.288, 100.00) (1048.576, 100.00)};

    \addplot[reframed-red, thick, mark=x] coordinates{(2.048, 0.00) (4.096, 0.00) (8.192, 0.00) (16.384, 0.00) (32.768, 0.00) (65.536, 0.00) (131.072, 0.00) (262.144, 11.00) (524.288, 41.00) (1048.576, 85.00)};

  \end{axis}
\end{tikzpicture}
    \caption{Accuracy of determining the speed difference.}
    \end{subfigure}
    \caption{Scene-to-movie temporal alignment accuracy as a function of scene duration, comparing cross-correlation \citep{soldan-2022-mad} against the RANSAC-based method of \citet{han-2024-autoad_iii}. Cross-correlation reaches perfect accuracy for scenes of $\geq$32 seconds, while the RANSAC method is reliable only at much longer durations; we therefore adopt cross-correlation in \protect\reframed.}
    \label{fig:scene-alignment}
\end{figure}
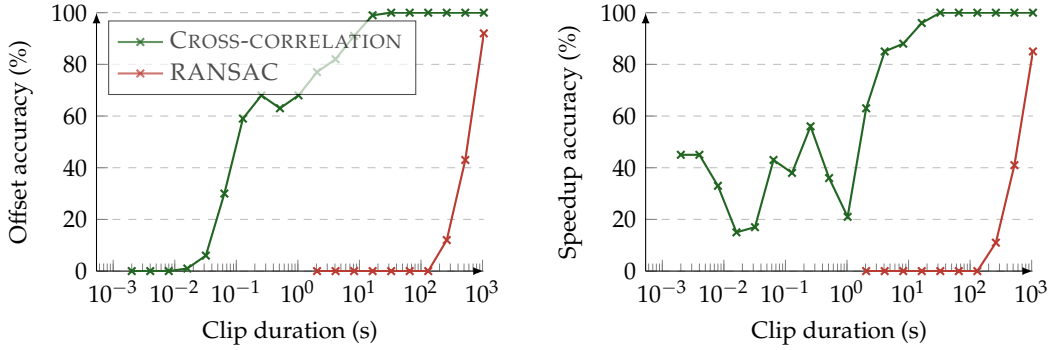

Figure~\ref{fig:scene-alignment} provides results.
We can see that the success of the methods depends greatly on the length of the scene to be aligned.
The only approach to achieve perfect performance is using cross-correlation on scenes of at least 32 seconds.
The RANSAC-based method performs poorly without very long scene durations.
Our experiment here also suggests why the RANSAC method is inferior---it relies at its core on applying many cross-correlation to 3 second windows, and we have shown here that cross-correlation is very noisy at this time-scale.

In \reframed, we opt to use cross-correlation.
In order to guarantee the highest possible quality assurance, we additionally require that all the offsets for a video from a given movie impute an OLS regression line between the US and UK AD tracks with RMSE $\le1$.
We state that we have high confidence in the alignment of the resulting videos based on the fact that it would be extremely unlikely for $\geq$5 independent alignment offsets to maintain linear consistency across the entire movie if one or more was erroneous.

\section{Challenge Set Annotation}\label{appx:sec:challenge}

The data in the \reframed challenge set of full movies serves both as an evaluation for AD generation and as an evaluation material for further data collection steps that we perform.
Annotation for this set was performed manually.

\paragraph{High-level Overview.}
There are 10 movies, and each movie is provided with 2-3 AD transcripts.
The movies are either 2023/2024 movies with award winning AD tracks (2/10), have three independently-created AD versions (1/10), or are the same movies as in MAD-v3-eval (7/10).
The exact frame that starts each new scene in every movie is labeled.
Each AD segment is labeled with a label of the scene whose content it describes.
The screenplays scenes are mapped to the movie scenes.
Boundaries between sequences of scenes are obtained from the official movie release chapter boundaries.
English SDH and English dialogue subtitles are from the official movie release.

\paragraph{Details.}

For each movie in our benchmark, we require the availability of independently created US and UK AD versions for use as multiple references.
For seven out of the ten movies in MAD-v3-eval, we were able to identify and purchase movies with the other AD version to the version in the previous data.
We add two movies which received awards for the quality of their AD tracks in the AD Awards gala and/or EGA Hermes Awards and also have US and UK AD versions, namely Dune Part Two (2024) and Barbie (2023).
We add a final movie for which we were able to obtain three independent AD tracks (two US versions, one UK version), namely Priscilla (2023).
The benchmark now contains movies that span two decades (2005--2024) and a wide range of genres.

We contracted the film post-production company to provide us with professional AD transcripts for each of the 21 (9 movies with two references, 1 movie with three references) tracks.
An annotator manually determined if PAL--NTSC speedups and/or offsets are required between the versions and ensure that all versions are temporally aligned, determined the scene boundaries in the movie and determined which AD segments correspond to which scene.
The process was as follows: (1) using professional subtitling software, we labeled the exact frame when each scene changes, following a strict definition of scene changes as a change of location or time, (2) using these boundaries, the movie video, and the timecoded AD, we label the scene to which each AD segment is describing content.
Screenplays for 4/10 of the movies are provided by MovieSum \citep{saxena-2024-moviesum}, and for the remaining 6 we follow \citeposs{saxena-2024-moviesum} parsing method to obtain parsed screenplays, which involves full manual review and correction.
The annotator manually labels the alignment between the movie scenes and the scenes in the screenplays from all ten movies.
There was no access to the screenplay to do the scene segmentation, as a result one-to-many and many-to-one alignments are possible.
Chapter boundaries are provided directly by our purchased versions of the movies, which we use as sequence boundaries.
We obtain professional dialogue subtitles by applying OCR (Tesseract 5.5.0) to the official movie subtitles and manually reviewing (using Subtitle Edit software).
We apply this to the English SDH and English dialogue subtitles, or if only the former is available (4/10 cases) then we manually edit the SDH version to remove text other than the dialogue subtitles.

The only potentially subjective task of the above is determining scene boundaries from the movie.
We do not perform any double-annotation and hence cannot report inter-annotator agreement on the task.
One of the movies in our data is also provided as a part of MovieNet \citep{huang-2020-movienet}, which provides data that is the established benchmark for movie scene segmentation.
The MovieNet annotation of scene segmentation is crowd-sourced and based on the decisions of an annotator (who likely has not watched the full movies) provided access only to stills from neighbouring shots (determined via automatic shot detection).
For a movie in both datasets, \textit{Les Miserables} (2012), we were surprised to find their annotation reports 228 scenes, while ours has only 113.
We inspected their boundaries and found them to consistently oversegment, in particular in settings when neighbouring shots are visually different but in fact just a different view of the same scene.
As an independent reference, the screenplay of this movie has 133 scenes (some of which were likely removed for the final version).

\section{New Data Collection Techniques}\label{sec:new_data_collection_techniques}

Working with screenplays brings challenges, both in terms of parsing the PDFs available of them and aligning them to the final movie.

Screenplays are released as 100+ page PDFs, with a reasonably standard scene-segmented structure that is displayed via indentations and font formatting.
The task of parsing such documents into their structured contents has been addressed by prior work \citep{turetsky-2004-screenplay,agarwal-2014-parsing,winer-2017-automated}.
However, the task remains challenging due to different formatting styles \citep{gil-2011-extraction} and in practice several hours of human supervision is needed per movie for perfect parsing \citep{saxena-2024-moviesum}.
We do not tackle this task and instead rely on the gold-standard machine-parsed screenplays from the MovieSum dataset \citep{saxena-2024-moviesum}.

However, we found current alignment methods to be insufficient for our purposes.
In our approach to screenplay alignment, we define the task at the scene-level as follows: given $\mathcal{S} = \{s_i\}_{i=1}^S$ scenes from a machine-parseable screenplay and the dialogue and AD from $\mathcal{M} = \{m_j\}_{j=1}^M$ scenes in the final movie, compute the alignment $A \subseteq \mathcal{S} \times \mathcal{M}$.
The screenplay's structured format provides exact segmentation into the $S$ scenes, but scene segmentation is not available for the final movie.
Our insight here is that AD will frequently cue movie scene changes; this is a requirement stipulated in both American and British AD guidelines.
If we can detect such cues, then we have a high-precision and scalable method of determining movie scene boundaries.

\subsection{Movie-Screenplay Alignment}\label{sec:screenplay-alignment}

\paragraph{Evaluation and Baselines.}
Our evaluation metric for movie-screenplay alignment is weighted $F_1$ score, where the weights are based on the lengths of the scenes in the movie.
We will compare our methods against the previous state-of-the-art: aligning a movie's dialogue subtitles with dialogue in the movie's screenplay by computing a dynamic time warping (DTW) over word tokens.
In this baseline, we assign any screenplay dialogue word with an exact match to a movie dialogue word the scene the movie word is from.

DTW over word matches is frequently applied for the task of movie-screenplay alignment \citep{everingham-2006-hello,sivic-2009-who,bouritsas-2018-multimodal,papalampidi-2021-movie}.
Nevertheless, results from \citet{lambert-2013-reordering} suggest that DTW may not be the ideal algorithm, because it favours on-diagonal matches in the alignment.
This is a particular problem for screenplay drafts before a post-production or even shooting version.
\citet{lambert-2013-reordering} propose to upweight the score that results from matches; this is akin to minimizing edit distance, which we will employ as a second baseline.\footnote{We use the implementation from \url{https://github.com/jitsi/jiwer}}

\paragraph{Vecalign-based Alignment.}
In our approach, we compute embeddings for each scene in the movie and each scene in the screenplay, and then use dynamic programming to align the two.
We additionally treat the potential for a screenplay scene to have been removed at production, or for a new scene to be added at production.
\citet{thompson-2019-vecalign} provide methods designed for compiling machine translation corpora that can be of use here, which they call Vecalign.
In particular they define a cost in the alignment algorithm which allows scenes to be skipped in either the movie or the screenplay, without alignment.
We use a fixed skip cost $c_{skip}=0.4$.
Vecalign also treats one-to-many and many-to-one alignments by allowing for merge operations between contiguous sequences of scenes.
We use mean-pooling as the merge operation.
We allow for alignment blocks of up to 10 scenes.
Vecalign also allows for, albeit discourages in the cost function, many-to-many alignments; we disallow these via search constraints so that the final alignment is strictly monotonic over the movie scenes.
Vecalign also incorporates methods that reduce the alignment time complexity from quadratic to linear via a recursive approximation; we do not apply these methods and optimize over the full alignment.

The cost $c(s, m)$ of aligning a candidate block of screenplay scenes $s \subseteq \mathcal{S}$ to a candidate block of movie scenes $m \subseteq \mathcal{M}$ is defined as:
\begin{equation}
    c(s, m) = \frac{\big(1 - \cos(\bar{\mathbf{s}}, \bar{\mathbf{m}})\big) \cdot |s| \cdot |m|}{\frac{1}{2M} \sum_{j=1}^{M} \big(1 - \cos(\bar{\mathbf{s}}, \mathbf{m}_j)\big) + \frac{1}{2S} \sum_{i=1}^{S} \big(1 - \cos(\mathbf{s}_i, \bar{\mathbf{m}})\big)}
\end{equation}

where boldface denotes scene embeddings, bar notation denotes embeddings after mean-pooling, and $|\cdot|$ denotes the number of scenes in a block.
The numerator scales the similarity cost by the number of scenes in the block alignment, and the denominator normalizes by the mean similarity between the candidate block and all scenes.

With initialization $D(0,0)=0$, $D(i,0) = i \cdot c_{\text{skip}}$, and $D(0,j) = j \cdot c_{\text{skip}}$, the dynamic programming recurrence relation is as follows:
\begin{equation*}
    D(i, j) = \min \begin{cases}
      D(i-1, j) + c_{\text{skip}} & \text{(skip screenplay scene)} \\
      D(i, j-1) + c_{\text{skip}} & \text{(skip movie scene)} \\
      \displaystyle \min_{|s|, |m|} \Big[ D(i-|s|, j-|m|) + c\big(\mathcal{S}_{i-|s|:i}, \mathcal{M}_{j-|m|:j}\big) \Big] & \text{(block match)}
   \end{cases}
\end{equation*}

where the search space is constrained such that $\min(|s|, |m|) = 1$ and $|s| + |m| \le 10$.
The final alignment $A$ is then recovered by backtracking along the minimal-cost path from $D(S, M)$ to $D(0,0)$ and collecting the matched scenes.

\paragraph{Embeddings.}
For the movies, we will define the embeddings based on a time-sorted concatenation of the subtitles (SDH or dialogue-only) and/or AD (UK and/or US versions) in each scene in our gold-standard annotation; for the screenplay we define the embedding based on the dialogue and scene descriptions in each scene in the order that they appear.
Embeddings are computed according to \citet{zhang-2025-qwen3embedding}.

\paragraph{Results.}
Results are in Table~\ref{tab:screenplay_alignment}.
The word-based DTW baseline achieves $F_1=62.3$, and the minimum edit distance variant achieves $F_1=63.8$.
Performance with Vecalign applied to the same dialogue is comparable at $F_1=64.2$.
Using instead SDH leads to an improved $F_1=68.0$ (+3.8), demonstrating that the non-verbal audio captions contain useful information for the alignment.
We can outperform these dialogue-based alignment approaches using only one of UK or US AD ($F_1=71.3$ and $F_1=73.0$), demonstrating that AD also contains useful information for the alignment.
Using both AD versions results in better performance than using only one ($F_1=74.5$, +3.2 and +1.5 for AD versions).
(Table~\ref{tab:screenplay_alignment} also gives results for other combinations of input data.)
Our best results ($F_1$=80.9) are based on using SDH and both AD versions with Vecalign.
In our experiments so far, we have used the Qwen3 Embedding 8B model for computing the embeddings.
Using the 4B or 0.6B variants of this model leads to worse performance ($F_1$=78.8 and $F_1$=71.8, respectively).
We apply this best alignment pipeline to all movies with screenplays in \reframed.

\begin{table}
\centering
\begin{tabular}{llcccc}
\toprule
\textbf{Method} & \textbf{Data} & \textbf{Model} & \textbf{$P$} & \textbf{$R$} & \textbf{$F_1$} \\
\midrule
Text-based DTW & dialogue & --- & 63.7 & 62.4 & 62.3 \\
Text-based MED & dialogue & --- & 66.4 & 63.1 & 63.8 \\ \midrule

Vecalign & dialogue & Qwen3 E 8B  &   81.4  &   53.4  &   64.2  \\
Vecalign & SDH & Qwen3 E 8B  &   82.8  &   58.5  &   68.0    \\
Vecalign & UK AD & Qwen3 E 8B     &   76.9  &   67.1  &   71.3   \\
Vecalign & US AD & Qwen3 E 8B &   77.0  &   70.0  &   73.0   \\
Vecalign & UK AD + US AD & Qwen3 E 8B   &   76.9  &   72.8  &   74.5   \\
Vecalign & dialogue + UK AD & Qwen3 E 8B    &   82.1  &   73.7  &   77.3   \\
Vecalign & SDH + UK AD & Qwen3 E 8B     &   83.6  &   73.1  &   77.5     \\
Vecalign & dialogue + US AD & Qwen3 E 8B     &   84.7  &   73.9  &   78.6    \\
Vecalign & SDH + US AD & Qwen3 E 8B   &   85.1  &   73.9  &   78.7     \\
Vecalign & dialogue + UK AD + US AD & Qwen3 E 8B    &   83.1  &   77.4  &   79.9     \\ \midrule
Vecalign & SDH + UK AD + US AD & Qwen3 E 0.6B &   87.5  &   62.4  &   71.8    \\
Vecalign & SDH + UK AD + US AD & Qwen3 E 4B &   85.3  &   73.8  &   78.8    \\
Vecalign & SDH + UK AD + US AD & Qwen3 E 8B &   84.9  &   77.7  &   \textbf{80.9}    \\
\bottomrule
\end{tabular}
\caption{Screenplay alignment results. DTW = dynamic time warping, MED = minimum edit distance. Text-based baselines are in first block, ablation results are in second block, and best results based on embedding models of three sizes are in final block.}
\label{tab:screenplay_alignment}
\end{table}

\subsection{Movie Scene Segmentation from AD}\label{sec:scene-segmentation}

\paragraph{AD scene segmentation.}
We detect scene change cues in AD by finetuning RoBERTa-large \citep{liu-2019-roberta} on our manually annotated data.
The task is for a model to classify each segment of AD as starting a new scene or not, given the context of surrounding AD segments.
Our evaluation metric is binary $F_1$ score.
We perform 10-fold cross-validation and report the macro-average.
Our baseline will assume that all AD segments that begin with the word `Now' (this is a word US AD guidelines specify can be used to cue scene changes) start a new scene.

We compare two formulations of the training task.
The first is called `single' and involves performing training and inference on independent sliding windows of six consecutive segments from individual AD transcripts.
The second is called `double' and involves performing training and inference on paired sliding windows from US and UK transcripts to predict congruent boundaries across both.
For `single', we collect many samples of `[CLS] 3 segments [SEP] 3 segments' based on a sliding window of 6 segments.
The learning task is binary classification on the CLS token of whether the [SEP] token represents a scene boundary or not.
We collect all possible samples independently of each of the US and UK AD transcripts.
`Double' utilizes our prior that both the US and UK AD versions are describing the same movie scene changes.
We formulate the task so that the model can only predict congruent scene boundaries in both AD versions.
The approach involves jointly training on sequences of US and UK AD using the following template: `[CLS] 3 segments of US AD [SEP] 3 segments of UK AD [SEP] 3 segments of US AD [SEP] 3 segments of UK AD'.
The two versions may not begin narrating a scene at the same time.
In fact, in 8.7\% of scene boundaries in our data, after the first AD version starts a new scene, the other AD version starts at least one new segment of the previous scene, before moving to the next scene.
The boundary is fuzzy, and with this approach and two sliding windows, we can treat this.
We allow for the sliding windows to separate at the start of segment number four (the potentially new scene) up to 10 seconds, or the next AD if there is none within 10 seconds for one of the versions.
The training task is now whether the second and third [SEP] tokens represent scene boundaries to the same next scene.
To ensure the predicted scene boundaries are consistent, we use greedy inference, discarding predictions that are incompatible with earlier ones.
We do not want our model to learn any signals present in our gold-standard AD transcripts, but not in automatic ASR transcripts.
Hence we train on transcripts from Speechmatics, segmented according to the gold-standard segmentation.
The full set of training hyperparameters is provided in Table~\ref{tab:scene_seg_params}.

\begin{table}
\begin{minipage}[t]{0.45\textwidth}

\centering

\begin{small}
\begin{tabular}{lc}
\toprule
\textbf{Parameter} & \textbf{Value} \\
\midrule
Block size & 512 \\
Learning rate & 1e-5 \\
Batch size & 64 \\
Epochs & 3 \\
Weight decay & 0.01 \\
Warmup ratio & 0.06 \\
\bottomrule
\end{tabular}
\end{small}
\captionof{table}{Hyperparameters for the scene segmentation fine-tuning.}
\label{tab:scene_seg_params}
\end{minipage}
\hspace{0.05\textwidth}
\begin{minipage}[t]{0.45\textwidth}
\centering
\begin{small}
\begin{tabular}{lc}
\toprule
\textbf{Parameter} & \textbf{Value} \\
\midrule
Block size & 128 \\
Learning rate & 1e-5 \\
Batch size & 32 \\
Epochs & 3 \\
Weight decay & 0.01 \\
Warmup ratio & 0.06 \\
\bottomrule
\end{tabular}
\end{small}
\captionof{table}{Hyperparameters for the token classification AD splitting model.}
\label{tab:ad_split_params}
\end{minipage}
\end{table}

\begin{table*}
\centering
\begin{minipage}[t]{0.46\textwidth}
\begin{tabular}{@{}l@{~}lp{0.9cm}p{0.9cm}p{1.2cm}@{}}
\toprule
& & $P$ & $R$ & $F_1$ \\
\midrule
Baseline && --- & 11.9$_{\pm0.0}$ & --- \\ \midrule
RoBERTa & sin. & 59.3$_{\pm0.4}$ & 55.3$_{\pm0.1}$ & 56.6$_{\pm0.2}$ \\
       (base)        & dou. & 73.2$_{\pm1.1}$ & 45.4$_{\pm0.3}$ & 55.5$_{\pm0.4}$ \\ \midrule
RoBERTa& sin. & 65.0$_{\pm0.2}$ & \textbf{55.8}$_{\pm0.1}$ & \textbf{59.5}$_{\pm0.2}$ \\
       (large)        & dou. & \textbf{76.1}$_{\pm0.4}$ & 49.2$_{\pm0.7}$ & \textbf{59.2}$_{\pm0.4}$ \\
\bottomrule
\end{tabular}
\captionof{table}{Scene segmentation performance (binary $F_1$, precision $P$, recall $R$; 10-fold CV, $\pm$ std). \textit{sin.}: per-version training; \textit{dou.}: joint US/UK training for congruent boundaries. Baseline: segments starting with ``Now'' begin a new scene ($P$ undefined as some tracks never use it).}
\label{tab:scene-seg}
\end{minipage}
\hspace{.5cm}
\begin{minipage}[t]{0.46\textwidth}
\centering
\begin{tabular}{@{}l@{~}p{0.9cm}p{0.9cm}p{1.2cm}@{}}
\toprule
 & $P$ & $R$ & $F_1$ \\
\midrule
Baseline & 38.0$_{\pm0.0}$ & 38.4$_{\pm0.0}$ & 36.9$_{\pm0.0}$ \\ \midrule
RoBERTa (base)  & 64.9$_{\pm0.1}$ & 66.2$_{\pm0.0}$ & 63.7$_{\pm0.1}$ \\
RoBERTa (large) & \textbf{65.8}$_{\pm0.0}$ & \textbf{68.5}$_{\pm0.1}$ & \textbf{64.8}$_{\pm0.0}$ \\
\bottomrule
\end{tabular}
\captionof{table}{AD splitting performance (character-level $F_1$, precision $P$, recall $R$; 10-fold CV, $\pm$ std). Task: classify whether each character ends a description element. Baseline: split at every comma.}
\label{tab:ad-frag}
\end{minipage}
\end{table*}

Results are in Table~\ref{tab:scene-seg}.
The baseline approach achieves a recall score of 11.9, but precision is undefined since some of the movie ADs never use `now'.
The `single' and `double' approaches with the RoBERTa-large model achieve similar $F_1$ measures of 59.5 and 59.2, respectively.
Using the smaller RoBERTa-base model results in 56.6 and 55.5.
Precision and recall metrics show that they are behaving differently: `double' achieves higher precision, yet lower recall.
This is intuitive, since the `double' relies on a model's prediction that a scene change is distinguishable in both transcripts.
Since `double' brings the additional benefit of the scene boundaries guaranteed to be congruent in the two versions, we consider it best and use it.
For the earlier video extract from \textit{The Girl with the Dragon Tattoo}, this model successfully segments into its three constituent scenes: at a tailor's shop where the gift is purchased, at Salander's apartment where she writes a card, and the final street scene in Figure~\ref{fig:dragon-tattoo}; as a result, the Vecalign-based alignment is able to successfully align each of these three scenes with its corresponding scene in the screenplay.

\paragraph{Incorporating video scene boundaries.}
The above procedure provides congruent scene labels for each AD segment in our two AD versions.
\reframed contains AD for individual video scenes, which can also provide a signal of scene boundaries.
In order to assign AD and subtitle segments to each video, we include all segments that are fully contained in the temporal span of the video.
Furthermore, for the screenplay alignment to assign a particular video a number of screenplay scenes, we must enforce that scenes do not span over the boundary of the start or end of a video -- in such cases, we need to divide the scene labels further.
We also wish to transfer these scene labels to the available dialogue and SDH data.
Since some videos in the training set have temporal overlap, we need a general re-assignment algorithm that caters to this.

Our scene re-assignment algorithm meets all of the following requirements: (1) scene boundaries according to the AD classifier are retained; (2) temporal scene boundaries are created for the start and end of every video; and (3) the data associated with each video can be reconstructed by concatenating the data associated with a particular set of scene labels.

This is achieved by constructing an intermediate scene label for every segment before mapping these labels to final shared scene identifiers. Each AD segment is initialized with the scene label assigned by the AD classifier. We then append the identifiers of all videos in which the segment is fully contained. To distinguish different positions relative to video boundaries, we partition the movie timeline at every video start and end time. Each resulting interval is associated with the set of videos active during that interval, which may be none. For each segment, we assign the interval whose active video set matches the segment's fully contained video set, when such an interval is unique. If no unique matching interval exists, the segment is assigned a boundary-spanning interval covering the relevant video-boundary regions. Thus, each AD segment receives an intermediate label consisting of its classifier scene, its fully contained video set, and its position relative to the global video-boundary partition.

For subtitle and SDH segments fully contained in videos, we infer the classifier scene from AD segments in the same video. When multiple AD scenes occur in the video, the subtitle is assigned to the one with greatest temporal overlap. For subtitle and SDH segments outside videos, we infer the classifier scene from global intervals derived from the first occurrence of each AD classifier scene across both AD versions, partitioning the full movie timeline from zero to the latest segment or video end.
The classification again uses greatest temporal overlap. If a video contains subtitle or SDH segments but no AD segments, we add a subtitle-only scene label for that video setting. Finally, all intermediate labels across the two AD versions, subtitles, and SDH subtitles are mapped to shared consecutive scene identifiers in order of their first occurrence.
For the screenplay alignment, each movie scene of length less than 10 seconds and not a part of any video is excluded from the alignment.

\subsection{AD Splitting}\label{sec:fragmentation}

ASR systems output transcripts at the sentence level, which are potentially long (the longest in our data is 74 words).
To be consistent with our evaluation data, we would like to segment transcribed AD sentences into the separate elements they describe.
This task is only defined fully with access to the movie visuals, but we expect that reasonable performance is possible using only the textual sentences as input.
Our operationalisation draws inspiration from \citet{frohmann-2024-segment}, who formulate the task of text segmentation as classifying whether each token in a text ends a segment or not.
We create training data by splitting Speechmatics AD transcript sentences according to the segmentation in our gold standard.
This results in 16,496 sentences split into 21,675 segments (26.45\% of sentences have more than one segment).
We evaluate using character-level $F_1$ on segment-ending characters.
Our baseline creates an AD split at every comma.
Table~\ref{tab:ad_split_params} provides the full set of hyperparameters used.

Results are in Table~\ref{tab:ad-frag}.
The comma baseline achieves $F_1$=36.9, while our best model achieves $F_1$=64.8.
Evidently the task is not defined perfectly, likely in part because of our text-only formulation, but we deem this satisfactory performance.
Manual inspection shows that the resulting model is quite robust to punctuation and complex syntax.
For example, the AD `His right hand forms a cannon, which he aims down the corridor before inserting a probe into the mainframe terminal at a radio telescope array, three dishes shift their positions.' becomes 4 segments (starting `His', `which', `before' and `at').
This example also demonstrates how preprocessing our data with this segmentation model is an important pre-processing step before running the scene segmentation model, because the segment beginning `at' actually starts a new scene.

\section{More Details of AD Transcription Experiments}\label{appx:sec:asr}

We evaluate ASR-only performance using gold-standard diarization; metrics include segment error rate (SER) and word error rate (WER) (which is derived from substitution rate (SR), deletion rate (DR) and insertion rate (IR)).
Binary diarization error rate (DER) gives diarization performance; it is derived from miss rate (MR) and false alarm rate (FAR).
We evaluate the error rate of timestamps against collars of $\alpha$ seconds around the reference timestamps.
Character names in the reference are manually identified and character name error rate is a micro average over reference characters spelt correctly.
Finally, overall ASR metrics show end-to-end performance, i.e. after automatic diarization.
All text is pre-processed according to \citet{radford-2023-whisper}.\footnote{\url{https://github.com/openai/whisper/tree/main/whisper/normalizers}}

We can also evaluate the quality of the MAD-v2-eval transcripts against our new reference.
The models we compare are as follows:
\begin{itemize}
\itemsep0pt
    \item \textbf{The system used for AD transcription in MAD-v2-train and CMD-AD datasets} \citep{han-2023-autoad_i, han-2024-autoad_iii}\textbf{: WhisperX.} It is an open-source system that combines multiple open-source projects. We also tried customizing the system to use \texttt{gpt-4o-transcribe} for the ASR.
    \item \textbf{The system used for AD transcription in the MAD-v1-train dataset} \citep{soldan-2022-mad}\textbf{: Microsoft Batch API system.} Since our data consists of both American English and British English accents. We investigated variants with and without continuous language identification between British and American English; in summary, including this language identification brings substantial performance gains. We show results including using this language identification.
    \item \textbf{A system previously not applied to the task, but generally-reported good performance: Speechmatics Batch API} \texttt{standard} and \texttt{enhanced} systems, global English.
\end{itemize}

Characters play a central role in movie narratives, and hence it is vital that they are transcribed correctly.
We apply pre-existing methods for guiding ASR models to spell particular words correctly, which for us is character names.
We collect lists of all character names in movie credits, except those that are non-unique (e.g. `Citizen') or ending in a numeric character (e.g. `Citizen 1').\footnote{These are available on `fullcredits' pages on IMDb.}
\begin{itemize}
\itemsep0pt
    \item Whisper: we override Whisper's in-context tokens with a script containing all character names. We compared three different script templates and proceeded with the best one: a comma separated list of the character names, which outperformed specifying the name of the movie and a template more realistic of a real conversation.
    \item Microsoft: vocabulary adaptation is not available in Batch API.
    \item Speechmatics: API-based vocabulary adaptation with the full list of character names, including both complete names and their constituent parts.
\end{itemize}

For a full set of numerical results, see Table~\ref{appx:tab:standard} for without character name adaptation, and Table~\ref{appx:tab:character_guided} with character name adaptation.

We performed a spot check of the remaining mistakes in the Speechmatics transcription.
The majority of the mistakes fall into one of four categories: (1) remaining character misspellings (particularly for non-standard name spellings), (2) incorrectly transcribed articles (sometimes they are dropped entirely, sometimes swapped for other articles), (3) incorrectly transcribed homophones, and (4) imperfections in the evaluation normalization script (particularly with relation to compound words).
Furthermore, we found the output to often be sentence-segmented incorrectly.
We provide a remedy for this for \reframed based on applying our AD splitting model (see Section~\ref{sec:fragmentation}).

\begin{table}[t]
\centering
\resizebox{\textwidth}{!}{%
\begin{tabular}{@{}l@{~}l@{~}r@{\hspace{2mm}}r@{\hspace{2mm}}r@{\hspace{2mm}}r@{\hspace{2mm}}r@{\hspace{2mm}}r@{\hspace{2mm}}r@{\hspace{2mm}}r@{\hspace{2mm}}r@{\hspace{2mm}}r@{\hspace{2mm}}r@{\hspace{2mm}}r@{\hspace{2mm}}r@{\hspace{2mm}}r@{\hspace{2mm}}r@{\hspace{2mm}}r@{\hspace{2mm}}r@{\hspace{2mm}}@{}}
\toprule
& & \multicolumn{5}{c}{\bfseries ASR} & \multicolumn{3}{c}{\bfseries Diarization} & \multicolumn{2}{c}{\bfseries Names} & \multicolumn{3}{c}{\bfseries Times} & \multicolumn{4}{c}{\bfseries Overall} \\
\cmidrule(lr){3-7} \cmidrule(lr){8-10} \cmidrule(lr){11-12} \cmidrule(lr){13-15} \cmidrule(lr){16-19}
& & {{SER}} & {{WER}} & {{SR}} & {{DR}} & {{IR}} & {{DER}} & {{MR}} & {{FAR}} & {{Micro}} & {{Macro}} & {{$0.2$}} & {{$0.4$}} & {{$0.5$}} & {{WER}} & {{SR}} & {{DR}} & {{IR}} \\
\midrule
\textbf{WhisperX} &  \texttt{large-v3-turbo} & 23.5 & 6.6 & 2.9 & 2.7 & 1.0 & 14.9 & 3.7 & 11.3 & 22.3 & 22.2 & 5.1 & 1.0 & 0.8 & 13.7 & 3.6 & 2.4 & 7.7 \\
&  \texttt{gpt-4o-transcribe} & 49.0 & 38.6 & 3.2 & 34.6 & 0.7 & 42.0 & 34.8 & 7.3 & 51.4 & 54.9 & 9.6 & 5.3 & 5.1 & 37.9 & 4.6 & 32.0 & 1.3 \\
& large-v3 & 23.5 & 8.0 & 3.1 & 3.8 & 1.1 & 16.7 & 4.7 & 11.9 & 21.7 & 21.7 & 5.2 & 1.1 & 0.9 & 14.6 & 4.2 & 3.1 & 7.3 \\
& large-v2 & 27.3 & 11.3 & 4.0 & 6.0 & 1.4 & 17.5 & 6.8 & 10.7 & 28.4 & 27.4 & 5.3 & 1.2 & 1.0 & 17.0 & 5.6 & 4.7 & 6.7 \\
& large & 23.5 & 8.0 & 3.1 & 3.8 & 1.1 & 16.7 & 4.7 & 11.9 & 21.7 & 21.7 & 5.2 & 1.1 & 0.9 & 14.6 & 4.2 & 3.1 & 7.3 \\
& medium & 27.4 & 10.5 & 4.2 & 5.1 & 1.3 & 16.7 & 5.9 & 10.8 & 28.1 & 29.2 & 5.2 & 1.1 & 0.9 & 16.3 & 5.8 & 3.9 & 6.7 \\
& medium.en & 25.1 & 7.2 & 3.1 & 2.9 & 1.1 & 14.4 & 4.0 & 10.4 & 23.9 & 23.7 & 5.3 & 1.1 & 1.0 & 14.1 & 3.7 & 2.6 & 7.7 \\
& small.en & 28.1 & 7.6 & 3.8 & 2.7 & 1.2 & 14.3 & 3.3 & 11.0 & 25.3 & 25.0 & 5.1 & 1.0 & 0.8 & 14.6 & 4.3 & 2.5 & 7.8 \\
& small & 29.6 & 8.4 & 4.4 & 2.6 & 1.4 & 13.9 & 3.3 & 10.6 & 27.4 & 28.4 & 5.3 & 1.2 & 1.0 & 15.3 & 5.0 & 2.3 & 7.9 \\
& base.en & 35.1 & 10.1 & 6.1 & 2.2 & 1.8 & 13.8 & 2.7 & 11.1 & 30.7 & 31.8 & 5.0 & 0.9 & 0.8 & 17.2 & 6.6 & 2.1 & 8.6 \\
& base & 40.0 & 12.9 & 7.5 & 3.2 & 2.2 & 14.1 & 3.4 & 10.7 & 32.9 & 35.7 & 5.1 & 1.1 & 0.9 & 19.6 & 8.2 & 2.9 & 8.6 \\
& tiny.en & 43.7 & 14.4 & 8.9 & 2.9 & 2.6 & 13.4 & 3.1 & 10.3 & 36.4 & 38.6 & 5.0 & 0.9 & 0.8 & 21.2 & 9.6 & 2.5 & 9.2 \\
& tiny & 51.5 & 19.1 & 11.7 & 3.6 & 3.8 & 14.4 & 3.8 & 10.7 & 41.1 & 45.0 & 5.1 & 1.1 & 1.0 & 25.6 & 12.4 & 3.1 & 10.2 \\
\textbf{Microsoft} & \texttt{standard} & 32.5 & 7.5 & 4.5 & 1.2 & 1.8 & 20.6 & 0.4 & 20.2 & 25.9 & 26.1 & 2.8 & 0.3 & 0.2 & 17.8 & 4.6 & 0.6 & 12.6 \\
\textbf{Speechmatics} & \texttt{standard} & 25.8 & 5.6 & 3.7 & 0.9 & 0.9 & 3.0 & 0.5 & 2.5 & 22.9 & 24.4 & 1.9 & \textbf{0.0} & \textbf{0.0} & 5.8 & 3.6 & 0.9 & 1.3 \\
&  \texttt{enhanced} & \textbf{18.3} & \textbf{3.6} & \textbf{2.3} & \textbf{0.7} & \textbf{0.6} & \textbf{2.5} & \textbf{0.3} & \textbf{2.2} & \textbf{16.2} & \textbf{17.8} & \textbf{0.7} & \textbf{0.0} & \textbf{0.0} & \textbf{3.6} & \textbf{2.3} & \textbf{0.5} & \textbf{0.8} \\
\bottomrule
\end{tabular}
}
\caption{AD transcription results \emph{without} character-name adaptation (\%, lower better; best in \textbf{bold}). Columns as in Table~\ref{tab:asr}, with additionally a macro-averaged character-name error rate and a $0.5$s timestamp collar. Full per-system version of Table~\ref{tab:asr}; adapted counterpart in Table~\ref{appx:tab:character_guided}.}
\label{appx:tab:standard}
\end{table}

\begin{table}[t]
\centering
\resizebox{\textwidth}{!}{%
\begin{tabular}{@{}l@{~}lr@{\hspace{2mm}}r@{\hspace{2mm}}r@{\hspace{2mm}}r@{\hspace{2mm}}r@{\hspace{2mm}}r@{\hspace{2mm}}r@{\hspace{2mm}}r@{\hspace{2mm}}r@{\hspace{2mm}}r@{\hspace{2mm}}r@{\hspace{2mm}}r@{\hspace{2mm}}r@{\hspace{2mm}}r@{\hspace{2mm}}r@{\hspace{2mm}}r@{\hspace{2mm}}r@{\hspace{2mm}}@{}}
\toprule
& & \multicolumn{5}{c}{\bfseries ASR} & \multicolumn{3}{c}{\bfseries Diarization} & \multicolumn{2}{c}{\bfseries Names} & \multicolumn{3}{c}{\bfseries Times} & \multicolumn{4}{c}{\bfseries Overall} \\
\cmidrule(lr){3-7} \cmidrule(lr){8-10} \cmidrule(lr){11-12} \cmidrule(lr){13-15} \cmidrule(lr){16-19}
& & {{SER}} & {{WER}} & {{SR}} & {{DR}} & {{IR}} & {{DER}} & {{MR}} & {{FAR}} & {{Micro}} & {{Macro}} & {{$0.2$}} & {{$0.4$}} & {{$0.5$}} & {{WER}} & {{SR}} & {{DR}} & {{IR}} \\
\midrule
\textbf{WhisperX} & \texttt{large-v3-turbo} & 17.6 & 5.4 & 1.7 & 2.6 & 1.1 & 14.1 & 3.4 & 10.7 & 7.2 & 9.6 & 5.3 & 1.0 & 0.9 & 12.8 & 2.2 & 2.5 & 8.1 \\
& \texttt{gpt-4o-transcribe} & 34.1 & 22.3 & 2.5 & 18.8 & 1.0 & 30.7 & 19.0 & 11.7 & 30.7 & 32.6 & 8.0 & 3.8 & 3.5 & 24.4 & 4.3 & 16.5 & 3.6 \\
& large-v3 & 18.4 & 8.1 & 1.9 & 5.2 & 1.0 & 16.1 & 5.9 & 10.2 & 8.7 & 11.0 & 5.7 & 1.5 & 1.3 & 14.2 & 2.7 & 4.6 & 6.8 \\
& large-v2 & 20.2 & 10.2 & 2.4 & 6.7 & 1.1 & 18.1 & 7.4 & 10.7 & 12.0 & 14.9 & 5.7 & 1.7 & 1.4 & 15.4 & 3.8 & 5.5 & 6.2 \\
& large & 18.4 & 8.1 & 1.9 & 5.2 & 1.0 & 16.1 & 5.9 & 10.2 & 8.7 & 11.0 & 5.7 & 1.5 & 1.3 & 14.2 & 2.7 & 4.6 & 6.8 \\
& medium & 22.4 & 11.3 & 2.6 & 7.6 & 1.1 & 18.4 & 8.2 & 10.3 & 12.1 & 14.1 & 5.8 & 1.6 & 1.4 & 15.8 & 4.1 & 6.2 & 5.5 \\
& medium.en & 19.5 & 8.1 & 1.7 & 5.4 & 1.0 & 15.9 & 6.2 & 9.7 & 8.3 & 11.3 & 5.7 & 1.6 & 1.4 & 13.6 & 2.4 & 4.8 & 6.3 \\
& small.en & 21.5 & 7.3 & 2.4 & 3.8 & 1.1 & 14.7 & 4.2 & 10.5 & 7.4 & 10.8 & 5.5 & 1.3 & 1.2 & 13.4 & 3.0 & 3.5 & 7.0 \\
& small & 23.5 & 9.0 & 2.8 & 5.0 & 1.2 & 16.0 & 5.7 & 10.3 & 8.8 & 14.5 & 5.6 & 1.4 & 1.2 & 15.1 & 3.6 & 4.5 & 7.0 \\
& base.en & 28.5 & 8.5 & 4.4 & 2.6 & 1.4 & 13.6 & 3.0 & 10.6 & 11.5 & 13.2 & 5.3 & 0.9 & 0.8 & 15.3 & 5.1 & 2.4 & 7.7 \\
& base & 32.0 & 10.0 & 5.5 & 2.6 & 1.8 & 13.6 & 3.0 & 10.6 & 8.9 & 14.8 & 5.1 & 0.9 & 0.8 & 16.9 & 6.3 & 2.3 & 8.4 \\
& tiny.en & 36.8 & 12.3 & 7.2 & 2.8 & 2.4 & 13.8 & 3.0 & 10.8 & 12.4 & 15.0 & 5.2 & 1.0 & 0.8 & 19.1 & 7.9 & 2.4 & 8.8 \\
& tiny & 46.0 & 17.8 & 9.7 & 4.9 & 3.2 & 15.9 & 5.1 & 10.8 & 19.1 & 21.6 & 5.5 & 1.3 & 1.2 & 23.6 & 10.9 & 3.9 & 8.8 \\
\textbf{Speechmatics} & \texttt{standard} & 23.2 & 4.9 & 3.1 & 0.9 & 0.9 & 3.1 & 0.4 & 2.7 & 15.6 & 14.5 & 1.9 & 0.0 & 0.0 & 5.2 & 3.1 & 0.8 & 1.3 \\
& \texttt{enhanced} & \textbf{14.0} & \textbf{2.8} & \textbf{1.5} & \textbf{0.7} & \textbf{0.6} & \textbf{2.7} & \textbf{0.2 }& \textbf{2.5} & \textbf{4.8} & \textbf{7.9} & \textbf{0.8} & \textbf{0.1} & \textbf{0.0} & \textbf{2.9 }& \textbf{1.5} & \textbf{0.5} & \textbf{0.8} \\
\bottomrule
\end{tabular}
}
\caption{AD transcription results \emph{with} character-name adaptation (\%, lower better; best in \textbf{bold}). Columns as in Table~\ref{appx:tab:standard}. Character-name-adapted counterpart of Table~\ref{appx:tab:standard}; adaptation details in Appendix~\ref{appx:sec:asr}.}
\label{appx:tab:character_guided}
\end{table}

\section{Subtitles}\label{appx:sec:subtitles}

Our approach matches movie dialogue ASR with subtitle files available online, and allows for a constant shift between the two.
We collect English SDH and dialogue-only subtitles from \url{opensubtitles.org}.
The data provides many subtitle versions for each movie --- we assign the highest quality score to subtitles with labels `trusted' or `subtranslator' and otherwise scores based on the subtitle's download count.
We use the non-AD transcription from our AD track transcripts.
For each movie, we try candidate subtitles in decreasing order of quality score, accepting the first where at least 90\% of matched words have timestamps within 1 second of the ASR transcript, requiring a minimum of 50 matched timestamp pairs.\footnote{This is inspired by the popular (but now unfortunately deprecated) tool `subsync', as used for the MovieNet dataset \citep{huang-2020-movienet}).}
For each candidate subtitle we try first without applying a constant offset, and if that fails the two conditions, we add the offset that maximises the proportion of inliers.

We apply a data cleaning pipeline to the resulting subtitles, as well as manual review (see also Section~\ref{appx:dataset-review}).
\begin{itemize}
    \item We apply the `fix common errors' tool available in Subtitle Edit v5.0.0, disabling only steps that would alter line breaking or sentence casing.
    \item Next we normalize the text by removing text that concerns renderer state, collapsing repeated spacing and adding spacing after leading hyphens (speaker-change format in subtitles). We correct a recurring error in which the capital letter I has been transcribed as a lowercase l within otherwise fully uppercase segments.
    \item We manually review segments with URLs, and other potential attribute identifiers and remove if it is deemed that these are not part of the original subtitles.
    \item We enforce a minimum inter-segment gap of 24 ms and require strictly positive duration on every segment. Offending segments are first trimmed against the window defined by their neighbours; if trimming yields an empty interval, the segment is re-timed at a target reading rate of 20 characters per second, borrowing up to 200 ms from an adjacent segment longer than 500 ms as a last resort.
\end{itemize}

\section{Dataset Compilation Review Steps}\label{appx:dataset-review}

For the scenes in our dataset, we have reviewed all titles manually and filtered out videos which are not scenes from a final movie (e.g. deleted scenes, trailers, interviews).
We manually determine which speaker label corresponds to the AD narrator.
Based on a manual review of the transcripts, we keep movies with two AD transcripts that appear to be created independently (as opposed to being identical or one being an edit of the other).
Based on manual review of the dialogue and SDH subtitles, we found 3 marked as normal which were in fact SDH and 4 subtitles marked as SDH which were in fact normal; only the remaining are kept.
The US AD tracks are all at the same speed as the videos; the UK AD tracks either require a PAL-NTSC speed-down, or are at the same speed.
We keep movies for which $\geq$80\% and $\geq$ 5 of the videos have the same predicted speed difference, and only keep videos with this predicted speed difference.
We manually remove AD transcription after the start of movie end credits.
Before sending the validation and test split movie scenes with their automatically aligned AD audio tracks to the company performing manual transcription, we reviewed the videos for their alignment.
Out of 106 scenes in the test set, two had imperfect alignment and were discarded; out of 99 in the validation set, three were discarded.
For these evaluation scenes, we also remove those with any kind of advertisement (typically a couple seconds at the end promoting the physical release); 4/106 in the test set and none in the validation set.
Finally, for these scenes we manually reviewed scenes from the same movie that had any temporal overlap and discard scenes until this is not the case; 9/106 from the test set and none from the validation set.
The final test set has 91 scenes and the final validation set has 96 scenes.

\section{Dataset Splits}\label{sec:splitting}

\begin{figure}
    \centering
    \begin{tikzpicture}
  \begin{axis}[
      width = \linewidth,
      height = 8cm,
      major x tick style = transparent,
      ybar=2*\pgflinewidth,
      bar width=3pt,
      ymajorgrids = true,
      ylabel = {Proportion},
      symbolic x coords={Action,Thriller,Comedy,Drama,Adventure,Sci-Fi,Crime,Horror,Fantasy,Mystery,Romance,Biography,Family,Music,Musical,History,Sport,War,Western},
      xtick = data,
      ymin=0, ymax=0.16,
      scaled y ticks = false,
      enlarge x limits=0.05,
      yticklabel style={/pgf/number format/fixed, font=\small},
      ymin=0,
      grid style=dashed,
      ylabel style={font=\small},
      xlabel style={font=\small},
      xticklabel style={font=\small},
      legend cell align=left,
      x tick label style={rotate=45, anchor=east, align=right, font=\footnotesize},
      axis x line*=bottom,
      axis y line*=left,
      axis line style={->, >=latex},
      legend pos=north east,
      legend style={
          font=\small,
          cells={anchor=west},
          opacity=0.7,
        },
    ]
    \addplot[style={draw=reframed-green,fill=reframed-green,mark=none}]
    coordinates {(Action,0.1263) (Thriller,0.1253) (Comedy,0.1189) (Drama,0.1178) (Adventure,0.0892) (Sci-Fi,0.0701) (Crime,0.0626) (Horror,0.0520) (Fantasy,0.0499) (Mystery,0.0488) (Romance,0.0456) (Biography,0.0223) (Family,0.0180) (Music,0.0149) (Musical,0.0106) (History,0.0096) (Sport,0.0074) (War,0.0064) (Western,0.0042)};
    \addplot[style={draw=reframed-silver,fill=reframed-silver,mark=none}]
    coordinates {(Action,0.1481) (Thriller,0.1111) (Comedy,0.1111) (Drama,0.1111) (Adventure,0.0741) (Sci-Fi,0.1111) (Crime,0.0741) (Horror,0.0370) (Fantasy,0.0370) (Mystery,0.0741) (Romance,0.0741) (Biography,0.0370) (Family,0.0000) (Music,0.0000) (Musical,0.0000) (History,0.0000) (Sport,0.0000) (War,0.0000) (Western,0.0000)};
    \addplot[style={draw=reframed-gold,fill=reframed-gold,mark=none}]
    coordinates {(Action,0.1389) (Thriller,0.1389) (Comedy,0.1111) (Drama,0.1111) (Adventure,0.1111) (Sci-Fi,0.0833) (Crime,0.0556) (Horror,0.0556) (Fantasy,0.0556) (Mystery,0.0556) (Romance,0.0556) (Biography,0.0000) (Family,0.0000) (Music,0.0000) (Musical,0.0000) (History,0.0000) (Sport,0.0278) (War,0.0000) (Western,0.0000)};
    \legend{Train,Val,Test}
  \end{axis}
\end{tikzpicture}
    \caption{The final distribution over genres in each of train, validation and test splits in \protect\reframed. We have applied an approach that maximises the distributional similarity over genres across the three splits.}
    \label{fig:stratified_split}
\end{figure}
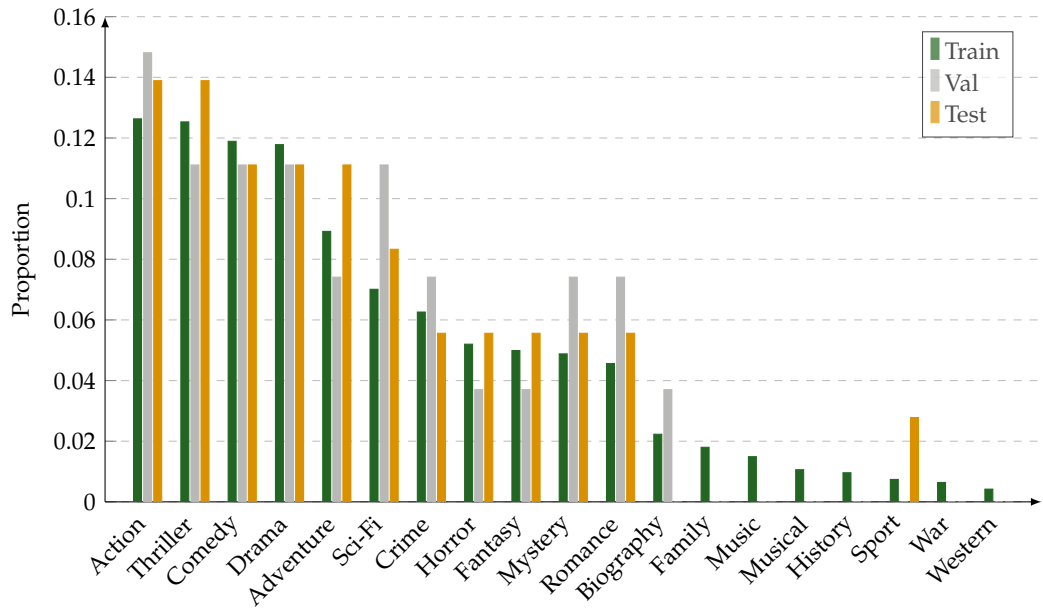

Genre labels are obtained from IMDb, where each film is assigned one or more genre descriptors.
Our partitioning into train, validation and test splits employs a distributional stratification strategy to align the label frequencies of the validation and test sets with the training set.
For each evaluation set, 10 movies are selected from the pool of candidate movies by minimizing the Jensen-Shannon divergence between the samples and the train genre distribution.
Acceptable solutions may have zero representation of a genre that is represented in the train set, hence our choice of Jensen-Shannon divergence.
Samples not selected for test are returned to the train set; we do not do this for validation, as these were movies used in prior datasets (LSMDC, MAD and CMD-AD).
This is formalized in Algorithm~\ref{genre-algorithm} and the final genre split is visualized in Figure~\ref{fig:stratified_split}.

\begin{algorithm}
\caption{Genre stratification via Jensen-Shannon divergence}
\label{genre-algorithm}
\begin{algorithmic}[1]
\State \textbf{Input:} Initial partitions $D_{\text{train}}, D_{\text{pool\_val}}, D_{\text{pool\_test}}$; desired subset size $k$
\State \textbf{Output:} Final splits $S_{\text{train}}, S_{\text{val}}, S_{\text{test}}$
\State $\mathcal{L} \leftarrow \text{unique labels in } D_{\text{train}}$ \hfill $\triangleright$ \text{Target label vocabulary}
\State $P \leftarrow \text{Normalize} \sum_{i \in D_{\text{train}}} \text{OneHot}_{G_i, \mathcal{L}}$ \hfill $\triangleright$ \text{Target genre distribution}
\Function{FindBest}{$D_{\text{pool}}, P, k$}
    \State $\delta_{\text{min}} \leftarrow \infty$
    \For{subset $s \in \text{Combinations}_{D_{\text{pool}}, k}$} \hfill $\triangleright$ \text{Iterate through all possible subsets}
        \State $Q \leftarrow \text{Normalize} \sum_{j \in s} \text{OneHot}_{G_j, \mathcal{L}}$ \hfill $\triangleright$ \text{Candidate subset distribution}
        \State $M \leftarrow 0.5 P + 0.5 Q$ \hfill $\triangleright$ \text{Mixture distribution}
        \State $\text{kl}_{\text{target}} \leftarrow 0.5 \sum P_i \log \frac{P_i}{M_i}$ \hfill $\triangleright$ \text{Target divergence from mixture distribution}
        \State $\text{kl}_{\text{subset}} \leftarrow 0.5 \sum Q_i \log \frac{Q_i}{M_i}$ \hfill $\triangleright$ \text{Subset divergence from mixture distribution}
        \State $\text{js} \leftarrow \text{kl}_{\text{target}} + \text{kl}_{\text{subset}}$ \hfill $\triangleright$ \text{Total symmetric Jensen-Shannon distance}
        \If{$\text{js} < \delta_{\text{min}}$}
        \State $\delta_{\text{min}} \leftarrow \text{js}, s^{*} \leftarrow s$ \hfill $\triangleright$ \text{Keep subset that best approximates target}
        \EndIf
    \EndFor
    \State \Return $s^{*}$
\EndFunction
\State $S_{\text{val}} \leftarrow \text{FindBest}_{D_{\text{pool\_val}}, P, k}$
\State $S_{\text{test}} \leftarrow \text{FindBest}_{D_{\text{pool\_test}}, P, k}$
\State $S_{\text{train}} \leftarrow D_{\text{train}} \cup D_{\text{pool\_test}} \setminus S_{\text{test}}$ \hfill $\triangleright$
\end{algorithmic}
\end{algorithm}

\section{Face Detection Labeling}\label{appx:faces}

Every character mention in both of the two references of the test split scenes was manually labeled without use of automatic tools.
If a character is named differently in the two references, we selected one.\footnote{There were six cases of this (three first name vs. surname, three acronyms/nicknames).}
We then run an off-the-shelf face-detection tool\footnote{\url{https://github.com/ageitgey/face_recognition}} on frames sampled at one frame per second on the input video.
The first discernable face corresponding to each character in each video was annotated.
Of 315 character-video pairs, 268 were successfully labeled.
The remaining concern cases when the character's face is never seen directly in the video (e.g. only from behind), the character is seen only for less than 1 second at a time in the video, or the face detection tool failed to recognize the face.
Of these unlabeled cases, 68.1\% have a labeled face from another video from the same movie which future work could use.
We maintain our closed-loop design and do not prompt with the names of unlabeled characters.

\section{Example US AD, UK AD and Screenplay Scene}\label{appx:intro-example}

\vspace{1em}
\noindent
\begin{minipage}[t]{0.48\textwidth}
{\small
\textbf{US AD}\label{appx:ex:dragon-tattoo}
\begin{enumerate}[label=\alph*., nosep, leftmargin=*]
    \item Night-time.
    \item Snow drifts to the ground
    \item as Salander glides her motorcycle down a hill
    \item and parks across the street
    \item from Blomkvist's apartment building.
    \item Dismounting, she removes her helmet,
    \item then takes the garment bag with her gift
    \item off the bike's rear cargo rack.
    \item She sees Blomkvist and Erika leaving the building
    \item and watches unnoticed as he drapes
    \item an arm over her shoulders.
    \item The couple heads down a walkway
    \item to a waiting taxi at the top of a hill.
    \item Salander watches woundedly
    \item as they climb into the cab,
    \item then lowers her heartbroken gaze.
    \item Turning away, she tosses Blomkvist's gift
    \item into a dumpster against the side of a building.
    \item She glances at the cab as it drives off,
    \item then dons her helmet
    \item and mounts her bike.
    \item She speeds off down the cobblestone street,
    \item her figure disappearing from view as she rounds a corner.
\end{enumerate}
}
\end{minipage}
\hfill
\begin{minipage}[t]{0.48\textwidth}
{\small
\textbf{UK AD}\label{appx:ex:dragon-tattoo-uk}
\begin{enumerate}[label=\alph*., nosep, leftmargin=*]
    \item Night.
    \item Lisbeth rides her motorcycle down a cobbled street
    \item with snow piled at the curbside.
    \item She parks on a corner outside Mikael's apartment
    \item and removes her crash helmet.
    \item She takes a package from the back of the bike,
    \item but stops dead when she sees Mikael
    \item leave his apartment with Erika.
    \item She watches numbly as the couple walk away
    \item with their arms around each other.
    \item Lisbeth looks down as Mikael and Erika
    \item get into a waiting taxi.
    \item Lisbeth turns away and flings the package which has the card
    \item attached to it into a nearby dumpster.
    \item She puts her crash helmet back on,
    \item gets on her bike
    \item and starts it up.
    \item A solitary figure, she rides off into the night.
\end{enumerate}
}
\end{minipage}

\lstdefinestyle{xmlstyle}{
    basicstyle=\small\ttfamily,
    backgroundcolor=\color{reframed-gray!10},
    frame=single,
    framerule=0.5pt,
    rulecolor=\color{reframed-silver},
    breaklines=true,
    breakatwhitespace=true,
    xleftmargin=8pt,
    xrightmargin=8pt,
    showstringspaces=false,
    columns=flexible,
    morestring=[b]{>},
    morestring=[b]{<},
    stringstyle=\color{black},
}

\

\begin{lstlisting}[style=xmlstyle]
<scene>
<stage_direction>EXT. STOCKHOLM - EVENING</stage_direction>
<scene_description>Snow's falling, but unlike the brutal winter storms last year, it's just heavy enough to dust the city in white powder that reflects the Christmas lights in the trees. The beauty of her city pleases Salander as she rides through it on her motorcycle. Or maybe it's something she's feeling as she turns onto Blomkvist's street. She parks. Gathers the garment bag and card, walks under construction scaffolding toward his apartment building. From around the corner ahead, Blomkvist and Erika appear. He says something and she laughs, puts her arm around his waist and lays her head against his neck scarf. Salander stops mid-stride, ducks into an alcove under the scaffolding, waits as long as it should take them to get inside the building, then peers out again - They're not inside. They've stopped just outside the building's entrance. They're embracing. And the longer they stand there together, the sharper the pain stabs at Salander. Finally they separate, but only to allow Blomkvist to unlock the door. Erika takes his hand then, and they disappear inside. Salander can't move. Waits for the pain to subside - which it eventually does - but only to be replaced with a sense of helplessness. Then that subsides, replaced with a facade of insouciance. The richest girl in Sweden emerges from the alcove and walks back the way she came - tossing the garment bag and card into a construction dumpster - climbs onto her Honda - starts it - and drives off - Probably forever.</scene_description>
</scene>
\end{lstlisting}

\section{Further Details on Evaluation Metrics}

For the test set, evaluation is performed at the level of each video; for the challenge set, at the sequence-level.
Movie-level scores are the mean of the video or sequence scores, and the final metrics are a macro-average across all movies.
References are already segmented into description elements; generations should also be defined to be at the level of individual visual description elements.

As evaluation preprocessing, we segment all generations into sentences with the \texttt{sat-12l-sm} model from \citet{frohmann-2024-segment}.
We then further segment each resulting sentence into description elements using our learned AD splitting model.
We re-purpose the splitting modeling used for the training set of \reframed: re-training directly on the gold AD boundaries (as opposed to boundaries mapped onto the Speechmatics transcripts).
To enhance the robustness of this splitting model, we apply the model initialization and data augmentation that \citet{frohmann-2024-segment} used for \texttt{sat-12l-sm} training.
Cross-validation on the gold data shows that this trained model performs similarly to the learned splitting model used for the \reframed dataset ($F_1$=66.0, vs. a comma-splitting baseline $F_1$=38.4).

\subsection{Alignment-based Evaluation}\label{sub:soda_alignment}

SODA computes an optimal monotonic alignment between generated and reference descriptions, using METEOR as the pairwise content score. The alignment preserves temporal order and allows unmatched generated or reference descriptions.
Sorted by start time, let us denote the sequence of $n$ reference description elements as $\mathcal{D} = (d_i)_{i=1}^n$ and the sequence of $\hat{n}$ generated description elements as $\hat{\mathcal{D}} = (\hat{d}_j)_{j=1}^{\hat{n}}$.
Using the \textrm{METEOR} evaluation metric, we then define the following:
\[
M(i, j) = \mathrm{METEOR}(d_i, \hat{d}_j).
\]
Let $S(i, j)$ be the best alignment score between the first $i$ reference and first $j$ generated descriptions, computed recursively as:
\[
S(i, j) =
\max
\begin{cases}
S(i-1, j)\\
S(i, j-1) \\
S(i-1, j-1)+M(i, j)
\end{cases}
\]
with $S(i, 0)=S(0, j)=0$ and ties broken in favor of the final case.

Backtracking yields the optimal alignment $\mathcal{A}^\star\subseteq [|\mathcal{D}|]\times[|\hat{\mathcal{D}}|]$. We then compute SODA-M as the F1 score over aligned METEOR content:
\[
P = \frac{\sum_{(i,j)\in\mathcal{A}^\star}
\mathrm{METEOR}(d_i,\hat{d}_j)}{|\hat{\mathcal{D}}|}, \quad
R = \frac{\sum_{(i,j)\in\mathcal{A}^\star}
\mathrm{METEOR}(d_i,\hat{d}_j)}{|\mathcal{D}|},
\]
\[
\mathrm{SODA\text{-}M} = \frac{2PR}{P+R} = \frac{\sum_{(i,j)\in\mathcal{A}^\star}
\mathrm{METEOR}(d_i,\hat{d}_j)}{\tfrac{1}{2}\left(|\mathcal{D}|+|\hat{\mathcal{D}}|\right)}
\]
For temporal grounding, SODA-T measures whether aligned descriptions are placed close in time.
Let $m(I_i)$ denote the midpoint of the temporal interval $I_i $ of description element $d_i$, and $\tau$ the temporal tolerance:
\[
\mathrm{SODA\text{-}T}
=
\frac{1}{|\mathcal{D}|}
\sum_{(i,j)\in\mathcal{A}^\star}
\mathbf{1}\bigl[|m(I_i)-m(\hat{I}_j)|<\tau\bigr].
\]

Following prior SODA implementations, we take the maximum of each score across references.

\subsection{QA Evaluation}\label{sec:qa-prompts}

We prompt an LLM with each segment of AD and 1 minute transcript of contextual AD (segments that end after 30 seconds before the target segment and end before 30 seconds after the target segment), and instruct the model to generate up to 3 questions and multiple choice answers of what is described.
The prompt we used is as follows:

\begin{tcolorbox}[colback=white, colframe=reframed-silver, sharp corners,
  boxrule=0.5mm, breakable, title=Question Generation Prompt]
\begin{Verbatim}[fontsize=\small, breaklines=true, breakanywhere=true, breaksymbolleft={}]
You are provided a transcript of one or more English audio description sentences from a 1-minute section of a movie. Based on one segment of text contained within the transcript, your task is to create 0, 1, 2, or 3 English questions and multiple-choice answers.

These questions will later be used for an evaluation task where the respondent has access to several minutes of a different audio description transcript of the same section of the same movie. The respondent will not be told which segment your question refers to.

Guidelines that your questions must follow:

    1. The question must include sufficient detail to uniquely identify the specific aspect under question from the wider context. If the aspect under question is strongly related to something also described elsewhere, or something that plausibly could happen nearby in the movie, do not create a question about it.

    2. The question must ask directly about the movie world (never use meta-phrasing such as 'described in', 'segment', 'scene', 'audio description').

    3. The question must concern content addressed in the provided segment of interest. You may use the full transcript for context, and it is acceptable for your question to only be answerable given context.

    4. For "who" questions about characters, always use the pronouns they/them/their. Otherwise, if referring to one or more characters, the question must always use character names. If a pronoun cannot be resolved to a character's name given the transcript context, do not create a question about it.

    5. The question must be phrased in such a way that there can feasibly be 5 possible answers, only one of which will be correct.

Guidelines that your correct_answer must follow:

    1. The answer should use wording similar to the segment of interest. If needed, you may apply simple reformulation so that the correct_answer is consistent with all the wrong_answers.
    
    2. The answer should have a length of between 1-4 words. The number of words in an answer is equal to the number of spaces plus one.
    
    3. If referring to one or more characters, the answer must always use the character's names. If a pronoun cannot be resolved to a character's name given the transcript context, do not create a question about it.

Guidelines that your wrong_answers must follow:

    1. There must be exactly four wrong answers.
    
    2. Wrong answers should be between 1-4 words and of similar style as the correct answer. If the correct_answer uses a particular article, every wrong answer should use it too. The correct answer should not stand out because of a surface feature.
    
    3. Wrong answers should all be both plausible and absolutely wrong. You should create them using your general knowledge. The correct answer should not stand out because by guessing it is more plausible than your wrong_answers.

Before your final response, check that all of the above conditions are met. If there are no suitable questions, it is fully acceptable to respond with zero questions. If only 1 or 2 questions are possible, only respond with this many questions. There is no penalty for creating fewer than 3 questions.

Your response must be a valid JSON list containing up to three objects. Each of these objects must have three keys: "question" mapped to a string of your question, "correct_answer" mapped to a string of your correct answer, and "wrong_answers" mapped to an array of exactly four strings of your wrong answers.

Example

Full transcript: Bright light fills our view and the caretaker's tea kettle steams. In a bedroom, Harry Potter stirs from a restless sleep. He gropes for his glasses. She goes to Ron. Harry rubs the jagged scar on his forehead. Later, the kids walk through a forest with Ron's father, his sister Ginny; and twin brothers Fred and George. A stout, smiling man hikes toward them.

Segment of interest: In a bedroom, Harry Potter stirs from a restless sleep.

[{"question": "Who rubs their forehead?", "correct_answer": "Harry", "wrong_answers": ["the caretaker", "Ron", "Ginny", "Fred"]}, {"question": "What shape is the scar on Harry's forehead?", "correct_answer": "jagged", "wrong_answers": ["circular", "crescent", "oval", "straight"]}, {"question": "Where on Harry's body is there a jagged scar?", "correct_answer": "forehead", "wrong_answers": ["cheek", "neck", "arm", "calf"]}]

Your task

Full transcript: INSERT_FULL_TRANSCRIPT

Segment of interest: INSERT_SEGMENT_OF_INTEREST
\end{Verbatim}
\end{tcolorbox}

We filtered the resulting questions aggressively, keeping those that: pass QEval on both reference scripts, fail QEval on the same script without the segment on which the question was based, pass QEval-T on both reference scripts.
Table~\ref{tab:filtering-stages} gives a summary of our filtering stages and how many QA pairs remain after each stage.

\begin{table}[ht]
\centering
\begin{small}
\begin{tabular}{cp{6cm}rr}
\toprule
{\bf Stage} & {\bf description} & {\bf test QAs} & {\bf challenge QAs} \\
\midrule
0 & All generated questions & 14,981 & 58,681 \\
1 & Pass QEval on the source reference & 13,737 & 53,704 \\
2 & Fail QEval on the source reference without the source segment & 5,042 & 20,829 \\
3 & Pass QEval on the alternative reference & 2,182 & 8,335 \\
4 & Pass QEval-T on the source reference & 1,987 & 7,715 \\
5 & Pass QEval-T on the alternative reference & 1,517 & 5,560 \\
\bottomrule
\end{tabular}
\end{small}
\caption{Number of QA pairs remaining after each filtering stage.}
\label{tab:filtering-stages}
\end{table}

QA answering is performed with a base LLM using a basic prompt template identical to the multiple-choice data used in the model's mid-training.\footnote{%
  \{\texttt{doc}\}%
  \textbackslash\texttt{nQuestion: }%
  \{\texttt{question}\}%
  \textbackslash\texttt{nA. }%
  \{\texttt{answer\_a}\}%
  \textbackslash\texttt{n...}%
  \textbackslash\texttt{nE. }%
  \{\texttt{answer\_e}\}%
  \textbackslash\texttt{nAnswer:}}
We compare the resulting space-prefixed A-E logits.

We access this LLM evaluator's 6T token training corpus for contamination of the textual AD using the infini-gram tool \citep{liu-2024-infinigram}.
Results show zero exact matches against AD segments containing greater than 10 tokens, leading us to determine the risk of contamination is low (Figure~\ref{fig:contamination_ad}; Figure \ref{fig:contamination_subs} provides a reference analysis of dialogue subtitles).

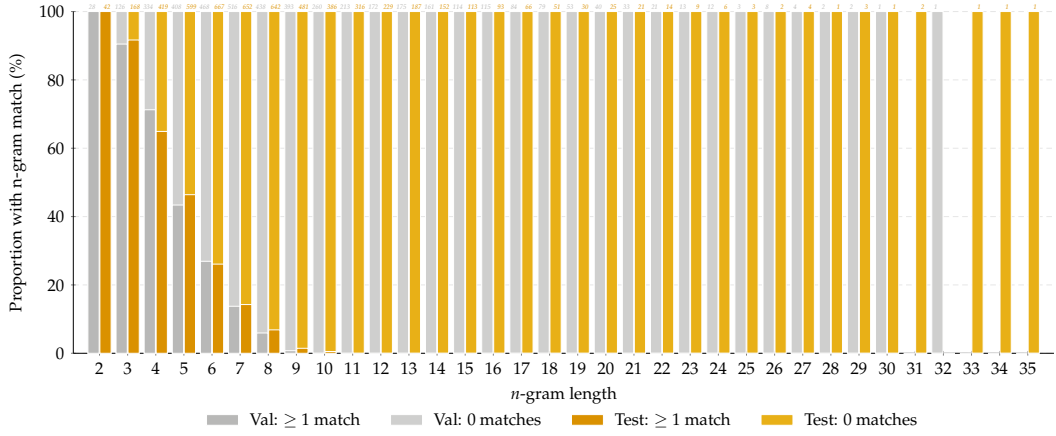
\begin{figure}[t]
    \centering
    \resizebox{\textwidth}{!}{%
  \begin{tikzpicture}
    \begin{axis}[
        width=20cm,
        height=8cm,
        ybar=1*\pgflinewidth,
        bar width=6pt,
        axis lines=left,
        every outer x axis line/.append style={-},
        every outer y axis line/.append style={-},
        clip=false,
        ymajorgrids=true,
        grid style={thin, gray!20, densely dashed},
        ymin=0, ymax=100,
        ytick={0,20,40,60,80,100},
        ylabel={Proportion with n-gram match (\%)},
        xlabel={$n$-gram length},
        xtick=data,
        xticklabel style={font=\footnotesize},
        yticklabel style={font=\footnotesize},
        ylabel style={font=\small},
        xlabel style={font=\small},
        tick style={color=black!50, thin},
        major tick length=2.5pt,
        legend style={
            at={(0.5,-0.15)},
            anchor=north,
            legend columns=4,
            font=\footnotesize,
            draw=none,
            column sep=4pt,
            /tikz/every even column/.append style={column sep=14pt},
          },
        enlarge x limits={abs=14pt},
      ]

      \addplot [bar shift=-3.2pt, fill=reframed-gray, draw=white, line width=0.3pt, mark=none, forget plot]
      coordinates {
          (2,100.00)
          (3,100.00)
          (4,100.00)
          (5,100.00)
          (6,100.00)
          (7,100.00)
          (8,100.00)
          (9,100.00)
          (10,100.00)
          (11,100.00)
          (12,100.00)
          (13,100.00)
          (14,100.00)
          (15,100.00)
          (16,100.00)
          (17,100.00)
          (18,100.00)
          (19,100.00)
          (20,100.00)
          (21,100.00)
          (22,100.00)
          (23,100.00)
          (24,100.00)
          (25,100.00)
          (26,100.00)
          (27,100.00)
          (28,100.00)
          (29,100.00)
          (30,100.00)
          (31,0.00)
          (32,100.00)
          (33,0.00)
          (34,0.00)
          (35,0.00)
        };

      \addplot [bar shift=3.2pt, fill=reframed-yellow, draw=white, line width=0.3pt, mark=none,forget plot]
      coordinates {
          (2,100.00)
          (3,100.00)
          (4,100.00)
          (5,100.00)
          (6,100.00)
          (7,100.00)
          (8,100.00)
          (9,100.00)
          (10,100.00)
          (11,100.00)
          (12,100.00)
          (13,100.00)
          (14,100.00)
          (15,100.00)
          (16,100.00)
          (17,100.00)
          (18,100.00)
          (19,100.00)
          (20,100.00)
          (21,100.00)
          (22,100.00)
          (23,100.00)
          (24,100.00)
          (25,100.00)
          (26,100.00)
          (27,100.00)
          (28,100.00)
          (29,100.00)
          (30,100.00)
          (31,100.00)
          (32,0.00)
          (33,100.00)
          (34,100.00)
          (35,100.00)
        };

      \addplot [bar shift=-3.2pt, fill=reframed-silver, draw=white, line width=0.3pt,
        mark=none, forget plot]
      coordinates {
          (2,100.00)
          (3,90.48)
          (4,71.26)
          (5,43.38)
          (6,26.92)
          (7,13.76)
          (8,5.94)
          (9,0.76)
          (10,0.00)
          (11,0.00)
          (12,0.00)
          (13,0.00)
          (14,0.00)
          (15,0.00)
          (16,0.00)
          (17,0.00)
          (18,0.00)
          (19,0.00)
          (20,0.00)
          (21,0.00)
          (22,0.00)
          (23,0.00)
          (24,0.00)
          (25,0.00)
          (26,0.00)
          (27,0.00)
          (28,0.00)
          (29,0.00)
          (30,0.00)
          (31,0.00)
          (32,0.00)
          (33,0.00)
          (34,0.00)
          (35,0.00)
        };

      \addplot [bar shift=3.2pt, fill=reframed-gold, draw=white, line width=0.3pt,
        mark=none, forget plot]
      coordinates {
          (2,100.00)
          (3,91.67)
          (4,64.92)
          (5,46.41)
          (6,26.09)
          (7,14.26)
          (8,6.85)
          (9,1.46)
          (10,0.52)
          (11,0.00)
          (12,0.00)
          (13,0.00)
          (14,0.00)
          (15,0.00)
          (16,0.00)
          (17,0.00)
          (18,0.00)
          (19,0.00)
          (20,0.00)
          (21,0.00)
          (22,0.00)
          (23,0.00)
          (24,0.00)
          (25,0.00)
          (26,0.00)
          (27,0.00)
          (28,0.00)
          (29,0.00)
          (30,0.00)
          (31,0.00)
          (32,0.00)
          (33,0.00)
          (34,0.00)
          (35,0.00)
        };

      \addplot [draw=none, fill=none, mark=none, forget plot,
        nodes near coords, point meta=explicit symbolic,
        every node near coord/.style={
            rotate=0,
            anchor=south,
            xshift=-4pt,
            yshift=1pt,
            font=\fontsize{3.5}{4}\selectfont\itshape,
            inner sep=0pt,
          }]
      coordinates {
          (2,100.00)[{\textcolor{reframed-silver}{28}}]
          (3,100.00)[{\textcolor{reframed-silver}{126}}]
          (4,100.00)[{\textcolor{reframed-silver}{334}}]
          (5,100.00)[{\textcolor{reframed-silver}{408}}]
          (6,100.00)[{\textcolor{reframed-silver}{468}}]
          (7,100.00)[{\textcolor{reframed-silver}{516}}]
          (8,100.00)[{\textcolor{reframed-silver}{438}}]
          (9,100.00)[{\textcolor{reframed-silver}{393}}]
          (10,100.00)[{\textcolor{reframed-silver}{260}}]
          (11,100.00)[{\textcolor{reframed-silver}{213}}]
          (12,100.00)[{\textcolor{reframed-silver}{172}}]
          (13,100.00)[{\textcolor{reframed-silver}{175}}]
          (14,100.00)[{\textcolor{reframed-silver}{161}}]
          (15,100.00)[{\textcolor{reframed-silver}{114}}]
          (16,100.00)[{\textcolor{reframed-silver}{115}}]
          (17,100.00)[{\textcolor{reframed-silver}{84}}]
          (18,100.00)[{\textcolor{reframed-silver}{79}}]
          (19,100.00)[{\textcolor{reframed-silver}{53}}]
          (20,100.00)[{\textcolor{reframed-silver}{40}}]
          (21,100.00)[{\textcolor{reframed-silver}{33}}]
          (22,100.00)[{\textcolor{reframed-silver}{21}}]
          (23,100.00)[{\textcolor{reframed-silver}{13}}]
          (24,100.00)[{\textcolor{reframed-silver}{12}}]
          (25,100.00)[{\textcolor{reframed-silver}{3}}]
          (26,100.00)[{\textcolor{reframed-silver}{8}}]
          (27,100.00)[{\textcolor{reframed-silver}{4}}]
          (28,100.00)[{\textcolor{reframed-silver}{2}}]
          (29,100.00)[{\textcolor{reframed-silver}{2}}]
          (30,100.00)[{\textcolor{reframed-silver}{1}}]
          (32,100.00)[{\textcolor{reframed-silver}{1}}]
        };

      \addplot [draw=none, fill=none, mark=none, forget plot,
        nodes near coords, point meta=explicit symbolic,
        every node near coord/.style={
            rotate=0,
            anchor=south,
            xshift=4pt,
            yshift=1pt,
            font=\fontsize{3.5}{4}\selectfont\itshape,
            inner sep=0pt,
          }]
      coordinates {
          (2,100.00)[{\textcolor{reframed-gold}{42}}]
          (3,100.00)[{\textcolor{reframed-gold}{168}}]
          (4,100.00)[{\textcolor{reframed-gold}{419}}]
          (5,100.00)[{\textcolor{reframed-gold}{599}}]
          (6,100.00)[{\textcolor{reframed-gold}{667}}]
          (7,100.00)[{\textcolor{reframed-gold}{652}}]
          (8,100.00)[{\textcolor{reframed-gold}{642}}]
          (9,100.00)[{\textcolor{reframed-gold}{481}}]
          (10,100.00)[{\textcolor{reframed-gold}{386}}]
          (11,100.00)[{\textcolor{reframed-gold}{316}}]
          (12,100.00)[{\textcolor{reframed-gold}{229}}]
          (13,100.00)[{\textcolor{reframed-gold}{187}}]
          (14,100.00)[{\textcolor{reframed-gold}{152}}]
          (15,100.00)[{\textcolor{reframed-gold}{113}}]
          (16,100.00)[{\textcolor{reframed-gold}{93}}]
          (17,100.00)[{\textcolor{reframed-gold}{66}}]
          (18,100.00)[{\textcolor{reframed-gold}{51}}]
          (19,100.00)[{\textcolor{reframed-gold}{30}}]
          (20,100.00)[{\textcolor{reframed-gold}{25}}]
          (21,100.00)[{\textcolor{reframed-gold}{21}}]
          (22,100.00)[{\textcolor{reframed-gold}{14}}]
          (23,100.00)[{\textcolor{reframed-gold}{9}}]
          (24,100.00)[{\textcolor{reframed-gold}{6}}]
          (25,100.00)[{\textcolor{reframed-gold}{3}}]
          (26,100.00)[{\textcolor{reframed-gold}{2}}]
          (27,100.00)[{\textcolor{reframed-gold}{4}}]
          (28,100.00)[{\textcolor{reframed-gold}{1}}]
          (29,100.00)[{\textcolor{reframed-gold}{3}}]
          (30,100.00)[{\textcolor{reframed-gold}{1}}]
          (31,100.00)[{\textcolor{reframed-gold}{2}}]
          (33,100.00)[{\textcolor{reframed-gold}{1}}]
          (34,100.00)[{\textcolor{reframed-gold}{1}}]
          (35,100.00)[{\textcolor{reframed-gold}{1}}]
        };

      \addlegendimage{area legend, fill=reframed-silver, draw=white, line width=0.3pt}
      \addlegendentry{Val: $\geq$ 1 match}
      \addlegendimage{area legend, fill=reframed-gray, draw=white, line width=0.3pt}
      \addlegendentry{Val: 0 matches}
      \addlegendimage{area legend, fill=reframed-gold, draw=white, line width=0.3pt}
      \addlegendentry{Test: $\geq$ 1 match}
      \addlegendimage{area legend, fill=reframed-yellow, draw=white, line width=0.3pt}
      \addlegendentry{Test: 0 matches}

    \end{axis}
  \end{tikzpicture}
}
    \caption{Proportion of \emph{AD} segments with at least one exact n-gram match in the LLM evaluator's training corpus across validation and test splits. The numerical values above each bar indicate the absolute count of segments evaluated at that specific length. The distribution reaches zero matches for segments of length greater than 10 tokens, which is below the 13-gram threshold set by \citet{brown-2020-few}, suggesting very low contamination of our AD evaluation data.}
    \label{fig:contamination_ad}
\end{figure}
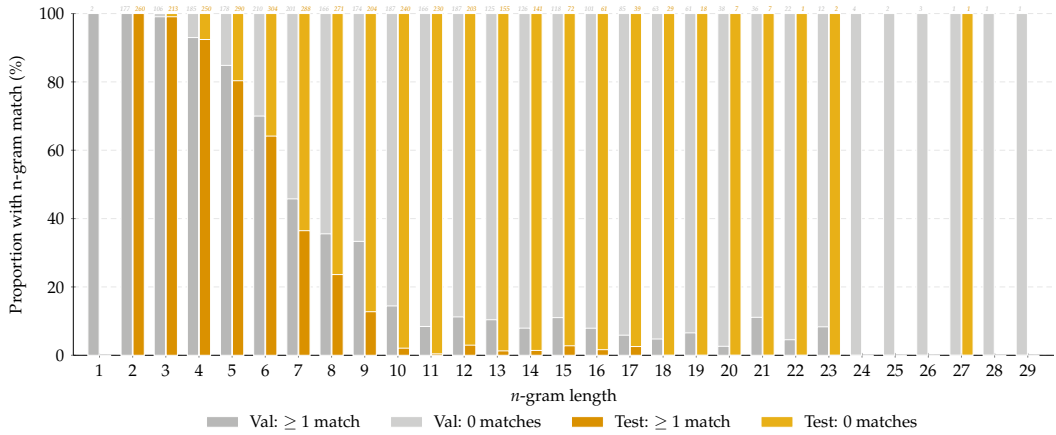
\begin{figure}
    \centering
    \resizebox{\textwidth}{!}{%
  \begin{tikzpicture}
    \begin{axis}[
        width=20cm,
        height=8cm,
        ybar=1*\pgflinewidth,
        bar width=6pt,
        axis lines=left,
        every outer x axis line/.append style={-},
        every outer y axis line/.append style={-},
        clip=false,
        ymajorgrids=true,
        grid style={thin, gray!20, densely dashed},
        ymin=0, ymax=100,
        ytick={0,20,40,60,80,100},
        ylabel={Proportion with n-gram match (\%)},
        xlabel={$n$-gram length},
        xtick=data,
        xticklabel style={font=\footnotesize},
        yticklabel style={font=\footnotesize},
        ylabel style={font=\small},
        xlabel style={font=\small},
        tick style={color=black!50, thin},
        major tick length=2.5pt,
        legend style={
            at={(0.5,-0.15)},
            anchor=north,
            legend columns=4,
            font=\footnotesize,
            draw=none,
            column sep=4pt,
            /tikz/every even column/.append style={column sep=14pt},
          },
        enlarge x limits={abs=14pt},
      ]

      \addplot [bar shift=-3.2pt, fill=reframed-gray, draw=white, line width=0.3pt, mark=none, forget plot]
      coordinates {
          (1,100.00)
          (2,100.00)
          (3,100.00)
          (4,100.00)
          (5,100.00)
          (6,100.00)
          (7,100.00)
          (8,100.00)
          (9,100.00)
          (10,100.00)
          (11,100.00)
          (12,100.00)
          (13,100.00)
          (14,100.00)
          (15,100.00)
          (16,100.00)
          (17,100.00)
          (18,100.00)
          (19,100.00)
          (20,100.00)
          (21,100.00)
          (22,100.00)
          (23,100.00)
          (24,100.00)
          (25,100.00)
          (26,100.00)
          (27,100.00)
          (28,100.00)
          (29,100.00)
        };

      \addplot [bar shift=3.2pt, fill=reframed-yellow, draw=white, line width=0.3pt, mark=none, forget plot]
      coordinates {
          (1,0.00)
          (2,100.00)
          (3,100.00)
          (4,100.00)
          (5,100.00)
          (6,100.00)
          (7,100.00)
          (8,100.00)
          (9,100.00)
          (10,100.00)
          (11,100.00)
          (12,100.00)
          (13,100.00)
          (14,100.00)
          (15,100.00)
          (16,100.00)
          (17,100.00)
          (18,100.00)
          (19,100.00)
          (20,100.00)
          (21,100.00)
          (22,100.00)
          (23,100.00)
          (24,0.00)
          (25,0.00)
          (26,0.00)
          (27,100.00)
          (28,0.00)
          (29,0.00)
        };

      \addplot [bar shift=-3.2pt, fill=reframed-silver, draw=white, line width=0.3pt,
        mark=none, forget plot]
      coordinates {
          (1,100.00)
          (2,100.00)
          (3,99.06)
          (4,92.97)
          (5,84.83)
          (6,70.00)
          (7,45.77)
          (8,35.54)
          (9,33.33)
          (10,14.44)
          (11,8.43)
          (12,11.23)
          (13,10.40)
          (14,7.94)
          (15,11.02)
          (16,7.92)
          (17,5.88)
          (18,4.76)
          (19,6.56)
          (20,2.63)
          (21,11.11)
          (22,4.55)
          (23,8.33)
          (24,0.00)
          (25,0.00)
          (26,0.00)
          (27,0.00)
          (28,0.00)
          (29,0.00)
        };

      \addplot [bar shift=3.2pt, fill=reframed-gold, draw=white, line width=0.3pt,
        mark=none, forget plot]
      coordinates {
          (1,0.00)
          (2,100.00)
          (3,99.06)
          (4,92.40)
          (5,80.34)
          (6,64.14)
          (7,36.46)
          (8,23.62)
          (9,12.75)
          (10,2.08)
          (11,0.43)
          (12,2.96)
          (13,1.29)
          (14,1.42)
          (15,2.78)
          (16,1.64)
          (17,2.56)
          (18,0.00)
          (19,0.00)
          (20,0.00)
          (21,0.00)
          (22,0.00)
          (23,0.00)
          (24,0.00)
          (25,0.00)
          (26,0.00)
          (27,0.00)
          (28,0.00)
          (29,0.00)
        };

      \addplot [draw=none, fill=none, mark=none, forget plot,
        nodes near coords, point meta=explicit symbolic,
        every node near coord/.style={
            rotate=0,
            anchor=south,
            xshift=-4pt,
            yshift=1pt,
            font=\fontsize{3.5}{4}\selectfont\itshape,
            inner sep=0pt,
          }]
      coordinates {
          (1,100.00)[{\textcolor{reframed-silver}{2}}]
          (2,100.00)[{\textcolor{reframed-silver}{177}}]
          (3,100.00)[{\textcolor{reframed-silver}{106}}]
          (4,100.00)[{\textcolor{reframed-silver}{185}}]
          (5,100.00)[{\textcolor{reframed-silver}{178}}]
          (6,100.00)[{\textcolor{reframed-silver}{210}}]
          (7,100.00)[{\textcolor{reframed-silver}{201}}]
          (8,100.00)[{\textcolor{reframed-silver}{166}}]
          (9,100.00)[{\textcolor{reframed-silver}{174}}]
          (10,100.00)[{\textcolor{reframed-silver}{187}}]
          (11,100.00)[{\textcolor{reframed-silver}{166}}]
          (12,100.00)[{\textcolor{reframed-silver}{187}}]
          (13,100.00)[{\textcolor{reframed-silver}{125}}]
          (14,100.00)[{\textcolor{reframed-silver}{126}}]
          (15,100.00)[{\textcolor{reframed-silver}{118}}]
          (16,100.00)[{\textcolor{reframed-silver}{101}}]
          (17,100.00)[{\textcolor{reframed-silver}{85}}]
          (18,100.00)[{\textcolor{reframed-silver}{63}}]
          (19,100.00)[{\textcolor{reframed-silver}{61}}]
          (20,100.00)[{\textcolor{reframed-silver}{38}}]
          (21,100.00)[{\textcolor{reframed-silver}{36}}]
          (22,100.00)[{\textcolor{reframed-silver}{22}}]
          (23,100.00)[{\textcolor{reframed-silver}{12}}]
          (24,100.00)[{\textcolor{reframed-silver}{4}}]
          (25,100.00)[{\textcolor{reframed-silver}{2}}]
          (26,100.00)[{\textcolor{reframed-silver}{3}}]
          (27,100.00)[{\textcolor{reframed-silver}{1}}]
          (28,100.00)[{\textcolor{reframed-silver}{1}}]
          (29,100.00)[{\textcolor{reframed-silver}{1}}]
        };

      \addplot [draw=none, fill=none, mark=none, forget plot,
        nodes near coords, point meta=explicit symbolic,
        every node near coord/.style={
            rotate=0,
            anchor=south,
            xshift=4pt,
            yshift=1pt,
            font=\fontsize{3.5}{4}\selectfont\itshape,
            inner sep=0pt,
          }]
      coordinates {
          (2,100.00)[{\textcolor{reframed-gold}{260}}]
          (3,100.00)[{\textcolor{reframed-gold}{213}}]
          (4,100.00)[{\textcolor{reframed-gold}{250}}]
          (5,100.00)[{\textcolor{reframed-gold}{290}}]
          (6,100.00)[{\textcolor{reframed-gold}{304}}]
          (7,100.00)[{\textcolor{reframed-gold}{288}}]
          (8,100.00)[{\textcolor{reframed-gold}{271}}]
          (9,100.00)[{\textcolor{reframed-gold}{204}}]
          (10,100.00)[{\textcolor{reframed-gold}{240}}]
          (11,100.00)[{\textcolor{reframed-gold}{230}}]
          (12,100.00)[{\textcolor{reframed-gold}{203}}]
          (13,100.00)[{\textcolor{reframed-gold}{155}}]
          (14,100.00)[{\textcolor{reframed-gold}{141}}]
          (15,100.00)[{\textcolor{reframed-gold}{72}}]
          (16,100.00)[{\textcolor{reframed-gold}{61}}]
          (17,100.00)[{\textcolor{reframed-gold}{39}}]
          (18,100.00)[{\textcolor{reframed-gold}{29}}]
          (19,100.00)[{\textcolor{reframed-gold}{18}}]
          (20,100.00)[{\textcolor{reframed-gold}{7}}]
          (21,100.00)[{\textcolor{reframed-gold}{7}}]
          (22,100.00)[{\textcolor{reframed-gold}{1}}]
          (23,100.00)[{\textcolor{reframed-gold}{2}}]
          (27,100.00)[{\textcolor{reframed-gold}{1}}]
        };

      \addlegendimage{area legend, fill=reframed-silver, draw=white, line width=0.3pt}
      \addlegendentry{Val: $\geq$ 1 match}
      \addlegendimage{area legend, fill=reframed-gray, draw=white, line width=0.3pt}
      \addlegendentry{Val: 0 matches}
      \addlegendimage{area legend, fill=reframed-gold, draw=white, line width=0.3pt}
      \addlegendentry{Test: $\geq$ 1 match}
      \addlegendimage{area legend, fill=reframed-yellow, draw=white, line width=0.3pt}
      \addlegendentry{Test: 0 matches}

    \end{axis}
  \end{tikzpicture}
}
    \caption{Proportion of \emph{dialogue subtitle} segments with at least one exact n-gram match (a direct match or match with newlines replaced with spaces) in the LLM evaluator's training corpus across validation and test splits. The numerical values above each bar indicate the absolute count of segments evaluated at that specific length. In both splits, we find matches for grams up to length 17 tokens, demonstrating contamination. The contamination is noticeably higher for the validation set as compared to the test set (which contains newer movies from 2023--2025).}
    \label{fig:contamination_subs}
\end{figure}

\section{LLM AD Generation}

\subsection{Prompt}\label{sec:prompts}
\begin{tcolorbox}[colback=white, colframe=reframed-silver, sharp corners,
  boxrule=0.5mm, breakable, title=LLM Prompt]
\begin{Verbatim}[fontsize=\small, breaklines=true, breakanywhere=true, breaksymbolleft={}]
Task description:

You are provided with the video (no audio) and dialogue subtitles for a full movie. Your task is to generate an audio description script for one part of the movie.

Audio description is a verbal narration of key visual content in a video. Descriptions must be inserted into gaps between dialogue.

Output format:

Return exactly one valid JSON object and no other text. Each key represents a gap between dialogue. Each value is an array of description objects. Each description object contains a textual description and timestamps for when it should be narrated. The following illustrates the required structure:

{"1st DIALOGUE GAP START --> DIALOGUE GAP END": [{"text": "TEXT", "start": "START", "end": "END"}, {"text": "TEXT", "start": "START", "end": "END"}, {"text": "TEXT", "start": "START", "end": "END"}], "2nd DIALOGUE GAP START --> DIALOGUE GAP END": [{"text": "TEXT", "start": "START", "end": "END"}]}

- DIALOGUE GAP START --> DIALOGUE GAP END takes the form "TIMESTAMP_FORMAT --> TIMESTAMP_FORMAT" and is the temporal interval when descriptions can be inserted.
- TEXT is a textual description.
- START and END are timestamps with format "TIMESTAMP_FORMAT".

Rules:

- Each value must be an array of zero, one, or more description objects, each containing exactly the keys "text", "start", and "end".
- For each description object:
  - TEXT is one sentence or shorter
  - # words in TEXT approx 3 x (END - START) in seconds
  - START >= DIALOGUE GAP START
  - END <= DIALOGUE GAP END
  - END > START
- Description objects must never overlap in time

Audio description script for prior parts:

INSERT_PRIOR_AUDIO_DESCRIPTION

Required part:

You should now generate an audio description script for the part of the movie that starts at PART_START_TIME and ends at PART_END_TIME.

Gaps between dialogue in the required part of the movie:

The following is the required output template, with the dialogue gaps in the required part of the movie included as keys:

OUTPUT_TEMPLATE

Generate the JSON with the values filled in. Your JSON must include all keys in the template above.
\end{Verbatim}
\end{tcolorbox}

For chunked video inputs, we adapt the prompt to state that the supplied video is one part of the movie rather than the full movie, remove the absolute start and end times of the required part, and express the subtitles, dialogue gaps, and generated timestamps relative to the beginning of the chunk.

Video (1FPS, no audio) is the first part of every prompt, followed by dialogue subtitles.
Dialogue subtitles take the form \texttt{START --> END TEXT}. For dialogue gaps, we include those that are at least one second in duration.
Timestamps are formatted according to model video preprocessing (\texttt{MM:SS} for Gemini and \texttt{<S.S seconds>} for Qwen).
For chunked video generation, relative output timestamps are converted to absolute movie times.

Iterative prompts include previously generated descriptions that were successfully parsed and validated against the prompt instructions:
\begin{itemize}
\itemsep0pt
    \item Malformed JSON (between first and last braces) and responses terminating due to maximum output length are treated as empty output.
    \item We retain only keys that match an eligible dialogue gap and whose values are arrays. For Qwen, we additionally accept otherwise exact gap keys from which the angle brackets around timestamps are missing.
    \item We retain only description objects containing the required \texttt{text}, \texttt{start}, and \texttt{end} fields, with nonempty textual descriptions and parseable timestamps.
    \item We require each segment to have a strictly positive duration and to be fully contained within the dialogue gap identified by its key.
    \item Within each gap, we remove duplicate segments, order the remaining segments by start and end time, and discard any segment that overlaps the preceding retained segment.
\end{itemize}
For chunked video input with iterative generation on our challenge set, timestamps risk becoming ambiguous.
From the second ten-minute chunk onwards, timestamps in the accumulated AD script refer to the movie timeline, while those in the required output refer to the current chunk's timeline.
To address this ambiguity, we compared providing the AD script for prior chunks as prose text, without timestamps, against providing it in the script format.
Results are in Table~\ref{tab:chunked-prompt}.
The two prompting approaches are largely indistinguishable, with small improvements in the QA-based metrics for the prose variant, which we therefore adopt.

\begin{table}[h]
    \centering
    \begin{small}
\begin{tabular}{l *{6}{S[table-format=2.1, table-column-width=1.2cm]}}
    \toprule
    & \multicolumn{1}{r}{\bf CIDEr} & \multicolumn{1}{r}{\bf METEOR} & \multicolumn{1}{r}{\bf SODA-M} & \multicolumn{1}{r}{\bf SODA-T} & \multicolumn{1}{r}{\bf QEval} & \multicolumn{1}{r}{\bf QEval-T} \\
    \midrule
    Qwen prose & 13.2 {~(0--2)} & 8.1 {~(0--0)} & 7.4 {~(0--2)} & 38.4 {~(2--0)} & 43.9 {~(3--0)} & 17.3 {~(3--0)} \\
    Qwen script & 13.3 {~(0--2)} & 7.8 {~(0--0)} & 7.4 {~(0--2)} & 35.4 {~(0--1)} & 43.4 {~(2--1)} & 16.7 {~(1--1)} \\
    Gemini prose & 19.0 {~(2--0)} & 8.2 {~(0--0)} & 7.9 {~(2--0)} & 36.0 {~(0--0)} & 42.3 {~(1--2)} & 16.4 {~(1--1)} \\
    Gemini script & 19.6 {~(2--0)} & 8.2 {~(0--0)} & 7.9 {~(2--0)} & 34.6 {~(0--1)} & 40.9 {~(0--3)} & 15.8 {~(0--3)} \\
    \bottomrule
    \end{tabular}
    \end{small}
    \caption{Iterative prompting strategy for chunked video input on our challenge set. Numbers in brackets represent wins and losses in the significance tests for that metric (W--L).}
    \label{tab:chunked-prompt}
\end{table}

\newpage
\subsection{Inference Configuration}
\label{appx:inference-configuration}

Table~\ref{tab:inference-params} shows our inference configurations used with the Qwen 3.5 27B and Gemini 3.1 Flash-Lite LLMs.

\begin{table}[H]
\centering
\small

\begin{tabular}{@{}lll@{}}
\toprule
\textbf{Parameter} & \textbf{Gemini 3.1 Flash-Lite} & \textbf{Qwen3.5-27B} \\
\midrule
Temperature & 1.0  & 1.0   \\
Top-$p$     & 0.95 & 0.95  \\
Top-$k$     & 64   & 20    \\
Min-$p$     & --   & 0.0   \\
Presence penalty     & --   & 1.5   \\
Repetition penalty   & --   & 1.0   \\
Max output tokens    & 65\,536      & 81\,920 \\
Thinking      & \texttt{high} & enabled  \\
Frame rate (FPS)      & 1  & 1  \\
\texttt{longest\_edge}   & --   & 469\,762\,048 \\
\texttt{shortest\_edge}  & --   & 4\,096 \\
RoPE type       & --   & YaRN ($\theta{=}10^7$, factor${=}4$) \\
\bottomrule
\end{tabular}
\caption{\label{tab:inference-params}Inference parameters for each LLM.}
\end{table}

\newpage

\section{Complementary Results}\label{appx:sec:extra-results}

\begin{table}[h]
\centering
\begin{small}
\begin{tabular}{l *{6}{S[table-format=2.1, table-column-width=1.2cm]}}
\toprule
& \multicolumn{1}{r}{\bf CIDEr} & \multicolumn{1}{r}{\bf METEOR} & \multicolumn{1}{r}{\bf SODA-M} & \multicolumn{1}{r}{\bf SODA-T} & \multicolumn{1}{r}{\bf QEval} & \multicolumn{1}{r}{\bf QEval-T} \\
\midrule
Expert & 51.4  & 20.1 & 16.3  & 81.0 & 69.6 & 61.2  \\
\midrule
Random & 1.3 {~(1--0)} & 4.4 {~(1--0)} & 4.0 {~(1--0)} & 27.1 {~(1--0)} & 34.1 {~(0--1)} & 2.3 {~(1--0)} \\
Empty & 0.0 {~(0--1)} & 0.0 {~(0--1)} & 0.0 {~(0--1)} & 0.0 {~(0--1)} & 41.4 {~(1--0)} & 0.0 {~(0--1)} \\
\midrule
Gemini (c) & 19.0 {~(4--0)} & 8.2 {~(4--0)} & 7.9 {~(5--0)} & 36.0 {~(3--0)} & 42.3 {~(3--1)} & 16.4 {~(4--1)} \\
Gemini (c, non-it.) & 18.1 {~(4--0)} & 7.9 {~(3--1)} & 7.7 {~(3--1)} & 34.5 {~(3--1)} & 42.0 {~(3--1)} & 15.7 {~(3--2)} \\
Gemini & 12.2 {~(1--2)} & 6.1 {~(1--3)} & 6.8 {~(1--3)} & 27.1 {~(1--3)} & 40.1 {~(0--4)} & 10.6 {~(2--3)} \\
Gemini (non-it.) & 10.5 {~(0--4)} & 5.6 {~(0--4)} & 6.5 {~(0--4)} & 23.4 {~(0--4)} & 40.4 {~(1--3)} & 9.4 {~(0--4)} \\
Qwen (c) & 13.2 {~(2--2)} & 8.1 {~(3--0)} & 7.4 {~(3--1)} & 38.4 {~(4--0)} & 43.9 {~(5--0)} & 17.3 {~(5--0)} \\
Qwen & 10.3 {~(0--3)} & 6.1 {~(0--3)} & 6.8 {~(0--3)} & 26.0 {~(0--3)} & 40.0 {~(0--3)} & 9.4 {~(0--4)} \\
\midrule
NarrAD (w/o cur.) & 13.3 {~(0--4)} & 7.9 {~(6--0)} & 8.5 {~(3--0)} & 55.0 {~(5--0)} & 45.5 {~(6--0)} & 19.1 {~(6--0)} \\
NarrAD & 14.6 {~(1--0)} & 7.3 {~(1--1)} & 8.5 {~(3--0)} & 49.7 {~(1--2)} & 43.3 {~(5--1)} & 18.7 {~(5--1)} \\
ShotbyShot (GPT4o) & 16.3 {~(2--0)} & 7.0 {~(1--1)} & 8.0 {~(1--3)} & 53.0 {~(5--0)} & 39.9 {~(4--2)} & 17.0 {~(4--2)} \\
UniAD & 15.9 {~(2--0)} & 7.3 {~(2--1)} & 8.2 {~(1--0)} & 48.5 {~(0--2)} & 36.8 {~(3--3)} & 11.3 {~(2--3)} \\
DistinctAD & 15.6 {~(2--0)} & 7.1 {~(1--1)} & 8.3 {~(3--0)} & 49.0 {~(0--2)} & 35.7 {~(1--4)} & 8.8 {~(0--6)} \\
ShotbyShot (Q/L) & 14.8 {~(1--0)} & 6.9 {~(1--2)} & 8.0 {~(1--3)} & 46.7 {~(0--3)} & 35.8 {~(1--4)} & 11.5 {~(2--3)} \\
AutoAD Zero & 13.4 {~(0--4)} & 6.5 {~(0--6)} & 7.6 {~(0--6)} & 47.1 {~(0--2)} & 34.7 {~(0--6)} & 9.8 {~(1--5)} \\
\bottomrule
\end{tabular}
\end{small}
\caption{Challenge set system performance under the six evaluation metrics. Numbers in brackets represent wins and losses in the significance tests for that metric (W--L), within a category. c=chunk, it.=iterative, cur.=curation, Q/L=Qwen/Llama.}
\label{tab:full-challenge-results}
\end{table}

\begin{table}[h]
\centering
\begin{small}
\begin{tabular}{l *{6}{S[table-format=2.1, table-column-width=1.2cm]}}
\toprule
& \multicolumn{1}{r}{\bf CIDEr} & \multicolumn{1}{r}{METEOR} & \multicolumn{1}{r}{\bf SODA-M} & \multicolumn{1}{r}{\bf SODA-T} & \multicolumn{1}{r}{\bf QEval} & \multicolumn{1}{r}{\bf QEval-T} \\
\midrule
Gemini 3.1 w/ faces & 22.5 {~(4--0)} & 9.7 {~(3--0)} & 8.9 {~(4--0)} & 52.2 {~(1--0)} & 44.7 {~(1--0)} & 23.1 {~(1--0)} \\
Qwen 3.5 w/ faces & 18.0 {~(2--1)} & 9.6 {~(3--0)} & 8.2 {~(2--1)} & 55.7 {~(2--0)} & 43.8 {~(1--0)} & 21.7 {~(1--0)} \\
Qwen 3.5 & 12.8 {~(1--2)} & 7.4 {~(1--2)} & 6.7 {~(1--2)} & 54.0 {~(2--0)} & 42.3 {~(1--0)} & 20.2 {~(1--0)} \\
Gemini 3.1 & 14.5 {~(1--1)} & 7.3 {~(1--2)} & 7.3 {~(1--1)} & 49.6 {~(1--2)} & 42.2 {~(1--0)} & 21.9 {~(1--0)} \\
Random & 1.1 {~(0--4)} & 3.9 {~(0--4)} & 3.8 {~(0--4)} & 39.5 {~(0--4)} & 32.4 {~(0--4)} & 7.8 {~(0--4)} \\
\bottomrule
\end{tabular}

\end{small}
\caption{Test set system performance under the six evaluation metrics. Numbers in brackets represent wins and losses in the significance tests for that metric (W--L). The systems are sorted by the sum of net wins across all metrics.}
\label{tab:test-set}
\end{table}

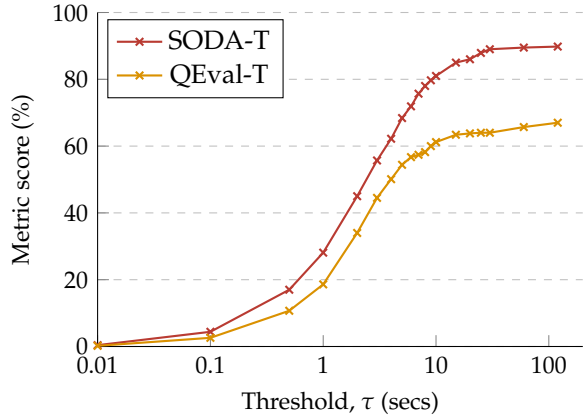
\begin{figure}[H]
    \centering
    \begin{tikzpicture}
  \begin{axis}[
      xmin=0.01, xmax=200,
      ymin=0, ymax=100,
      ymajorgrids=true,
      xlabel={Threshold, $\tau$ (secs)},
      ylabel={Metric score (\%)},
      xmode=log,
      log basis x=10,
      grid style=dashed,
      width=8cm,
      height=6cm,
      ylabel style={font=\small},
      xlabel style={font=\small},
      xticklabel style={font=\small},
      yticklabel style={font=\small},
      xtick={0.01, 0.1, 1, 10, 100},
      xticklabels={0.01, 0.1, 1, 10, 100},
      axis x line*=bottom,
      axis y line*=left,
      legend style={
          at={(0.02,0.98)},
          anchor=north west
        },
    ]
    \addplot[
      color=reframed-red,
      thick,
      mark=x,
      mark options={solid}
    ] coordinates {
        (0.01,  0.4)
        (0.1,   4.4)
        (0.5,  17.0)
        (1.0,  28.1)
        (2.0,  45.0)
        (3.0,  55.7)
        (4.0,  62.2)
        (5.0,  68.4)
        (6.0,  71.9)
        (7.0,  75.7)
        (8.0,  78.0)
        (9.0,  79.7)
        (10.0, 81.0)
        (15.0, 85.0)
        (20.0, 86.0)
        (25.0, 87.9)
        (30.0, 89.0)
        (60.0, 89.5)
        (120.0, 89.8)
      };
    \addlegendentry{SODA-T}

    \addplot[
      color=reframed-gold,
      thick,
      mark=x,
      mark options={solid}
    ] coordinates {
        (0.01,  0.2)
        (0.1,   2.6)
        (0.5,  10.7)
        (1.0,  18.6)
        (2.0,  34.0)
        (3.0,  44.5)
        (4.0,  50.1)
        (5.0,  54.4)
        (6.0,  56.7)
        (7.0,  57.4)
        (8.0,  58.2)
        (9.0,  60.0)
        (10.0, 61.2)
        (15.0, 63.4)
        (20.0, 63.8)
        (25.0, 64.0)
        (30.0, 64.0)
        (60.0, 65.7)
        (120.0, 67.0)
      };
    \addlegendentry{QEval-T}

  \end{axis}
\end{tikzpicture}
    \caption{Sweep of the threshold used in the SODA-T and QEval-T evaluation metrics. Results based on the expert human upper bound on the challenge set. The metrics plateau with scores lower than 100\%: SODA-T relies on alignment and unaligned description elements never score; QEval-T relies also on QA answer correctness, so tends towards QEval.}
    \label{fig:attribution-sweep}
\end{figure}

\begin{table}[t]
\centering
\small
\begin{tabular}{l *{10}{S[table-format=2.1, table-column-width=0.25cm]}}
\toprule
& \multicolumn{5}{c}{\textbf{CIDEr}}
& \multicolumn{5}{c}{\textbf{METEOR}} \\
\cmidrule(lr){2-6}
\cmidrule(lr){7-11}

\textbf{Gap}
& \multicolumn{1}{c}{\textbf{Expert}}
& \multicolumn{2}{c}{\textbf{Qwen 3.5}}
& \multicolumn{2}{c}{\textbf{Gemini 3.1}}
& \multicolumn{1}{c}{\textbf{Expert}}
& \multicolumn{2}{c}{\textbf{Qwen 3.5}}
& \multicolumn{2}{c}{\textbf{Gemini 3.1}} \\
\cmidrule(lr){3-4}
\cmidrule(lr){5-6}
\cmidrule(lr){8-9}
\cmidrule(lr){10-11}

\textbf{length} &
& \multicolumn{1}{c}{\textbf{Full}}
& \multicolumn{1}{c}{\textbf{Chunk}}
& \multicolumn{1}{c}{\textbf{Full}}
& \multicolumn{1}{c}{\textbf{Chunk}}
&
& \multicolumn{1}{c}{\textbf{Full}}
& \multicolumn{1}{c}{\textbf{Chunk}}
& \multicolumn{1}{c}{\textbf{Full}}
& \multicolumn{1}{c}{\textbf{Chunk}} \\
\midrule
1--5s   & 57.1 & 10.1 & 12.0 & 12.4 & 19.7 & 18.3 & 5.8 & 7.3 & 5.7 & 8.0 \\
5--10s  & 43.5 & 11.0 & 17.4 & 13.0 & 22.3 & 19.2 & 6.5 & 8.9 & 6.6 & 8.9 \\
10--20s & 50.0 & 5.4  & 11.2 & 7.4  & 11.7 & 21.9 & 5.8 & 8.7 & 6.6 & 8.8 \\
20--30s & 41.6 & 1.5  & 6.6  & 2.0  & 3.8  & 20.8 & 4.9 & 8.6 & 5.6 & 8.1 \\
30--60s & 22.6 & 0.8  & 2.0  & 1.7  & 1.2  & 21.4 & 5.3 & 8.2 & 5.7 & 7.6 \\
$>$60s  & 23.3 & 1.4  & 1.2  & 0.0  & 0.4  & 22.1 & 5.3 & 8.5 & 5.6 & 6.8 \\
\bottomrule
\end{tabular}
\caption{CIDEr and METEOR-based performance across varying dialogue gap lengths. Note the drop in CIDEr scores for dialogue gaps of length greater than 20 seconds, which is due to a length-difference penalty term in the metric. Note also particularly good Gemini performance for short gaps.}
\label{tab:gap_length}
\end{table}

\begin{table}[t]
\centering
\begin{small}
\begin{tabular}{lrrrrrrrrrr}
\toprule
\textbf{System}
& \multicolumn{2}{c}{\textbf{METEOR}}
& \multicolumn{2}{c}{\textbf{SODA-M}}
& \multicolumn{2}{c}{\textbf{SODA-T}}
& \multicolumn{2}{c}{\textbf{QEval}}
& \multicolumn{2}{c}{\textbf{QEval-T}} \\
\cmidrule(lr){2-3}
\cmidrule(lr){4-5}
\cmidrule(lr){6-7}
\cmidrule(lr){8-9}
\cmidrule(lr){10-11}
& \textbf{US} & \textbf{UK}
& \textbf{US} & \textbf{UK}
& \textbf{US} & \textbf{UK}
& \textbf{US} & \textbf{UK}
& \textbf{US} & \textbf{UK} \\
\midrule
Expert & 16.5 & 18.0 & 14.4 & 15.5 & 80.4 & 73.0 & 69.0 & 70.0 & 60.0 & 62.3 \\ \midrule
Qwen 3.5 & 4.8 & 5.3 & 6.0 & 6.6 & 20.9 & 21.8 & 39.6 & 40.3 & 8.8 & 9.8 \\
Qwen 3.5 (chunk) & 6.7 & 7.2 & 6.7 & 7.1 & 32.5 & 33.4 & 43.6 & 44.4 & 16.7 & 17.9 \\
Gemini 3.1 & 4.9 & 5.2 & 6.0 & 6.6 & 21.8 & 22.8 & 40.2 & 40.2 & 11.0 & 10.3 \\
Gemini 3.1 (chunk) & 6.7 & 7.0 & 7.0 & 7.6 & 29.8 & 31.5 & 42.5 & 42.3 & 17.0 & 15.9 \\
\bottomrule
\end{tabular}
\end{small}
\caption{Downstream performance against each of our two references (US and UK) individually. CIDEr results cannot be reported here as it relies on both references jointly.}
\label{tab:per-reference}
\end{table}

\begin{table}[t]
\centering
\begin{small}
\begin{tabular}{lrcr}
\toprule
 & \multicolumn{3}{c}{\textbf{Words per minute}} \\
\cmidrule(lr){2-4}
& \textbf{Full movie} & \textbf{Chunk} & \textbf{Test} \\
\midrule
US reference & 196.2 & ---    & 195.0 \\
UK reference & 201.5 & ---    & 197.6 \\
Gemini 3.1   & 110.3 & 118.3 & 157.0 \\
Qwen 3.5     & 104.9 & 145.0 & 182.9 \\
\bottomrule
\end{tabular}
\end{small}
\caption{Speech rate for professional AD references and LLM generations. References are tightly centred around roughly 200 words per minute, close to guideline recommendations. In contrast, Gemini and Qwen show inconsistent timing behaviour: with increasing length video input, generations become overly brief. On the test set of shorter videos, the models generate approximately three words per second, in accordance with their instructions.}
\label{tab:wpm}
\end{table}

\vspace{10cm}

\end{document}